\def\year{2016}\relax
\documentclass[letterpaper]{article}
\usepackage{aaai16}
\usepackage{times}
\usepackage{helvet}
\usepackage{courier}
\usepackage{floatrow}
\usepackage{algorithm}% http://ctan.org/pkg/algorithms
\usepackage[noend]{algpseudocode}
\usepackage{amsmath}
\usepackage{amssymb}
\usepackage{graphicx}
\usepackage{cleveref}
\usepackage{url}
\usepackage{tikz}
\newcommand{\squishlisttwo}{
 \begin{list}{$\bullet$}
  { \setlength{\itemsep}{0pt}
     \setlength{\parsep}{0pt}
    \setlength{\topsep}{0pt}
    \setlength{\partopsep}{0pt}
    \setlength{\leftmargin}{1em}
    \setlength{\labelwidth}{1.5em}
    \setlength{\labelsep}{0.5em} } }

\newcommand{\squishend}{
  \end{list}  }
  
\newcommand{\Sdom}{\mathcal{S}}
\newcommand{\Adom}{\mathcal{A}}
\newcommand{\Zprior}{\mathbf{z}_0}
\newcommand{\Sprior}{\mathbf{s}_0}
\newcommand{\Zhist}[1]{\mathbf{z}_{#1}}
\newcommand{\Shist}[1]{\mathbf{s}_{#1}}
\newcommand{\argmax}{\operatornamewithlimits{argmax}}
\newcommand{\ubar}[1]{\underline{#1}}
\newcommand{\obar}[1]{\overline{#1}}
\newcommand{\sdk} {\tau} % k, in my report, overlaps with covariance function
\newcommand{\rfn} {{R}} % reward function
\newtheorem{definition}{Definition}
\newtheorem{theorem}{Theorem}
\newtheorem{lemma}{Lemma}

\newtheorem{corollary}{Corollary}
\def\myproof{1} 
\nocopyright

\crefformat{footnote}{#2\footnotemark[#1]#3}

\begin{document}

\title{Gaussian Process Planning with Lipschitz Continuous Reward Functions: Towards Unifying  Bayesian Optimization, Active Learning, and Beyond}

\author{
Chun Kai Ling$^{\ast}$ \and Kian Hsiang Low$^{\ast}$ \and Patrick Jaillet$^{\dag}$\\
Department of Computer Science, National University of Singapore, Republic of Singapore$^{\ast}$\\
Department of Electrical Engineering and Computer Science, Massachusetts Institute of Technology, USA$^{\dag}$\\
\{chunkai, lowkh\}@comp.nus.edu.sg$^{\ast}$, jaillet@mit.edu$^{\dag}$
%Paper ID: \#2572\\
%Keywords: Non-myopic planning, Gaussian process, Bayesian optimization, active learning\\
%Topics: RU: Sequential Decision Making, ML: Active learning, ML: Reinforcement Learning,\\
%CSAI: Modeling and prediction of dynamic and spatiotemporal phenomena and systems
}
\maketitle

\begin{abstract}\begin{quote}\vspace{-1mm}
This paper presents a novel nonmyopic adaptive \emph{Gaussian process planning} (GPP) framework endowed with a general class of Lipschitz continuous reward functions that can unify some active learning/sensing and Bayesian optimization criteria  and offer practitioners some flexibility to specify their desired choices for defining new tasks/problems. In particular, it utilizes a principled Bayesian sequential decision problem framework for jointly and naturally optimizing the exploration-exploitation trade-off. In general, the resulting induced GPP policy cannot be derived exactly due to an uncountable set of candidate observations.
A key contribution of our work here thus lies in exploiting the Lipschitz continuity of the reward functions to solve for a nonmyopic adaptive \emph{$\epsilon$-optimal GPP} ($\epsilon$-GPP) policy.
To plan in real time, we further propose an asymptotically optimal, branch-and-bound anytime variant of $\epsilon$-GPP with performance guarantee.
We empirically demonstrate the effectiveness of our $\epsilon$-GPP policy and its anytime variant in Bayesian optimization and an energy harvesting task.\vspace{-1.9mm}
\end{quote}
\end{abstract}

\section{Introduction}\vspace{-0mm}
\label{sec:intro}
%
%To bridge the gap between recent advances in planning under uncertainty and machine learning, 
%To enhance the capability of planning and decision making under uncertainty algorithms to perform well and reliably in our real data-rich world,
The fundamental challenge of integrated planning and learning is to design an autonomous agent that can plan its actions to maximize its expected total rewards while interacting with an unknown task environment. 
Recent research efforts tackling this challenge have progressed from the use of simple Markov models assuming discrete-valued, independent observations (e.g., in \emph{Bayesian reinforcement learning} (BRL) \cite{Poupart2006}) to that of a rich class of Bayesian nonparametric \emph{Gaussian process} (GP) models characterizing continuous-valued, correlated observations
in order to represent the latent structure of more complex, possibly noisy task environments with higher fidelity. 
Such a challenge is posed by the following important problems in machine learning, among others:

\noindent
{\bf Active learning/sensing (AL).} In the context of environmental sensing (e.g., adaptive sampling in oceanography \cite{Leonard07}, traffic sensing \cite{LowUAI12,LowRSS13,LowTASE15}), its objective is to select the most informative (possibly noisy) observations for predicting a  spatially varying environmental field (i.e., task environment) modeled by a GP subject to some sampling budget constraints (e.g., number of sensors, energy consumption).
The rewards of an AL agent are defined based on some formal measure of predictive uncertainty such as the entropy or mutual information criterion. To resolve the issue of sub-optimality (i.e., local maxima) faced by greedy algorithms \cite{Guestrin08,LowAAMAS12,LowAAMAS14,YehongAAAI16}, recent developments have made nonmyopic AL computationally tractable with provable performance guarantees \cite{LowAAMAS13,NghiaICML14,LowICAPS09,LowAAMAS08,LowAAMAS11}, some of which have further investigated the performance advantage of adaptivity by proposing nonmyopic adaptive observation selection policies that depend on past observations.
%\cite{NghiaICML14,LowICAPS09} Singh09

\noindent
{\bf Bayesian optimization (BO).} Its objective is to select and gather the most informative (possibly noisy) observations for finding the global maximum of an unknown, highly complex (e.g., non-convex, no closed-form expression nor derivative) objective function (i.e., task environment) modeled by a GP given a sampling budget (e.g., number of costly function evaluations).
The rewards of a BO agent are defined using an improvement-based \cite{Brochu10} (e.g., \emph{probability of improvement} (PI) or \emph{expected improvement} (EI) over currently found maximum), entropy-based  \cite{Hennig12,Ghahramani14}, or \emph{upper confidence bound} (UCB) acquisition function \cite{Srinivas10}.
A limitation of most  BO algorithms is that they are myopic. %\cite{Hennig12,Ghahramani14,Srinivas10}, Villemonteix09
%as highlighted in \cite{Brochu10}. 
%Lizotte07,
To overcome this limitation, approximation algorithms for nonmyopic 
%non-adaptive \cite{Dey2014} and 
adaptive BO \cite{RamosUAI14,Osborne09} have been proposed, but their performances are not theoretically guaranteed.   
%We need to at least compare with EI, which outperforms PES in 1 dataset in that paper. If we do not compare with PES, I'm going to cite a weak reason given in his experimental section: "The entropy-based strategies (i.e., ES, PES) explore more aggressively than EI, thus resulting in their poorer performance in that dataset."

\noindent
{\bf General tasks/problems.} In practice, other types of rewards (e.g., logarithmic, unit step functions) need to be specified for an agent to plan and operate effectively in a given real-world task environment (e.g., natural phenomenon like wind or temperature) modeled by a GP, as detailed in Section~\ref{gppfram}.
%
%has exploited Monte Carlo tree search and an extreme assumption of maximum likelihood observations during planning for achieving efficient nonmyopic adaptive BO with UCB-based rewards, but has not provided any theoretical performance guarantee pertaining to the assumption. 
%
%
%The work in this paper is motivated by whether there are similarities in the structure of the above problems such that the challenge of integrated planning and control can 

As shall be elucidated later, similarities in the structure of the above problems motivate us to consider whether it is possible to tackle the overall challenge by devising a nonmyopic adaptive GP planning framework with a general class of reward functions unifying some AL and BO criteria and affording practitioners some flexibility to specify their desired choices for defining new tasks/problems. Such an integrated planning and learning framework has to address the exploration-exploitation trade-off common to the above problems: The agent faces a dilemma between gathering observations to maximize its expected total rewards given its current, possibly imprecise belief of the task environment (exploitation) vs. that to improve its belief to learn more about the environment (exploration).

This paper presents a novel nonmyopic adaptive \emph{Gaussian process planning} (GPP) framework endowed with a general class of Lipschitz continuous reward functions that can unify some AL and BO criteria (e.g., UCB) discussed earlier and offer practitioners some flexibility to specify their desired choices for defining new tasks/problems (Section~\ref{gppfram}).
In particular, it utilizes a principled Bayesian sequential decision problem framework for jointly and naturally optimizing the exploration-exploitation trade-off,
consequently allowing planning and learning to be integrated seamlessly and performed simultaneously 
%this eliminates the need to 
%therefore, unlike the UCB selection criterion of the greedy Bayesian optimization algorithm of \cite{Srinivas10}, it  does not have to explicitly consider an additional weighted exploration term (i.e., weighted GP posterior variance) in the reward function.
%Consequently, our proposed GPP framework can seamlessly integrate planning and learning and perform them simultaneously 
instead of separately \cite{Deisenroth15}.
In general, the resulting induced GPP policy cannot be derived exactly due to an uncountable set of candidate observations.
A key contribution of our work here thus lies in exploiting the Lipschitz continuity of the reward functions to solve for a nonmyopic adaptive \emph{$\epsilon$-optimal GPP} ($\epsilon$-GPP) policy given an arbitrarily user-specified loss bound $\epsilon$ (Section~\ref{eogpp}). 
%That is, our $\epsilon$-GPP policy can closely approximate the optimal expected total rewards achieved by the GPP policy within the loss bound $\epsilon$.
%, including for the special  case of maximum likelihood observations during planning.
%In contrast, the nonmyopic adaptive Bayesian optimization algorithm of \cite{RamosUAI14} has no provable performance guarantee pertaining to its extreme assumption of maximum likelihood observations.
%the optimal expected total rewards achieved by our $\epsilon$-GPP policy upper bounds that of .
To plan in real time, we further propose an asymptotically optimal, branch-and-bound anytime variant of $\epsilon$-GPP with performance guarantee.
Finally, we empirically evaluate the performances of our $\epsilon$-GPP policy and its anytime variant in BO and an energy harvesting task on simulated and real-world environmental fields (Section~\ref{expt}). 
%\footnote{Bayes-optimality is previously studied in reinforcement learning whose developed theories \cite{Poupart2006,NghiaIJCAI13b} cannot be applied here because their assumptions of discrete-valued observations and Markov property do not hold.} 
To ease exposition, the rest of this paper will be described by assuming the task environment to be an environmental field and the agent to be a mobile robot, which coincide with the setup of our experiments.\vspace{-2.3mm}
\section{Gaussian Process Planning (GPP)}\vspace{-0.2mm}
\label{gppfram}
{\bf Notations and Preliminaries.}
Let $\Sdom$ be the domain of an environmental field corresponding to a set of sampling  locations.
%A robot at location $s\in\Sdom$ can deterministically move to  choose from a finite set $\Adom(s) \subseteq \Sdom$ of reachable locations to  in a single time step.
%So, the robot at location $s$ may choose to move to any location $s' \in \Adom(s)$ in a single time step; 
%this transition is assumed to be , which is valid for large-scale applications.
At time step $t > 0$, a robot can deterministically move from its previous location $s_{t-1}$ to visit location $s_t\in\Adom(s_{t-1})$ and observes it by taking a corresponding realized (random) field measurement $z_t$ ($Z_t$) where $\Adom(s_{t-1}) \subseteq \Sdom$ denotes a finite set of sampling locations reachable from its previous location $s_{t-1}$ in a single time step. 
%The realization of $Z_t$ is denoted by $z_t$.
%
The state of the robot at its initial starting location $s_0$ is represented by prior observations/data $d_0\triangleq\langle \Sprior , \Zprior\rangle$ available before planning where 
$\Sprior$ and $\Zprior$ denote, respectively, vectors comprising locations visited/observed and corresponding field measurements taken by the robot prior to planning and $s_0$ is the last component of $\Sprior$.
Similarly, at time step $t>0$, the state of the robot at its current location $s_t$ is represented by observations/data $d_t \triangleq \langle \Shist{t}, \Zhist{t} \rangle$ where 
$\Shist{t}\triangleq\Sprior \oplus (s_1 , \cdots s_t)$ and $\Zhist{t}\triangleq\Zprior \oplus (z_1 , \cdots z_t)$ denote, respectively, vectors comprising locations visited/observed and corresponding field measurements taken by the robot up until time step $t$ and `$\oplus$' denotes vector concatenation.
%, and 
%Let $\Shist{t}\triangleq\Sprior \oplus (s_1 , \cdots s_t)$ be a vector containing all locations visited up to (and including) the location visited at time step $t$ and $\oplus$ denotes vector concatenation. Similarly, let $\Zhist{t}\triangleq\Zprior \oplus (z_1 , \cdots z_t)$ be a vector containing all measurements sampled until time step $t$. So, a robot's state after time step $t$ is represented by $d_t \triangleq \langle \Shist{t}, \Zhist{t} \rangle$ and its current location 
%$s_t$ is the last component of $\Shist{t}$.
At time step $t > 0$, the robot also receives a reward $\rfn(z_t, \Shist{t})$ to be defined later. \vspace{1mm}
%As before, denoted $r_t$ to be a realization of $R_t$.
%in Section~\ref{ch:rewardfunctions}.
%

\noindent
{\bf Modeling Environmental Fields with Gaussian Processes (GPs).}
\label{GPSSSS}
%
%The environmental field itself is modeled by a Gaussian Process (GP) and is denoted by $F$.
The GP can be used to model a spatially varying environmental field as follows:
The field is assumed to be a realization of a GP.
Each location $s \in \Sdom$ is associated with a latent field measurement $Y_s$.
Let $Y_{\Sdom} \triangleq \{Y_{s}\}_{s \in \Sdom}$ denote a GP, that is, every finite subset of $Y_{\Sdom}$ has a multivariate Gaussian distribution \cite{Rasmussen06}.
Then, the GP is fully specified by its \emph{prior} mean $\mu_s \triangleq \mathbb{E}[Y_s]$ and covariance 
$k_{ss'} \triangleq \text{cov}[Y_s,Y_{s'}]$ for all $s, s'\in \Sdom$,
the latter of which characterizes the spatial correlation structure of the environment field and can be defined using a covariance function.
A common choice is the squared exponential covariance function
$
%cov[F_{s}, F_{s'}] = \sigma_{\text{signal}}^2\exp\big(\frac{(s-s')^{\text{T}}M^{-2}(s-s')}{-2}\big) + \sigma_{\text{noise}}^2\delta_{pq}
k_{ss'} \triangleq \sigma_{y}^2\exp\{-0.5(s-s')^{\top}M^{-2}(s-s')\}$ where
$\sigma_{y}^2$ is the signal variance controlling the intensity of measurements and $M$ is a diagonal matrix with length-scale components $l_1$ and $l_2$ governing the degree of spatial correlation or ``similarity'' between measurements in the respective horizontal and vertical directions of the $2$D fields in our experiments.
%The squared exponential covariance function is adopted for the rest of the paper, although it should be noted that our results are in no way dependent on this choice.

The field measurements taken by the robot are assumed to be corrupted by Gaussian white noise, i.e., 
$Z_t \triangleq Y_{s_t}+\varepsilon$ where $\varepsilon\sim\mathcal{N}(0, \sigma_{n}^{2})$ and $\sigma_{n}^{2}$ is the noise variance.
Supposing the robot has gathered observations $d_t = \langle \Shist{t}, \Zhist{t} \rangle$ from time steps $0$ to $t$,
the GP model can perform probabilistic regression by using $d_t$ to predict the noisy measurement at any unobserved location $s_{t+1}\in\Adom(s_{t})$ as well as provide its  predictive uncertainty using a 
%Using a GP to model the environmental field allows us to perform probabilistic regression.
%Suppose the robot is in state $s_{t+1}$ and has past measurements and locations $d_t$.
%Conditioning on $d_t$ allows us to obtain the noisy 
Gaussian predictive distribution 
$p(z_{t+1}|d_t, s_{t+1})=\mathcal{N}(\mu_{s_{t+1}| d_t}, \sigma^2_{s_{t+1}|\Shist{t}})$ 
with the following \emph{posterior} mean and variance, respectively:\vspace{-1mm}
%\begin{equation} \label{eq:predictive}
%\hspace{-1.8mm}
%\begin{array}{rl}
$$
\begin{array}{rl}
\mu_{s_{t+1}| d_t}\hspace{-2.8mm} &\triangleq\displaystyle\mu_{s_{t+1}}+\Sigma_{s_{t+1}\Shist{t}} \Gamma_{\Shist{t}\Shist{t}}^{-1} (\Zhist{t}-\mu_{\Shist{t}})^{\top}\vspace{0.5mm}\\
\sigma^2_{s_{t+1}|\Shist{t}}\hspace{-2.8mm} &\triangleq \displaystyle k_{s_{t+1}s_{t+1}}+\sigma_{n}^2  - \Sigma_{s_{t+1}\Shist{t}} \Gamma_{\Shist{t}\Shist{t}}^{-1} \Sigma_{\Shist{t}s_{t+1}}\vspace{-1mm}
\end{array}
$$
%\end{array}
%\end{equation}
where $\mu_{\Shist{t}}$ is a row vector with mean components $\mu_s$ for every location $s$ of $\Shist{t}$, 
$\Sigma_{s_{t+1}\Shist{t}}$ is a row vector with covariance components $k_{s_{t+1}s}$ for every location $s$ of $\Shist{t}$, 
$\Sigma_{\Shist{t}s_{t+1}}$ is the transpose of $\Sigma_{s_{t+1}\Shist{t}}$, and $\Gamma_{\Shist{t}\Shist{t}}\triangleq\Sigma_{\Shist{t}\Shist{t}}+\sigma_{n}^2I$ such that $\Sigma_{\Shist{t}\Shist{t}}$ is a covariance matrix with components $k_{ss'}$ for every pair of locations $s, s'$ of $\Shist{t}$.
An important property of the GP model is that, unlike 
%GP posterior mean 
$\mu_{s_{t+1}| d_t}$, 
%GP posterior variance 
$\sigma^2_{s_{t+1}|\Shist{t}}$ is independent of 
%the realized measurements 
$\Zhist{t}$.\vspace{0.5mm}
%We assume that the hyperparameters $\{\sigma_\text{signal}, \sigma_{\text{noise}}, M\}$ are known in advance. This can be reasonably approximated with expert knowledge or by prior data.
%
%Before proceeding, we introduce some new notation. 
%For any vector $\mathbf{v}$, $|\mathbf{v}|$ denotes a vector containin absolute values of $\mathbf{v}$. That is, $|\mathbf{v}|_i=|\mathbf{v}_i|$ for all $i$.
%This should not be confused for the vector norm of $\mathbf{v}$.
%Additionally, we use $(\cdot)$ and $(\ast)$ as symbols representing the vector dot product and convolution between functions respectively.
%

\noindent
{\bf Problem Formulation.}
\label{proble}
To frame nonmyopic adaptive \emph{Gaussian process planning} (GPP) as a Bayesian sequential decision problem, let an adaptive policy $\pi$ be defined to sequentially decide the next location $\pi(d_t)\in\Adom(s_t)$ to be observed at each time step $t$ using observations $d_t$ over a finite planning horizon of $H$ time steps/stages (i.e., sampling budget of $H$ locations).
The value $V_0^\pi(d_0)$ under an adaptive policy $\pi$ is defined to be the expected total rewards achieved by its selected observations when starting with some prior observations $d_0$ and following $\pi$ thereafter and can be computed using the following $H$-stage Bellman equations:\vspace{-1mm}
%\begin{equation} 
%
%\hspace{-1.52mm}
$$
%\hspace{-1.8mm} 
 \begin{array}{rl}
  V_t^\pi(d_t) \hspace{-2.8mm} &\triangleq \displaystyle Q_t^\pi(d_t, \pi(d_{t}))\vspace{0.5mm}\\
  Q_t^\pi(d_t, s_{t+1}) \hspace{-2.8mm} &\triangleq \displaystyle \mathbb{E}[ \rfn(Z_{t+1}, \Shist{t+1})\ + \\
&\quad V_{t+1}^\pi(\langle \Shist{t+1}, \Zhist{t}\oplus Z_{t+1}\rangle)|d_t,s_{t+1} ]\vspace{-1mm}
\end{array}
$$  
%  \hspace{-1.51mm}
%\label{eq:ValFunDef}
%\end{equation}
for stages $t = 0,\ldots,H-1$ where $V_H^\pi(d_H)\triangleq 0$.
To solve the GPP problem, the notion of Bayes-optimality\if\myproof1\footnote{Bayes-optimality has been studied in discrete BRL \cite{Poupart2006} whose assumptions (Section~\ref{sec:intro}) do not hold in GPP. Continuous BRLs \cite{Ross09,Ross08} assume a known parametric form of observation function, the reward function to be independent of measurements and past states, and/or, when exploiting GP, maximum likelihood observations during planning with no provable performance guarantee.} \fi
%ICRA 2008 parametric form of transition and observation function is known.
%
%IROS 2009
%reward depends on current state
%but ours, reward depends additionally all past states and current observation
%
%max likelihood observation without guarantee.
%So, its solvers cannot be used here.
is exploited for selecting observations to achieve the largest possible expected total rewards with respect to all possible induced sequences of future Gaussian posterior beliefs $p(z_{t+1}|d_t, s_{t+1})$ for $t=0,\ldots,H-1$ to be discussed next.
Formally, this involves choosing an adaptive policy $\pi$ to maximize $V^{\pi}_0(d_0)$,  %\eqref{eq:ValFunDef}, 
which we call the GPP policy $\pi^\ast$. That is, $V^{\ast}_0(d_0)\triangleq V^{\pi^\ast}_0(d_0) = \max_\pi V^{\pi}_0(d_0)$. By plugging $\pi^\ast$  into $V_t^\pi(d_t)$ and $Q_t^\pi(d_t, s_{t+1})$ above,\vspace{-0.7mm}
%\eqref{eq:ValFunDef},
\begin{equation} 
\hspace{-1.2mm}
 \begin{array}{rl}
  V_t^*(d_t) \triangleq\hspace{-2.4mm} & \max_{s_{t+1} \in \Adom(s_{t})} Q_t^*(d_t, s_{t+1})\vspace{0.5mm}\\
  Q_t^*(d_t, s_{t+1}) \triangleq\hspace{-2.4mm} & \displaystyle \mathbb{E}[ \rfn(Z_{t+1}, \Shist{t+1})|d_t, s_{t+1} ]\ + \\
&  \mathbb{E}[V_{t+1}^*(\langle \Shist{t+1}, \Zhist{t}\oplus Z_{t+1}\rangle)|d_t, s_{t+1} ]\vspace{-0.6mm}
  \end{array}
  %\hspace{-1.51mm}
\label{eq:OptimalValFunDef}
\end{equation}
for stages $t = 0,\ldots,H-1$ where $V_H^*(d_H)\triangleq 0$.
To see how the GPP policy $\pi^\ast$ jointly and naturally optimizes the exploration-exploitation trade-off,
its selected location $\pi^\ast(d_t) = \arg\max_{s_{t+1} \in \Adom(s_{t})} Q_t^*(d_t, s_{t+1})$ at each time step $t$ affects both the immediate expected reward 
$\mathbb{E}[ \rfn(Z_{t+1}, \Shist{t}\oplus\pi^*(d_t))|d_t,\pi^*(d_t) ]$ given current belief $p(z_{t+1}|d_t, \pi^*(d_t))$ (i.e., exploitation) as well as the Gaussian posterior belief $p(z_{t+2}|\langle \Shist{t}\oplus\pi^*(d_t), \Zhist{t}\oplus z_{t+1}\rangle, \pi^*(\langle \Shist{t}\oplus\pi^*(d_t), \Zhist{t}\oplus z_{t+1}\rangle))$ at next time step $t+1$ (i.e., exploration), the latter of which influences expected future rewards $\mathbb{E}[V_{t+1}^*(\langle \Shist{t}\oplus\pi^*(d_t), \Zhist{t}\oplus Z_{t+1}\rangle)|d_t,\pi^*(d_t) ]$.

In general, the GPP policy $\pi^\ast$ cannot be derived exactly because the expectation terms in \eqref{eq:OptimalValFunDef} usually cannot be evaluated in closed form 
%\cite{Marco2008} 
due to an uncountable set of candidate measurements (Section~\ref{sec:intro}) 
except for degenerate cases like $\rfn(z_{t+1}, \Shist{t+1})$ being independent of $z_{t+1}$ 
%(Section~\ref{ch:rewardfunctions}) 
and $H\leq 2$. 
To overcome this difficulty, we will show in Section~\ref{eogpp} later how the Lipschitz continuity of the reward functions 
%(Section~\ref{ch:rewardfunctions}) 
can be exploited for 
theoretically guaranteeing the performance of our proposed nonmyopic adaptive $\epsilon$-optimal GPP policy, that is, the expected total rewards achieved by its selected observations closely approximates that of $\pi^{\ast}$ within an arbitrarily user-specified loss bound $\epsilon > 0$.\vspace{0.5mm}
%
% ($\epsilon$-GPP) policy given an arbitrarily user-specified loss bound $\epsilon$. That is, our $\epsilon$-GPP policy can  closely approximate the optimal expected total rewards achieved by the GPP policy within the loss bound $\epsilon$, including for the extreme case of maximum likelihood observations during planning.
%
%Attempting to compute the optimal policy or value function directly from \eqref{eq:OptimalValFunDef} is difficult.
%This is due to the expectation taken over a gaussian distribution, as well as the max operator in $V^*$.
%Thus, we are compelled to resort to approximations with provably correct error bounds. 
%This eventually bounds the policy loss of our proposed algorithm.
%
%

\noindent
{\bf Lipschitz Continuous Reward Functions.} 
\label{ch:rewardfunctions}
$\rfn(z_t, \Shist{t})\triangleq$ $\rfn_1(z_t) + \rfn_2(z_t) + \rfn_3(\Shist{t})$ where 
$\rfn_1$, $\rfn_2$, and $\rfn_3$ are user-de- fined reward functions that satisfy the conditions below:\vspace{-0mm}
\squishlisttwo
 \item $\rfn_{1}(z_t)$ is Lipschitz continuous in $z_t$ with Lipschitz constant $\ell_{1}$. 
So, $h_{\sigma}(u)\triangleq (\rfn_{1} \ast \mathcal{N}(0, \sigma^2))(u)$ 
 %= \int\rfn_{1}(z) p(u-z) \text{d}z = \int\rfn_{1}(u-z) p(z) \text{d}z
 %$\mathcal{N}$ refers to the density of any normal distribution some standard deviation $\sigma$.
%Then, $h_{\sigma}(u)$ 
is Lipschitz continuous in $u$ with $\ell_{1}$
where `$\ast$' denotes convolution;
 \item $\rfn_{2}(z_t)$: Define $g_{\sigma}(u)\triangleq(\rfn_{2} \ast \mathcal{N}(0, \sigma^2))(u)$ such that 
 %all the following conditions must be satisfied for any $\sigma$:
(a) $g_{\sigma}(u)$ is well-defined for all $u \in \mathbb{R}$,
 (b) $g_{\sigma}(u)$ can be evaluated in closed form or  computed up to an arbitrary precision in reasonable time for all $u \in \mathbb{R}$, and
(c) $g_{\sigma}(u)$ is Lipschitz continuous\footnote{\label{const}Unlike $\rfn_{1}$, $\rfn_{2}$ does not need to be Lipschitz continuous (or continuous); it must only be Lipschitz continuous after convolution with any Gaussian kernel. An example of $\rfn_{2}$ is  unit step function.} in $u$ with Lipschitz constant $\ell_{2}(\sigma)$;
 \item $\rfn_{3}(\Shist{t})$ only depends on locations $\Shist{t}$ visited/observed by the robot up until time step $t$ and is independent of realized measurement $z_{t}$.
It can be used to represent some sampling or motion costs or explicitly consider exploration by defining it as a function of $\sigma^2_{s_{t+1}|\Shist{t}}$.\vspace{-0mm}
\squishend
Using the above definition of $\rfn(z_t, \Shist{t})$, the immediate expected reward in \eqref{eq:OptimalValFunDef} evaluates to %\vspace{-1.6mm}
%Consider the immediate \textit{expected reward} of a robot with history $d_{t}$ just after moving to location $s_{t+1}$.
%Using the relationship between convolution and expectations, we obtain:
%\begin{equation} \label{eq:lipschitzmean}
% \begin{array}{l}
$
%\displaystyle 
\mathbb{E}[ \rfn(Z_{t+1}, \Shist{t+1})|d_t, s_{t+1} ]$ $= 
%\vspace{1mm}\\\displaystyle
(h_{\sigma_{s_{t+1}|\Shist{t}}} + g_{\sigma_{s_{t+1}|\Shist{t}}})\hspace{-1mm}\left(\mu_{s_{t+1}|d_{t}}\right) +\rfn_3(\Shist{t+1})%\vspace{-1.6mm}
$
% \end{array}
%\end{equation}
%Suppose $\Shist{t+1}$ is fixed. 
which is Lipschitz continuous in the realized measurements $\Zhist{t}$:
%, as shown in the following result:%\vspace{-1mm}
\begin{lemma} Let $\alpha(\Shist{t+1}) \triangleq \lVert\Sigma_{s_{t+1}\Shist{t}}\Gamma_{\Shist{t}\Shist{t}}^{-1}\rVert$ and $d_{t}'\triangleq\langle \Shist{t}, \Zhist{t}' \rangle$. %\vspace{-1mm} 
%\begin{equation}

\noindent
Then,$
%\hspace{-0.5mm}
% \begin{array}{l}
\displaystyle\left|\mathbb{E}[ \rfn(\hspace{-0.1mm}Z_{t+1},\hspace{-0.5mm} \Shist{t+1})|d_t,\hspace{-0.5mm} s_{t+1} ]\hspace{-0.8mm} - \hspace{-0.8mm}\mathbb{E}[ \rfn(\hspace{-0.1mm}Z_{t+1}, \hspace{-0.5mm}\Shist{t+1})|d'_t,\hspace{-0.5mm}s_{t+1} ] \right|$ $\leq \alpha(\Shist{t+1}) \left(\ell_1 + \ell_2(\sigma_{s_{t+1}|\Shist{t}})\right) \lVert\Zhist{t} - \Zhist{t}'\rVert%\vspace{-6mm}
% \end{array}
\ .$
% \label{geez}
%\end{equation}
\label{rewlc}
\end{lemma}
Its proof is in\if\myproof1 Appendix~\ref{geeez2}. \fi\if\myproof0 \cite{AA16}. \fi 
Lemma~\ref{rewlc} will be used to prove the Lipschitz continuity of $V^*_t$ in \eqref{eq:OptimalValFunDef} later.
%(Section~\ref{propertyv}), which can then be exploited for theoretically guaranteeing the performance of our nonmyopic adaptive $\epsilon$-optimal GPP policy. 
Before doing this, let us consider how the Lipschitz continuous reward functions defined above can unify some AL and BO criteria discussed in Section~\ref{sec:intro} and be used for defining new tasks/problems.\vspace{0.5mm}
%\textbf{Degenerate cases.} In the event that a mission is purely exploratory in nature, $\rfn(z_{t+1},\Shist{t+1})$ can be independent of $z_{t+1}$. These problems are prominent in Bayesian experimental design where one attempts to find a set of explored locations maximizing information gain that, in the context of GPs, only depends on locations. Closely related is the problem of non-adaptive maximum entropy sampling  where $\rfn(z_{t+1}, \Shist{t+1})=R_3(\Shist{t+1})=0.5\log(2 \pi e\sigma^2_{s_{t+1}|\Shist{t}})$ \cite{LowICAPS09}. Under these situations, planning reduces to a straightforward search algorithm and does not face the problem of having to consider an uncountable set of candidate measurements. As such, we will not focus on these applications.

\noindent
\textbf{Active learning/sensing (AL).} Setting $\rfn(z_{t+1}, \Shist{t+1})=R_3(\Shist{t+1})=0.5\log(2 \pi e\sigma^2_{s_{t+1}|\Shist{t}})$ yields the well-known nonmyopic AL algorithm called  \emph{maximum entropy sampling} (MES) \cite{Shewry87} which plans/decides locations with maximum entropy to be observed that
%, as proven in an equivalence result \cite{LowICAPS09}, 
minimize the posterior entropy remaining in the unobserved areas of the field.
Since $\rfn(z_{t+1}, \Shist{t+1})$ is independent of $z_{t+1}$, the expectations in \eqref{eq:OptimalValFunDef} go away, thus making MES non-adaptive and hence a straightforward search algorithm not plagued by the issue of uncountable set of candidate measurements. 
%Maximum mutual information sampling \cite{LowAAMAS13} can be analyzed likewise. 
As such, we will not focus on such a degenerate case.
This degeneracy vanishes when the environment field is instead a realization of log-Gaussian process.
Then, MES becomes adaptive \cite{LowICAPS09} and its reward function can be represented by our Lipschitz continuous reward functions: By setting $R_1(z_{t+1})=0$, $R_2$ and $g_{\sigma_{s_{t+1}|\Shist{t}}}$ as identity functions with $\ell_2(\sigma_{s_{t+1}|\Shist{t}})=1$, and $R_3(\Shist{t+1})=0.5\log(2 \pi e\sigma^2_{s_{t+1}|\Shist{t}})$, 
$\mathbb{E}[ \rfn(Z_{t+1}, \Shist{t+1})|d_t, s_{t+1} ] =\mu_{s_{t+1}| d_t} + 0.5\log(2 \pi e\sigma^2_{s_{t+1}|\Shist{t}})$.
% will resolve to the form $\rfn(Z_{t+1}, \Shist{t+1})=\log(\sqrt{(2 \pi e)^{i}|\Sigma_{\Shist{t+1}, \Shist{t+1}}|})+\mu_{s_{t+1}|dt}$ . The latter term implies that since $\mu_{s_{t+1}|dt}$ depends on $\mu_{s_{t+1}|dt}$ which in turn depends on past measurements.

\noindent
\textbf{Bayesian optimization (BO).} The greedy BO algorithm of \citeauthor{Srinivas10}~\shortcite{Srinivas10} utilizes the UCB selection criterion $\mu_{s_{t+1}| d_t} + \beta\sigma_{s_{t+1}|\Shist{t}}$ ($\beta\geq 0$) to approximately optimize the global BO objective of total  field measurements $\sum^{H}_{t=1} z_t$ taken by the robot or, equivalently, minimize its total regret.
% \cite{Brochu10}.
%Bayesian optimization algorithms of   and \cite{RamosUAI14} utilize the UCB selection criterion of the form $\mu_{s_{t+1}| d_t} + \beta\sigma_{s_{t+1}|\Shist{t}}$ ($\beta>0$) as the reward function. 
UCB can be represented by our Lipschitz continuous reward functions: By setting $R_1(z_{t+1})=0$, $R_2$ and $g_{\sigma_{s_{t+1}|\Shist{t}}}$ as identity functions with $\ell_2(\sigma_{s_{t+1}|\Shist{t}})=1$, and $R_3(\Shist{t+1})=\beta\sigma_{s_{t+1}|\Shist{t}}$, 
$\mathbb{E}[ \rfn(Z_{t+1}, \Shist{t+1})|d_t, s_{t+1} ] =\mu_{s_{t+1}| d_t} + \beta\sigma_{s_{t+1}|\Shist{t}}$.
In particular, when $\beta=0$, it can be derived that our GPP policy $\pi^*$ maximizes the \emph{expected} total field measurements taken by the robot, 
hence optimizing the exact global BO objective of \citeauthor{Srinivas10}~\shortcite{Srinivas10} in the expected sense.
So, unlike greedy UCB, our nonmyopic GPP framework does not have to explicitly consider an additional weighted exploration term (i.e., $\beta\sigma_{s_{t+1}|\Shist{t}}$) in its reward function because it can jointly and naturally optimize the exploration-exploitation trade-off, as explained earlier.
Nevertheless, if a stronger exploration behavior is desired (e.g., in online planning), then $\beta$ has to be fine-tuned.
Different from nonmyopic BO algorithm of \citeauthor{RamosUAI14}~\shortcite{RamosUAI14} using UCB-based rewards, our proposed nonmyopic $\epsilon$-optimal GPP policy (Section~\ref{eogpp}) does not need to impose an extreme assumption of maximum likelihood observations during planning and, more importantly, provides a %theoretical 
performance guarantee, including for the extreme assumption made by nonmyopic UCB. 
Our GPP framework differs from  nonmyopic BO algorithm of \citeauthor{Osborne09}~\shortcite{Osborne09} in that every selected observation contributes to the total field measurements taken by the robot instead of considering just the expected improvement for the last observation.
%As a result, unlike \cite{Osborne09}, 
So, it usually does not have to expend all the given sampling budget to find the global maximum.\vspace{0.5mm}
%
%Following this, the work of \cite{RamosUAI14} has exploited Monte Carlo tree search and an extreme assumption of maximum likelihood observations during planning for achieving efficient nonmyopic adaptive BO with UCB-based rewards, but has not provided any theoretical performance guarantee pertaining to the assumption. 
%
%This expression has been derived in [12], however, it presents a slight modification because we are considering the whole sequence of locations for reward calculation, not just the expected improvement for the last sample.
%
% including for the special  case of maximum likelihood observations during planning.
%In contrast, the nonmyopic adaptive Bayesian optimization algorithm of \cite{RamosUAI14} has no provable performance guarantee pertaining to its extreme assumption of maximum likelihood observations.
%
%
%
%The confidence bound criteria used in Bayes optimization features reward functions of the form The addition of a positive multiple of the posterior variance encourages an exploratory behavior. In particular, we obtain the UCB selection criterion \cite{Srinivas10} if we allow $\tau$ to vary with time and set $\Adom(s)=\Sdom$. While the confidence bound criterion explicitly encourages exploration by assigning high rewards to regions with high posterior variance, our framework has no need for this. ``True'' Bayes-optimality is achieved by planning in the space of posteriors.

\noindent
\textbf{General tasks/problems.}
In practice, the necessary reward function can be more complex than the ones specified above that are formed from an identity function of the field measurement.
%such that it is related to, but not exactly, the field measurement.
For example, consider the problem of placing wind turbines in optimal locations to maximize the total power production.
Though the average wind speed in a region can be modeled by a GP, the power output is not a linear function of the steady-state wind speed.
In fact, power production requires a certain minimum speed known as the cut-in speed. 
After this threshold is met, power output increases and eventually plateaus. 
Assuming the cut-in speed is $1$, this effect can be modeled with a logarithmic reward function\footnote{In reality, the speed-power relationship is not exactly logarithmic, but this approximation suffices for the purpose of modeling.}:
$\rfn(z_{t+1}, \Shist{t+1}) = \rfn_1(z_{t+1})$ gives a value of $\log(z_{t+1})$ if $z_{t+1}>1$, and $0$ otherwise where $\ell_1=1$.
To the best of our knowledge, $h_{\sigma_{s_{t+1}|\Shist{t}}}(u)$ has no closed-form expression. In\if\myproof1 Appendix~\ref{aeg}, \fi\if\myproof0 \cite{AA16}, \fi  
we present other interesting reward functions like unit step function\cref{const} and Gaussian distribution that can be represented by $\rfn(z_{t+1}, \Shist{t+1})$ and used in real-world tasks.
%
%where  and $\mathbf{B}(\Shist{t+1}) \triangleq \Sigma_{s_{t+1}\Shist{t}}(\Sigma_{\Shist{t}\Shist{t}} + \sigma_{n}^2I)^{-1}$ is a column vector containing the weights for measurements when applying GP regression.
%
%In the following section, we will prove several properties of $V_t^*$ to aid our approximations.
%

%{\bf .}
%\label{propertyv}
Theorem~\ref{th:LipschitzAll} below reveals that $V^*_t(d_t)$  \eqref{eq:OptimalValFunDef} with Lipschitz continuous reward functions is Lipschitz continuous in $\Zhist{t}$ with  Lipschitz constant $L_{t}(\Shist{t})$ defined below:\vspace{-1mm} 
\begin{definition}\label{def:ValFunGenLip}
Let $L_{H}(\Shist{H}) \triangleq 0$. For $t=0,\ldots,H-1$, define %\vspace{-2.5mm} 
%Suppose $\Shist{t}$ contains $i$ locations. 
%Recall that $\mathbf{B}(\Shist{t+1}) = \Big(\Sigma_{s_{t+1}\Shist{t}}(\Sigma_{\Shist{t}\Shist{t}} + \sigma_{\text{noise}}^2I)^{-1}\Big)^{\text{T}}$.
%$L_{t}(\Shist{t})$ is recursively defined to be the following $i$ element (column) vector containing non-negative reals:
$
%\label{eq:GenLipExact}
% \begin{array}{l}
  L_{t}(\Shist{t}) \triangleq \max_{s_{t+1} \in \Adom(s_t)} \alpha(\Shist{t+1}) \left(\ell_1 + \ell_2(\sigma_{s_{t+1}|\Shist{t}})\right) + L_{t+1}(\Shist{t+1})\sqrt{1+\alpha(\Shist{t+1})^2}\ .
 \vspace{-1mm} 
%\end{array}
$
\end{definition}
%The following theorem bounds the effect that measurements $\Zhist{t}$ has on $V_t^*(d_t)$.
\begin{theorem}[Lipschitz Continuity of $V^*_t$]\label{th:LipschitzAll}
% Suppose $d_t=\langle \Shist{t}, \Zhist{t} \rangle$, $d_{t}'=\langle \Shist{t}, \Zhist{t}' \rangle$. The following inequality holds 
For $t=0,\ldots,H$, 
% \begin{equation}
$|V^*_t(d_t) - V^*_t(d_{t}')| \le L_{t}(\Shist{t}) \lVert\Zhist{t}-\Zhist{t}'\rVert\ .$\vspace{-0.5mm} 
% \label{whatthe}
% \end{equation}
\end{theorem}
Its proof uses Lemma~\ref{rewlc} and is in\if\myproof1 Appendix~\ref{akaka}. \fi\if\myproof0 \cite{AA16}. \fi
The result below is a direct consequence of Theorem~\ref{th:LipschitzAll} and will be used to theoretically guarantee the performance of our proposed nonmyopic adaptive $\epsilon$-optimal GPP policy in Section~\ref{eogpp}:\vspace{-0.5mm} 
\begin{corollary} \label{th:LipschitzSingle}
%  $V^*_{t+1}(d_{t+1})=V^*_{t+1}(\langle \Shist{t+1}, \Zhist{t} \oplus z_{t+1} \rangle)$ is Lipschitz continuous in $z_{t+1}$ with Lipschitz constant equal to the last element of $L_{t+1}(\Shist{t+1})$.
 % Specifically, if $\Shist{t}$ contains $i$ locations, 
%By definition, $d_{t}=\langle \Shist{t}, \Zhist{t-1} \oplus z_{t} \rangle$. Let $d_{t}'\triangleq\langle \Shist{t}, \Zhist{t-1} \oplus z'_{t} \rangle$. Then, for $t=0,\ldots,H$,
%$|V^*_{t}(d_{t}) - V^*_{t}(d_{t}') | \le L_{t}(\Shist{t})| z_{t} -z_{t}'|$.
For $t=0,\ldots,H$, $|V^*_{t}(\langle \Shist{t}, \Zhist{t-1} \oplus z_{t} \rangle) - V^*_{t}(\langle \Shist{t}, \Zhist{t-1} \oplus z'_{t} \rangle) | \le L_{t}(\Shist{t})| z_{t} -z_{t}'|$.
\vspace{-1.5mm} 
\end{corollary}
%Corollary~\ref{th:LipschitzSingle} . 
%
%For ease of exposition, we define $L_{t}(\Shist{t})_{-1}$ to be the last element $L_{t}(\Shist{t})$. 
%In addition we define $\xi(\Shist{t}) \triangleq \ell_1 + L_{t}(\Shist{t})_{-1}$.
%
\section{$\epsilon$-Optimal GPP ($\epsilon$-GPP)}%\vspace{-3.5mm} 
\label{eogpp}
\label{ch:SamplingStrategy}
The key idea of constructing our proposed nonmyopic adaptive $\epsilon$-GPP policy is to approximate the expectation terms in \eqref{eq:OptimalValFunDef} at every stage using a form of deterministic sampling, as illustrated in the figure below.
%Instead of performing Monte Carlo sampling to obtain equally weighted sample measurements and subsequently deriving a probabilistic policy loss bound like that of \cite{NghiaICML14}, our proposed alternative is to consider deterministic sampling by 
%
%Equidistant sample measurements are selected from within some user-specified region (except for the extremes corresponding to the limits of this region) that are weighted according to the areas under their respective equally spaced intervals of the Gaussian predictive distribution $p(z_{t+1}|d_t, s_{t+1})$ centered at them (except for the extremes weighted by their respective tails). Consequently, a deterministic policy loss bound can be derived.
Specifically, the measurement space of $p(z_{t+1}|d_t, s_{t+1})$ is first partitioned into $n\geq 2$ intervals $\zeta_0, \ldots, \zeta_{n-1}$ such that intervals $\zeta_1, \ldots, \zeta_{n-2}$ are equally spaced within the bounded gray region $[\mu_{s_{t+1}|d_t} - \sdk\sigma_{s_{t+1}|\Shist{t}}, \mu_{s_{t+1}|d_t} + \sdk\sigma_{s_{t+1}|\Shist{t}}]$ specified by a user-defined width parameter $\tau\geq 0$ while intervals $\zeta_0$ and $\zeta_{n-1}$ span the two  infinitely long red tails.
%(i.e., $\zeta_0\triangleq [-\infty, \mu_{s_{t+1}|d_t} - \sdk\sigma_{s_{t+1}|\Shist{t}}]$ and $\zeta_{n-1}\triangleq[\mu_{s_{t+1}|d_t} + \sdk\sigma_{s_{t+1}|\Shist{t}}, \infty]$).
Note that $\tau>0$ requires $n>2$ for the partition to be valid.
%Define a sampling region $[\mu_{s_{t+1}|d_t} - \sdk\sigma_{s_{t+1}|\Shist{t}}, \mu_{s_{t+1}|d_t} + \sdk\sigma_{s_{t+1}|\Shist{t}}]$ in the measurement space of $Z_{t+1}$ such that its width can be controlled by the parameter $\sdk$.
The $n$ sample measurements $z^{0} \dots z^{n-1}$ are then selected by setting 
$z^0$
%\triangleq\mu_{s_{t+1}|d_t} - \sdk\sigma_{s_{t+1}|\Shist{t}}$ 
as upper limit of red interval $\zeta_0$, $z^{n-1}$
%\triangleq\mu_{s_{t+1}|d_t} + \sdk\sigma_{s_{t+1}|\Shist{t}}$ 
as lower limit of red interval $\zeta_{n-1}$,
and $z^1, \ldots, z^{n-2}$ as centers of the respective gray intervals $\zeta_1, \ldots, \zeta_{n-2}$.
% (i.e., $z^{i}\triangleq z^{0} + \frac{i-0.5}{n-2}(z^{n-1} - z^{0})$ for $i=1,\ldots,n-2$).
%$$
% z^{i} \triangleq \hspace{-0.5mm}\left\{
% \begin{array}{ll}
%  \displaystyle\mu_{s_{t+1}|d_t} - \sdk\sigma_{s_{t+1}|\Shist{t}} &\text{if}\ i = 0,\vspace{1mm}\\
%\displaystyle  \mu_{s_{t+1}|d_t} + \sdk\sigma_{s_{t+1}|\Shist{t}}  &\text{if}\  i = n-1,\vspace{1mm}\\
% \displaystyle z^{0} + \frac{i-0.5}{n-2}(z^{n-1} - z^{0}) &\text{if}\ 0 < i < n-1.
% \end{array}\right.
%$$
Next, the weights $w^{0} \dots w^{n-1}$ for the corresponding sample measurements $z^0, \ldots, z^{n-1}$ are defined as the areas under their respective intervals $\zeta_0, \ldots, \zeta_{n-1}$ of the Gaussian predictive distribution $p(z_{t+1}|d_t, s_{t+1})$.
%, that is, $w^0 = w^{n-1}\triangleq\Phi(-\sdk)$ and $w^{i} \triangleq \Phi(\frac{2i\sdk}{n-2}-\sdk) - \Phi(\frac{2(i-1)\sdk}{n-2}-\sdk)$ for $i=1,\ldots,n-2$.
%$$
% w^{i} \hspace{-0.5mm}\triangleq\hspace{-1mm}\left\{\hspace{-1.5mm}
% \begin{array}{ll}
 % \displaystyle\Phi(-\sdk)  &\hspace{-3mm} \text{if}\ i = 0 \vee i= n\hspace{-0.5mm}-\hspace{-0.5mm}1,\vspace{1mm}\\
%  \displaystyle \Phi\hspace{-1mm}\left(\frac{2i\sdk}{n\hspace{-0.5mm}-\hspace{-0.5mm}2}\hspace{-0.5mm}-\hspace{-0.5mm}\sdk\hspace{-0.5mm}\right)\hspace{-1mm} - \hspace{-0.5mm}\Phi\hspace{-1mm}\left(\frac{2(i\hspace{-0.5mm}-\hspace{-0.5mm}1)\sdk}{n\hspace{-0.5mm}-\hspace{-0.5mm}2}\hspace{-0.5mm}-\hspace{-0.5mm}\sdk\hspace{-0.5mm}\right)  &\hspace{-3mm}\text{if}\  0 \hspace{-0.5mm}< \hspace{-0.5mm}i \hspace{-0.5mm}< \hspace{-0.5mm}n\hspace{-0.5mm}-\hspace{-0.5mm}1.
% \end{array}\right.
%$$
So, $\sum^{n-1}_{i=0}w^i = 1$.
An example of such a partition is given in\if\myproof1 Appendix~\ref{dsampeg}. \fi\if\myproof0 \cite{AA16}. \fi
The selected sample measurements and their corresponding weights can be exploited for approximating $V^*_t$ with Lipschitz continuous reward functions \eqref{eq:OptimalValFunDef} 
%(Section~\ref{ch:rewardfunctions}) 
using the following $H$-stage Bellman equations:\vspace{-1.5mm}
\begin{equation} 
\hspace{-1.8mm}
 \begin{array}{rl}
  V_t^{\epsilon}(d_t) \triangleq\hspace{-2.4mm} &  \max_{s_{t+1} \in \Adom(s_{t})} Q_t^{\epsilon}(d_t, s_{t+1})\vspace{0.5mm}\\
  Q_t^{\epsilon}(d_t, s_{t+1}) \triangleq\hspace{-2.4mm} &  g_{\sigma_{s_{t+1}|\Shist{t}}}\hspace{-1mm}\left(\mu_{s_{t+1}|d_{t}}\right) +\rfn_3(\Shist{t+1})\ + \\
  &\displaystyle\sum^{n-1}_{i=0} w^i \hspace{-1mm}\left(R_1(z^{i}) + V_{t+1}^{\epsilon}(\langle \Shist{t+1}, \Zhist{t}\oplus z^{i}\rangle)\right)\vspace{-3mm}\hspace{-5.3mm}
  \end{array}
\label{eq:EpsilonValFunDef}
\end{equation}
for stages $t = 0,\ldots,H-1$ where $V_H^{\epsilon}(d_H)\triangleq 0$. 
The resulting induced $\epsilon$-GPP policy $\pi^\epsilon$  jointly and naturally optimizes the exploration-exploitation trade-off in a similar manner as that of the GPP policy $\pi^*$, as explained in Section~\ref{proble}.
It is interesting to note that setting $\sdk=0$ yields $z^{0}=\ldots = z^{n-1}=\mu_{s_{t+1}| d_t}$, which is equivalent to selecting a single sample measurement of $\mu_{s_{t+1}| d_t}$ with corresponding weight of $1$.
This is identical to the special case of maximum likelihood observations during planning which is the extreme assumption used by nonmyopic UCB \cite{RamosUAI14} for sampling to gain time efficiency.\vspace{0.5mm}
\begin{figure}
%\hspace{-3mm}
%{r}{7.4cm}\vspace{-5mm}
\resizebox{8.4cm}{!}{%
%\hspace{-1mm}
    \begin{tikzpicture}
      %\draw[->] (0,-0.5) -- (0,4.2) node[above] {$y$};
      \draw[scale=1,domain=-5:5,smooth,variable=\x,black, line width=3.0pt] plot ({\x},{4*exp(-\x*\x*0.5*0.2)}) (4,1.7) node[above, black] {$p(z_{t+1}|d_t, s_{t+1})$};
      
      \fill[fill=red] (4,0) -- plot [domain=4:5] ({\x},{4*exp(-\x*\x*0.5*0.2)}) -- (5,0) -- cycle;
      
      \fill[fill=red] (-5,0) -- plot [domain=-5:-4] ({\x},{4*exp(-\x*\x*0.5*0.2)}) -- (-3,0) -- cycle;
            
      \fill[fill=gray] (-4,0) -- plot [domain=-4:4] ({\x},{4*exp(-\x*\x*0.5*0.2)}) -- (4,0) -- cycle;      
      
      \foreach \x [evaluate = \x as \xpp using 4*exp(-\x*\x*0.5*0.2)] in {-4,-2.9,-1.65,-0.55, 0.55, 1.65,2.9,4} 
      \draw[dashed, black, line width=1pt] (\x, 0) -- (\x,\xpp );

\draw (3.15,2.54) node[black,above] {$\begin{array}{r}
z^{0} \triangleq \mu_{s_{t+1}|d_t}\hspace{-1mm}-\hspace{-0.5mm}\tau\sigma_{s_{t+1}|\mathbf{s}_{t}};\vspace{0mm}\\
z^{n-1} \triangleq \mu_{s_{t+1}|d_t}\hspace{-1mm}+\hspace{-0.5mm}\tau\sigma_{s_{t+1}|\mathbf{s}_{t}};\vspace{0mm}\\
z^{i}\hspace{-1mm} \triangleq\hspace{-0.5mm} z^{0}\hspace{-1mm} + \hspace{-1mm}\frac{i-0.5}{n-2}(z^{n-1}\hspace{-1mm} -\hspace{-1mm} z^{0})\\
\text{for}\ i = 1,\ldots,n-2.
\end{array}$};
\draw (-2.7,2.9) node[black,above] {$
\begin{array}{l}
w^{i}\hspace{-1mm} \triangleq\hspace{-0.5mm} \Phi(\frac{2i\tau}{n-2}\hspace{-1mm}-\hspace{-0.5mm}\tau) \hspace{-1mm}-\hspace{-1mm} \Phi(\frac{2(i-1)\tau}{n-2}\hspace{-1mm}-\hspace{-0.5mm}\tau)\\
\text{for}\ i = 1,\ldots,n-2;\vspace{0.5mm}\\
w^0 = w^{n-1}\triangleq\Phi(-\tau).
\end{array}
$};
  \draw (-4,0) -- ( -4,-0.1) node[below] {$z^{0}$} (-4.5,-0.03) node[above] {$w^0$};      
      \draw (-3.45,0) -- ( -3.45,-0.1) node[below] {$z^{1}$} ( -3.45,0.35) node[above] {$w^{1}$};      
       \draw (-2.275,-0.99) node[below] {$\ldots$} (-2.275,-0.56) node[gray, below] {$\ldots$} (-2.275,-0.35) node[below] {$\ldots$} (-2.275,1) node[above] {$\ldots$};
      \draw (-1.1,0) -- ( -1.1,-0.1) node[below] {$z^{i\text{-}1}$}  ( -1.1,1.6) node[above] {$w^{i\text{-}1}$};
      \draw ( 0,0) -- ( 0,-0.1) node[below] {$z^{i}$} ( 0, 1.9) node[above] {$w^{i}$};
      \draw ( 1.1,0) -- ( 1.1,-0.1) node[below] {$z^{i\text{+}1}$} ( 1.1,1.6) node[above] {$w^{i\text{+}1}$};
      \draw (4,0) -- ( 4,-0.1) node[below] {$z^{n\text{-}1}$} (4.5,-0.03) node[above] {$w^{n\text{-}1}$};      
      \draw (3.45,0) -- ( 3.45,-0.1) node[below] {$z^{n\text{-}2}$} ( 3.45,0.35) node[above] {$w^{n\text{-}2}$};      
       \draw (2.275,-0.99) node[below] {$\ldots$} (2.275,-0.56) node[gray, below] {$\ldots$} (2.275,-0.35) node[below] {$\ldots$} (2.275,1) node[above] {$\ldots$};         
    \draw (4.85,0) node[black,below] {$z_{t+1}$};
 
	% labels for zetas
	     \draw[<->, gray, line width =1.5] (-4, -0.7) -- node[black,below] {$\zeta_{1}$} ++ (1.1,0);
	     \draw[<->, gray, line width =1.5] (-1.65, -0.7) -- node[black,below] {$\zeta_{i\text{-}1}$} ++ (1.1,0);
	     \draw[<->, gray, line width =1.5] (-0.55, -0.7) -- node[black,below] {$\zeta_{i}$} ++ (1.1,0);
	     \draw[<->, gray, line width =1.5] (0.55, -0.7) -- node[black,below] {$\zeta_{i\text{+}1}$} ++ (1.1,0);	
	     	     \draw[<->, gray, line width =1.5] (2.9, -0.7) -- node[black,below] {$\zeta_{n\text{-}2}$} ++ (1.1,0);     
	\draw[->, red, line width =1.5] (-5, -0.7) -- node[black,below] {$\zeta_0$} ++ (1,0);
	
	\draw[<-, red, line width =1.5] (4.0, -0.7) -- node[black,below] {$\zeta_{n\text{-}1}$} ++ (1,0);
	
        \draw[->, black, line width =1] (-5,0) -- (5.13,0);
   
    \end{tikzpicture}
    }%
    
    \vspace{-6mm}
%\vspace{-5mm}\includegraphics[height=5cm]{dsamp}\vspace{-3.3mm}
\label{fig:dsamp}
\end{figure}

\noindent
{\bf Performance Guarantee.}
The difficulty in theoretically guaranteeing the performance of our $\epsilon$-GPP policy $\pi^{\epsilon}$ (i.e., relative to that of GPP policy $\pi^*$) lies in analyzing how 
the values of the width parameter $\tau$ and deterministic sampling size $n$ can be chosen to satisfy the user-specified loss bound $\epsilon$, as discussed below. 
The first step is to prove that $V_t^{\epsilon}$ in \eqref{eq:EpsilonValFunDef} approximates $V_t^{*}$ in \eqref{eq:OptimalValFunDef} closely for some chosen $\tau$ and $n$ values, which relies on the Lipschitz continuity of $V_t^{*}$ in Corollary~\ref{th:LipschitzSingle}.
%determining how its ?- Bayes optimality can be guaranteed by choosing appropriate values of .
%the truncated sampling parameters S and ? (Section 3.2). To achieve this, we have to formally under- stand how S and ? can be specified and varied in terms of the user-defined loss bound ?, budget of N sampling lo- cations, domain size |X | of the phenomenon, and proper- ties/parameters characterizing the spatial correlation struc- ture of the phenomenon (Section 2), as detailed below.
%
%$\lambda \leftarrow \frac{\epsilon}{H(H+1)}$
%$\epsilon$ is a user-defined value describing maximum allowable policy loss.
%The actual construction of the tree begins with the first call to \textsc{Compute$U$} in line~\ref{algline:dsStartTreeExpand}.
%Prior to that, several preprocessing steps are required.
%
%In the function \textsc{GetSamples} (line~\ref{algline:dsSelectPartitions} in Algorithm~\ref{alg:ds}), we wish to select $n$ measurements, $z^{0} \dots z^{n-1}$ and.
%We know that the robot is at location $s_{t+1}$, and has past locations/measurements $d_t$.
%Define $\kappa(\sdk)=\frac{1}{\sqrt{2\pi}} \exp \big(-\frac{\sdk^2}{2}\big) - \sdk \Phi(-\sdk)$. %and  $\eta(n,\sdk)=\frac{\sdk}{n}\big(\frac{1}{2}-\Phi(-\sdk)\big)$.
%Define $\xi(\Shist{t}) \triangleq \ell_1 + L_{t}(\Shist{t})$.
%Observe that computing $\xi(\Shist{t})$ for all \textit{reachable} $\Shist{t}, 0 \le t \le H$ can be done by directly implementing the recursive formulas in \ref{def:ValFunGenLip} and the definition of $\xi$.
Define $\Lambda(n, \sdk)$ to be equal to the value of $\sqrt{2/\pi}$ if $n\geq2\wedge \sdk=0$, and value of $\kappa(\sdk)+\eta(n,\sdk)$ if $n>2\wedge \sdk> 0$ where
$\kappa(\sdk)\triangleq\sqrt{{2/\pi}} \exp(-0.5{\sdk^2})- 2\sdk\Phi(-\sdk)$, $\eta(n,\sdk)\triangleq{2\sdk}(0.5-\Phi(-\sdk))/({n-2})$, and $\Phi$ is a standard normal CDF.\vspace{-0mm}
%
%\begin{equation*}
% \begin{array}{c}
% \Lambda(n, \sdk)\hspace{-0.5mm}\triangleq \hspace{-0.5mm}\left\{
% \begin{array}{ll}
%\displaystyle \sqrt{2/\pi} & \hspace{-0mm}\text{if}\ n\geq2\wedge \sdk=0,\vspace{1mm}\\ %2 \kappa(0) 
%\displaystyle\kappa(\sdk)+\eta(n,\sdk) &\hspace{-0mm} \text{if}\ n>2\wedge \sdk> 0;%\vspace{1mm}\\
%   \displaystyle  \hspace{-1mm}+\infty & \hspace{-1mm}\text{if}\ n=2\wedge \sdk > 0.
 %\end{array}\right.\vspace{1mm}\\
%\displaystyle\kappa(\sdk)\triangleq\sqrt{\frac{2}{\pi}} \exp\hspace{-1mm}\left(-\frac{\sdk^2}{2}\right)- 2\sdk\Phi(-\sdk)\ ,\vspace{1mm}\\ 
%\displaystyle\eta(n,\sdk)\triangleq\frac{2\sdk}{n-2}\left(\frac{1}{2}-\Phi(-\sdk) \right).
%\end{array}
 %\end{equation*}
%$m$ and $\sdk$ are selected such that:
%
%We set $m=n+2$.
%Corollary~\ref{th:LipschitzSingle} gives an upper bound of how much effect $\Delta z_{t+1}$ has on our optimal value function. Intuitively, this allows us to bound the single stage error due to the rectangular approximations made.In particular, the following theorem 2 provides a bound for the accuracy of our approximations, $V_t^{\epsilon}$.
%
\begin{theorem} \label{th:MultistageError} Suppose that $\lambda > 0$ is given.
For all $d_t$ and $t = 0,\ldots, H$, if\vspace{-1mm}
\begin{equation} \label{eq:PartitionIneq}
\lambda \ge \Lambda(n, \sdk) \sigma_{s_{t+1}|\Shist{t}}(\ell_1 + L_{t+1}(\Shist{t+1}))\vspace{-1mm}
\end{equation}
for all $s_{t+1}\in\Adom(s_t)$, then
%\begin{equation}\label{whack}
$|V_t^{\epsilon}(d_t) - V_t^{*}(d_t)| \le \lambda (H-t)\ .$\vspace{-0mm}
%\end{equation}
\end{theorem}
Its proof uses Corollary~\ref{th:LipschitzSingle} and is given in\if\myproof1 Appendix~\ref{jordan}. \fi\if\myproof0 \cite{AA16}. \fi

\noindent
\emph{Remark} $1$. From Theorem~\ref{th:MultistageError}, a tighter bound on the error $|V_t^{\epsilon}(d_t) - V_t^{*}(d_t)|$ can be achieved by decreasing the sampling budget of $H$ locations\footnote{\label{change}This changes $\epsilon$-GPP by reducing its planning horizon though.} and increasing the deterministic sampling size $n$; increasing $n$ reduces $\eta(n,\sdk)$ and hence $\Lambda(n,\sdk)$, which allows $\lambda$ to be reduced as well. The width parameter $\tau$ has a mixed effect on this error bound: Note that $\kappa(\sdk)$ ($\eta(n,\sdk)$) is proportional to some upper bound on the error incurred by the extreme sample measurements $z^0$ and $z^{n-1}$ ($z^1,\ldots,z^{n-2}$), as shown in\if\myproof1 Appendix~\ref{jordan}. \fi\if\myproof0 \cite{AA16}. \fi 
Increasing $\tau$ reduces $\kappa(\sdk)$ but unfortunately raises $\eta(n,\sdk)$.
So, in order to reduce $\Lambda(n,\sdk)$ further by increasing $\tau$, it has to be complemented by raising $n$ fast enough to keep $\eta(n,\sdk)$ from increasing.
This allows $\lambda$ to be reduced further as well.

\noindent
\emph{Remark} $2$. 
A feasible choice of $\tau$ and $n$ satisfying \eqref{eq:PartitionIneq} can be expressed  analytically in terms of the given $\lambda$ and hence computed prior to planning, as shown in\if\myproof1 Appendix~\ref{yayat}. \fi\if\myproof0 \cite{AA16}. \fi

\noindent
\emph{Remark} $3$. $\sigma_{s_{t+1}|\Shist{t}}$ and $L_{t+1}(\Shist{t+1})$ for all $\Shist{t+1}$ and $t=0,\ldots,H-1$ can be computed prior to planning as they depend on $\Shist{0}$ and all reachable locations from $s_0$ but not on their measurements.
%Precompute $\sigma(\Shist{t}), L_t(\Shist{t})$ for all reachable $\Shist{t}$ from $\Shist{0}$
%
%In particular, we iterate over all reachable \textit{location} histories and compute several properties of these histories. Note that the precomputed elements (line~\ref{algline:dsPreprocess}) only depend on the location history, and not measurements. During the actual tree expansion, we make extensive use of $\lambda$ which is an auxillary variable governing the maximum stage-wise error tolerance. A small $\lambda$ indicates a larger number of sample measurements to be made at each stage.
%
%We can see that our choice of $n$ directly affects the branching factor of our algorithm.
%Thus, it is advantageous to select a small $n$ that satisfies the inequality in \eqref{eq:PartitionIneq}.
%$\sdk$ can be chosen arbitarily as long as \eqref{eq:PartitionIneq} is satisfied. We are not aware of existing methods to minimize $n$ subject to \eqref{eq:PartitionIneq}.

Using Theorem~\ref{th:MultistageError}, the next step is to bound the performance loss of our $\epsilon$-GPP policy $\pi^{\epsilon}$ relative to that of GPP policy $\pi^*$, that is, policy $\pi^{\epsilon}$ is $\epsilon$-optimal:\vspace{-0mm}
%
%Theorem~\ref{th:SingleStageError} bounds the \textit{single stage} error due to finite sampling. When running this algorithm over $H$ time steps, these errors add up additively. This leads us to our main result.
\begin{theorem}\label{th:PolicyLoss} 
%Let $\epsilon\triangleq\lambda H(H+1)$ such that $\lambda$ satisfies \eqref{eq:PartitionIneq}. Then,
% Denote the policy computed by Algorithm~\ref{alg:ds} as $\pi^A$, using the sampling strategy described in Section~\ref{ch:SamplingStrategy}. Its coressponding value function, denoted by $V_{t}^A$ is as defined in \eqref{eq:ValFunDef}. The following bound for its policy loss holds:
 %\begin{equation*}
Given the user-specified loss bound $\epsilon>0$, $V_0^*(d_0)-V_0^{\pi^{\epsilon}}(d_0) \le \epsilon$ by 
%setting and 
substituting $\lambda=\epsilon/( H(H+1))$ into the choice of $\tau$ and $n$ stated in Remark $2$ above.\vspace{-0mm}
% \end{equation*}
\end{theorem}
Its proof is in\if\myproof1 Appendix~\ref{hender}. \fi\if\myproof0 \cite{AA16}. \fi
%\emph{Remark}. 
It can be observed from Theorem~\ref{th:PolicyLoss} that a tighter bound $\epsilon$ on the error $V_0^*(d_0)-V_0^{\pi^{\epsilon}}(d_0)$ can be achieved by decreasing the sampling budget of $H$ locations\cref{change} and increasing the deterministic sampling size $n$.
The effect of width parameter $\tau$ on this error bound $\epsilon$ is the same as that on the error bound of $|V_t^{\epsilon}(d_t) - V_t^{*}(d_t)|$, as explained in Remark $1$ above.
\vspace{1mm}
%\emph{Remark} $2$. A feasible choice of $\tau$ and $n$ can be computed analytically in terms of the user-specified loss bound $\epsilon$ prior to planning, as shown in Appendix~\ref{yayat}.

%The feasible choice of $\tau$ and $n$ reported in Appendix~\ref{yayat} that satisfy \eqref{eq:PartitionIneq} and can be computed analytically prior to planning can also be used here to meet 

%larger epsilon needed when larger posterior variance, smaller n, smaller/larger tau
%smaller epsilon needed when smaller posterior variance, smaller H, larger n, smaller/larger tau
%-tau/(n-2)erf(-tau/sqrt(2)) -tau erf(-tau/sqrt(2))) -tau
%-tau erf(-tau/sqrt(2))) (1/(n-2)+1)
%\phi from 0 to 1/2
%
%\begin{theorem} [\textbf{Complexity}]\label{th:FullTreeRuntime}
% Suppose $d_0$ has $i$ locations/measurements. 
% Let $|\mathcal{B}|$ be the maximum number of sample measurements taken at any measurement node.
% Similarly, denote $|\Adom|$ to be the the maximum number of actions at any given location.
% Under these conditions, Algorithm~\ref{alg:ds} requires $\mathcal{O}\big((i+H)^3|\Adom|^H+|\Adom \mathcal{B}|^H(i+H)\big)$ time and $\mathcal{O}\big((i+H)|\Adom|^H + (i+H)^2\big)$ space to output an action for a single time step.
%\end{theorem}
%\ref{th:PolicyLoss} provides a guarantee that $\pi^A$ will be $\epsilon$-optimal.
%

\noindent
{\bf Anytime $\epsilon$-GPP.} Unlike GPP policy $\pi^*$, our $\epsilon$-GPP policy $\pi^{\epsilon}$ can be derived exactly since its incurred time is independent of
the size of the uncountable set of candidate measurements.
However, expanding the entire search tree of $\epsilon$-GPP \eqref{eq:EpsilonValFunDef} incurs time containing a $\mathcal{O}(n^H)$ term and is not always necessary to achieve $\epsilon$-optimality in practice.
% since it has to compute the values $V_t^{\epsilon}(d_t)$  for all $(S|\mathcal{X}|)^N$ possible states $(n,z_\mathcal{D})$.
%To eliminate this expensive $\mathcal{O}(n^H)$ term
To mitigate this computational difficulty\footnote{\label{sucks}The value of $n$ is a bigger computational issue than that of $H$ when $\epsilon$ is small and in online planning.}, we propose an anytime variant of $\epsilon$-GPP that can produce a good policy fast and improve its approximation quality over time, as briefly discussed here and detailed with the pseudocode in\if\myproof1 Appendix~\ref{aegpp}. \fi\if\myproof0 \cite{AA16}. \fi
% (Algorithm~\ref{alg:ds-anytime})

%In practice, the value of $n$ is extremely sensitive to $\sigma_{s_{t+1}|\Shist{t}}(\ell_1 + L_{t+1}(\Shist{t+1}))$, thus making the version of $\epsilon$-GPP described in Section~\ref{eogpp} unsuitable for real-time applications. Furthermore, $\epsilon$-optimality may be attained before the search tree is fully expanded. These considerations drive us to apply branch-and-bound techniques to reduce computational efforts. In this section, we  briefly describe how a reasonable anytime variant of $\epsilon$-GPP can be derived.

The key intuition is to expand the sub-trees rooted at ``promising'' nodes with the highest weighted uncertainty of their corresponding values $V^{*}_t(d_t)$ so as to improve their estimates.
To represent such uncertainty at each encountered node, 
%on  ``promising'' sample measurements that are likely to improve our value estimates considerably.
 upper \& lower heuristic bounds (respectively, $\obar{V}_{t}^{*}(d_t)$ and $\ubar{V}_{t}^{*}(d_t)$) are maintained, like in \cite{Simmons06}.
%
%Instead of explicitly maintaining $V_{t}^{\epsilon}(d_t)$, upper and lower bounds for $V_{t}^{*}(d_t)$ and $Q_{t}^{*}(d_t, s_{t+1})$ are maintained for each node encountered in the search tree. These are denoted by $\obar{V}_{t}^{*}$, $\ubar{V}_{t}^{*}$, $\obar{Q}_{t}^{*}$, $\ubar{Q}_{t}^{*}$ and generalize the two-sided bounds proposed by Theorem~\ref{th:MultistageError}.
A partial construction of the entire tree is maintained
%Instead of expanding the tree depth-first,  at any point.
and expanded incrementally in each iteration of anytime $\epsilon$-GPP that incurs linear time in $n$ and 
%The tree is built incrementally in an iterative manner as resources permit.
%Each iteration 
comprises $3$ steps:
%by repeated calls to \textsc{BuildTree}

\noindent
\textbf{Node selection.} Traverse down the partially constructed tree by repeatedly selecting nodes with largest difference between their upper and lower bounds (i.e., uncertainty) 
%$\obar{V}_{t}^{*}(d_t)- \ubar{V}_{t}^{*}(d_t)$ 
discounted by  weight $w^{i^*}$ of its preceding sample measurement $z^{i^*}$ 
until an unexpanded node, denoted by  $d_t$, is reached.
%
%Explore the current partially constructed tree to find and select an unexpanded node . Let this node be .
%using the methods \textsc{SelectAction} and \textsc{SelectMeasurement}.
%This is done recursively in \textsc{BuildTree}. This node is denoted by .

\noindent
\textbf{Expand tree.} Construct a ``minimal'' sub-tree rooted at node $d_t$ by sampling all possible next locations and only their median sample measurements $z^{\bar{i}}$ recursively up to full height $H\hspace{-0.7mm}$.
%A minimal version of the original sub-tree rooted at node $d_t$ is built by calling the function \textsc{ExtendTree}.
%When constructing this tree, all possible actions are sampled. However, for each sampled action, only one candidate measurement ($z^{+}$) is expanded (line~\ref{algline:DefaultMeasurement}). This minimal tree is constructed recursively up to the full height of $H$.

\noindent
\textbf{Backpropagation.} Backpropagate bounds from the leaves of the newly constructed sub-tree to node $d_t$, during which the refined bounds of expanded nodes are used to inform the bounds of unexpanded siblings by exploiting the Lipschitz continuity of $V^*_t$ (Corollary~\ref{th:LipschitzSingle}), as explained in\if\myproof1 Appendix~\ref{aegpp}. \fi\if\myproof0 \cite{AA16}. \fi
%The bounds for unexpanded nodes are derived from the expanded sibling nodes via the function \textsc{RefineBounds}, which updates the upper and lower bounds of all siblings of a node whose bounds have been updated.
Backpropagate bounds to the root of the partially constructed tree in a similar manner.
%, calling \textsc{RefineBounds} as required.
%
%The value of unexpanded nodes is approximated to be the value of its closest sibling.
%
%Evaluating $Q_t^{\epsilon}$ \eqref{eq:EpsilonValFunDef} requires the computation of $V_{t+1}^{\epsilon}(\langle \Shist{t+1}, \Zhist{t}\oplus z^{i}\rangle)$ for all $z^{i}$. By Corollary~\ref{th:LipschitzSingle}, we know that $|V_{t+1}^{*}(\langle \Shist{t+1}, \Zhist{t}\oplus z^{i}\rangle) - V_{t+1}^{*}(\langle \Shist{t+1}, \Zhist{t}\oplus z^{j}\rangle)| \le L_{t+1}(\Shist{t+1})|z^{i}-z^{j}|$, which suggests that $V_{t+1}^{\epsilon}(\langle \Shist{t+1}, \Zhist{t}\oplus z^{i}\rangle)$ may be used to approximate $V_{t+1}^{\epsilon}(\langle \Shist{t+1}, \Zhist{t}\oplus z^{j}\rangle)$.
%The function \textsc{RefineBounds} updates the bounds of all sibling nodes just after $\obar{V}_{t+1}^{*}(d^{+})$ and $\ubar{V}_{t+1}^{*}(d^{+})$ are updated.
%The update done within the loop inside of \textsc{RefineBounds}, where $b$ represents the maximum possible difference in true values (Corollary~\ref{th:LipschitzSingle}).
%
%We briefly discuss a few important considerations.
%In our implementation, \textsc{SelectAction} selects the action with the highest upper bound. \textsc{SelectMeasurement} chooses the measurement with the highest difference in upper and lower bounds, weighted by $w^{i}$.

%SUM^{H-1}_{t=1} A^t(nt^2+ t^3) = H A^H(nH^2 + H^3) reduce to linear in n
By using the lower heuristic bound to produce our anytime $\epsilon$-GPP policy, its performance loss relative to that of GPP policy $\pi^*$ can be bounded, as proven in\if\myproof1 Appendix~\ref{aegpp}. \fi\if\myproof0 \cite{AA16}. \fi\vspace{-2.5mm}
%
%To select the best immediate action, the tree is traversed from its root $d_0$ in accordance to \eqref{eq:EpsilonValFunDef}.
%
% There are several ways to select the immediate action at the root. 
% It can be seen that if the branch and bound algorithm is run to completion, 
% the bounds computed will be at least as tight as in that described in Theorem~\ref{th:MultistageError}.
% Thus, setting our desired action to $\argmax_{s_{1} \in \Adom(s_0)}\big(\obar{Q_{0}^{*}}(d_0, s_{1}) + \ubar{Q_{0}^{*}}(d_0, s_{1})\big)/2$ (Line~\ref{algline:MCTSrealaction})
% ensures that Theorem~\ref{th:MultistageError} holds. Consequently, Theorem~\ref{th:PolicyLoss} also holds.
% Hence, running the branch and bound algorithm to completion is guaranteed to $\epsilon$-optimal.
% < Note that in experiments, we used a slightly different action selection method, which involves tree traversal. This is to ensure that the exact same action as full-tree search is returned >
%
\section{Experiments and Discussion}\vspace{-0.5mm}
\label{expt}
This section empirically evaluates the online planning performance and time efficiency of our $\epsilon$-GPP policy $\pi^{\epsilon}$ and its anytime variant   
under limited sampling budget in an energy harvesting task on a simulated wind speed field and in BO on simulated plankton density (chl-a) field and real-world log-potassium (lg-K) concentration (mg~l$^{-1}$) field\if\myproof1 (Appendix~\ref{treesize2}) \fi\if\myproof0 \cite{AA16} \fi 
of Broom's Barn farm \cite{Webster01}.
Each simulated (real-world lg-K) field is spatially distributed over a $0.95$~km by $0.95$~km ($520$~m by $440$~m) region discretized into a $20\times 20$ ($14\times 12$) grid of sampling locations. These fields are assumed to be realizations of GPs. The wind speed (chl-a) field is simulated using hyperparameters $\mu_s=0$,\footnote{Its actual prior mean is not zero; we have applied zero-mean GP to $Y_s-\mu_s$ for simplicity.}  $l_1=l_2=0.2236$ ($0.2$)~km, $\sigma_{n}^2=10^{-5}$, and $\sigma_{y}^2=1$.
%$l_1^2=l_2^2=0.05$ ($0.04$)~km$^2$, $\sigma_{n}^2=10^{-5}$, and $\sigma_{y}^2=1$
The hyperparameters $\mu_s=3.26$, $l_1=42.8$~m, $l_2=103.6$~m, $\sigma^2_{n}=0.0222$, and $\sigma^2_{y}=0.057$ of lg-K field are learned using maximum likelihood estimation \cite{Rasmussen06}.
The robot's initial starting location is near to the center of each simulated field and randomly selected for lg-K field.
It can move to any of its $4$ adjacent grid locations at each time step and is tasked to maximize its total rewards over $20$ time steps (i.e., sampling budget of $20$ locations).
%The hyper value for the log-k field are: 
%3.2624 2.5898 1.0693 0.2388 0.1490
%in the format of constant prior mean, characteristic lengthscale x 2, signal std dev, noise std dev

In BO, the performances of our $\epsilon$-GPP policy $\pi^{\epsilon}$ and its anytime variant are compared with that of state-of-the-art \emph{nonmyopic UCB} \cite{RamosUAI14}  and \emph{greedy PI, EI, UCB} \cite{Brochu10,Srinivas10}.
%Lizotte07,
Three performance metrics are used: 
(a) Total rewards achieved over the evolved time steps (i.e., higher total rewards imply less total regret in BO (Section~\ref{gppfram})), (b) maximum reward achieved during  experiment, and (c) search tree size in terms of no. of nodes (i.e., larger tree size implies higher incurred time). 
%the size of the tree expanded, the latter of which is measured by the number of action nodes expanded during search and correlates with computational costs.
%We demonstrate that superior performance may be obtained either by increasing the search horizon $H$ or by building a wider tree (reducing $\epsilon$). 
%Some computational trade-offs associated with this gain in performance are also examined.
%We present results in synthetic spatially correlated environment fields and a real-world temperature field.
All experiments are run on a Linux machine with Intel Core i$5$ at $1.7$~GHz.\vspace{0.5mm}
%
%\emph{Remark}.
%In Theorem~\ref{th:FullTreeRuntime}, we showed that the time complexity of the Full-Tree search is $\mathcal{O}\big((i+H)^3|\Adom|^H+|\Adom \mathcal{B}|^H(i+H)\big)$. In our experiments, $(i+H)^3|\Adom|^H$ is generally much smaller than $|\Adom \mathcal{B}|^H(i+H)$, since the robot begins with no measurement and $i$ is small. Thus, $(i+H)$ is near constant and $|\Adom \mathcal{B}|^H$ is the term which dominates. This coressponds to the size of the search tree. The assumption that $(i+H)^3$ is near constant is not justified if we consider longer experiments.
%

%{\bf Simulated Spatial Field.} 
%
%The domain $\mathcal{S}$ of reachable locations is given as a finite $20 \times 20$ grid.
%Each side of the grid is taken to be of length $1.0$ and
%the robot's location is described by its Cartesian coordinates $(x, y)$.
%The robot is permitted to move to any of its $4$ adjacent grid locations on the condition that it does not exit the grid. 
%Revisiting previously explored locations is legal.
%Initially, the robot is in the center of the map with coordinates $(0.5, 0.5)$ where it has already accumulated a single measurement.
%For each experiment, $30$ independent realizations of a GP field were generated using a given set of hyperparameters.
%The robot is tasked to maximize its total reward over a period of $20$ time steps.
%, given it is aware of the hyperparameters used to generate $Y$.
%We will empirically discuss the effects of varying the search horizon $H$ and tolerance $\epsilon$ on robots with logarithmic and linear reward models.
%

\noindent
{\bf Energy Harvesting Task on Simulated Wind Speed Field.}
%,Wettergreen05
A robotic rover equipped with a wind turbine is tasked to harvest energy/power from the wind while exploring a polar region \cite{Chen14}. It is driven by the logarithmic reward function described under `General tasks/problems' in Section~\ref{ch:rewardfunctions}. 
%Its functional form is given by 
%\begin{equation*}
%\rfn(z_{t+1}, \Shist{t+1}) = \rfn_1(z_{t+1}) =
%\begin{cases}
%\log(z_{t+1}) &\text{if}\ z_{t+1}>1,\\
%0 &\text{otherwise}.
%\end{cases} 
%\end{equation*}
%As discussed previously, $\rfn_1(z_{t+1})$ is globally Lipschitz continuous with $\ell_1=1$.
%, satisfying the criteria laid out in Section~\ref{GPSSSS}.
%The length-scale used in this experiment was $l_1^2=l_2^2=0.05$.
%The signal and noise variances are $\sigma_{n}^2=10^{-5}$ and $\sigma_{y}^2=1.0$.
%\begin{table}
%\centering
\begin{table}
{\setlength{\tabcolsep}{0em}
\begin{tabular}{p{\textwidth}}
\floatbox[{\capbeside\thisfloatsetup{capbesideposition={left,top},capbesidewidth=2.66cm}}]{figure}[\FBwidth]
{\caption{Graphs of total rewards and tree size of $\epsilon$-GPP policies with (a-b) online planning horizon $H'=4$ and varying $\epsilon$ and (c-d) varying $H'=1,2,3,4$ (respectively, $\epsilon=0.002, 0.06, 0.8, 5$)\vspace{-1mm}
}\label{fig:gpp_synthetic_log}}
{\hspace{-9mm}
\begin{tabular}{cc}
\hspace{-0mm}\includegraphics[width=2.77cm]{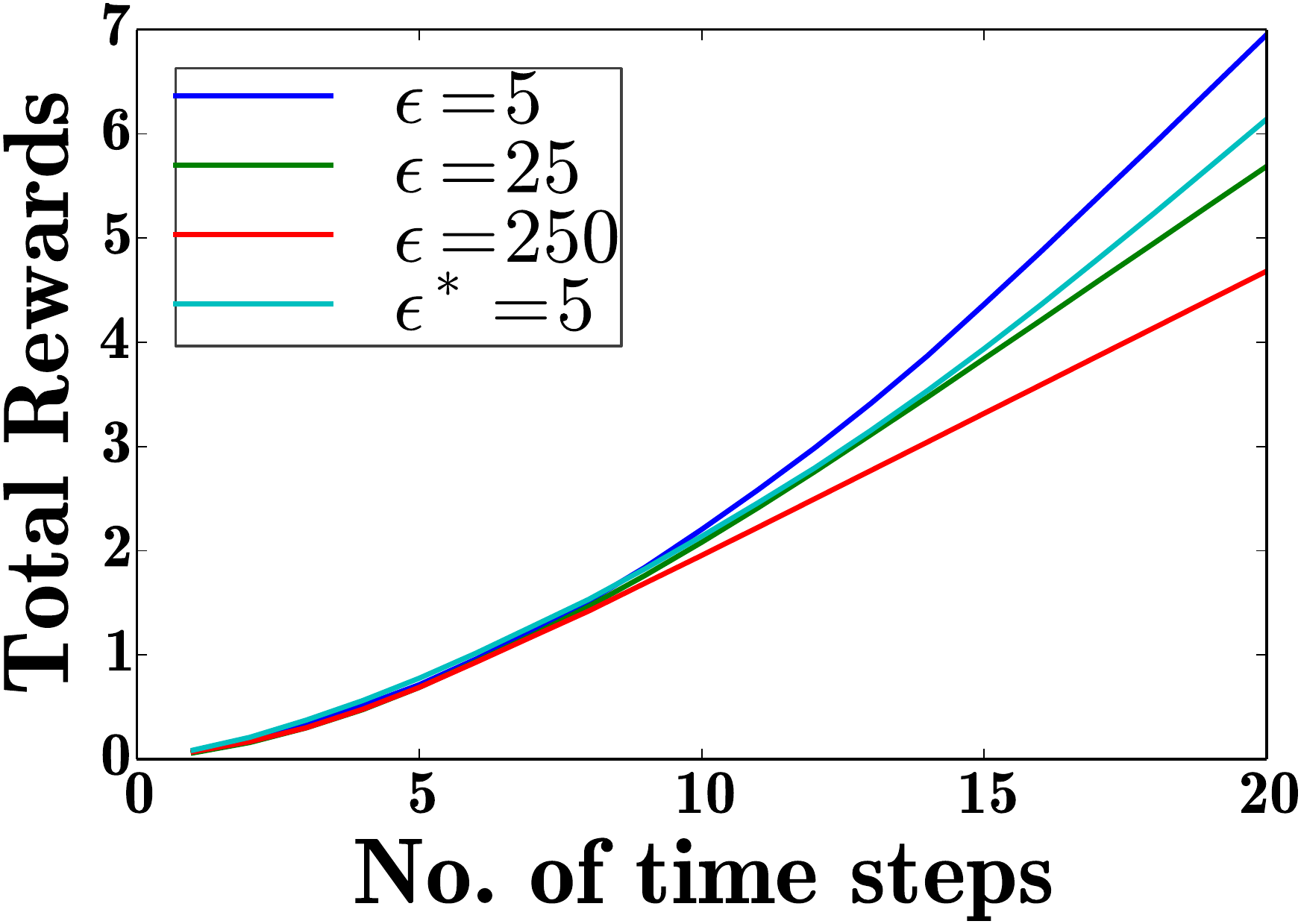} & 
\hspace{-0mm}\includegraphics[width=2.77cm]{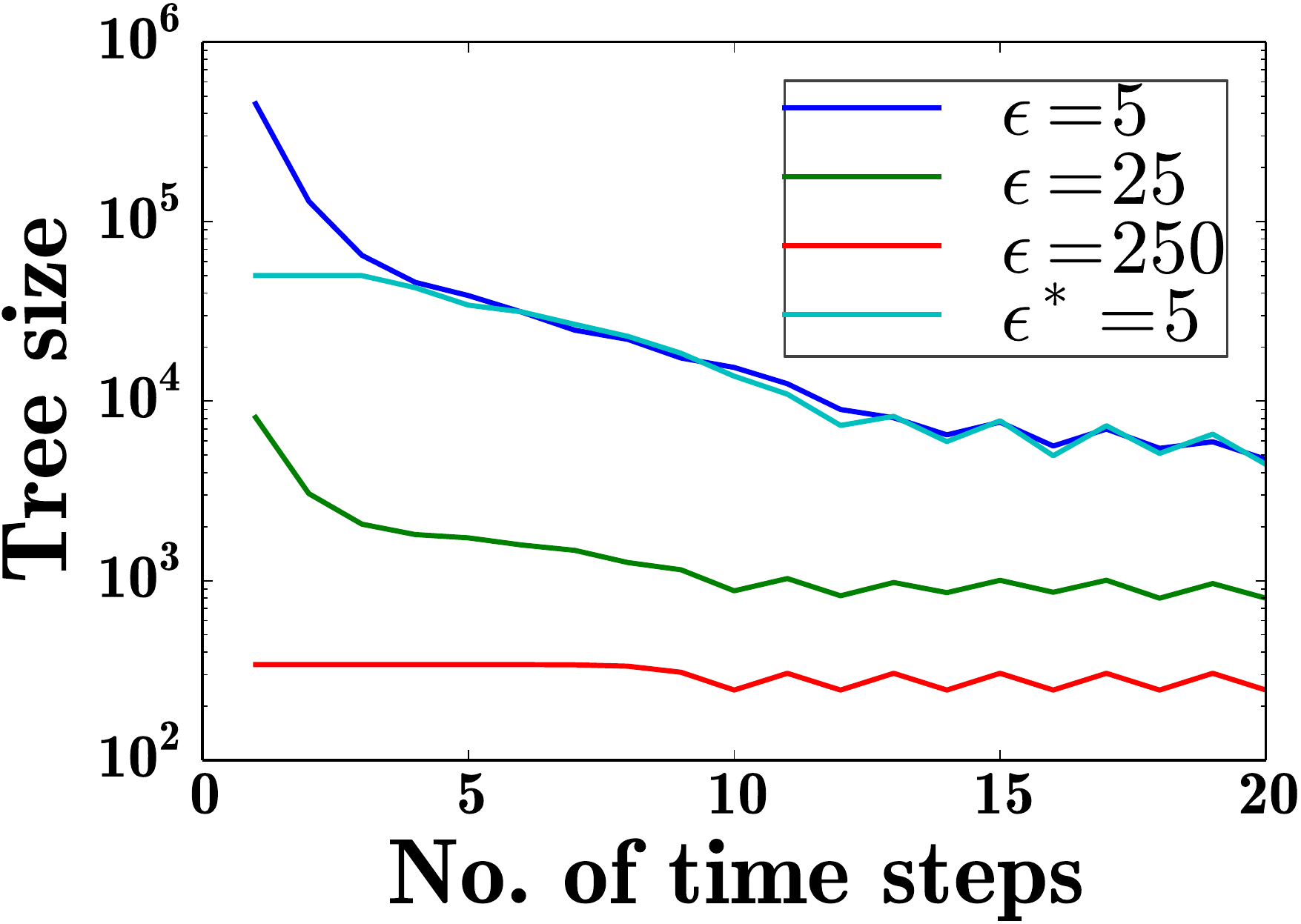}\vspace{-2mm}\\
\hspace{2mm}{\scriptsize (a)} & \hspace{4mm}{\scriptsize (b)} \vspace{-0mm}\\
\hspace{-0mm}\includegraphics[width=2.77cm]{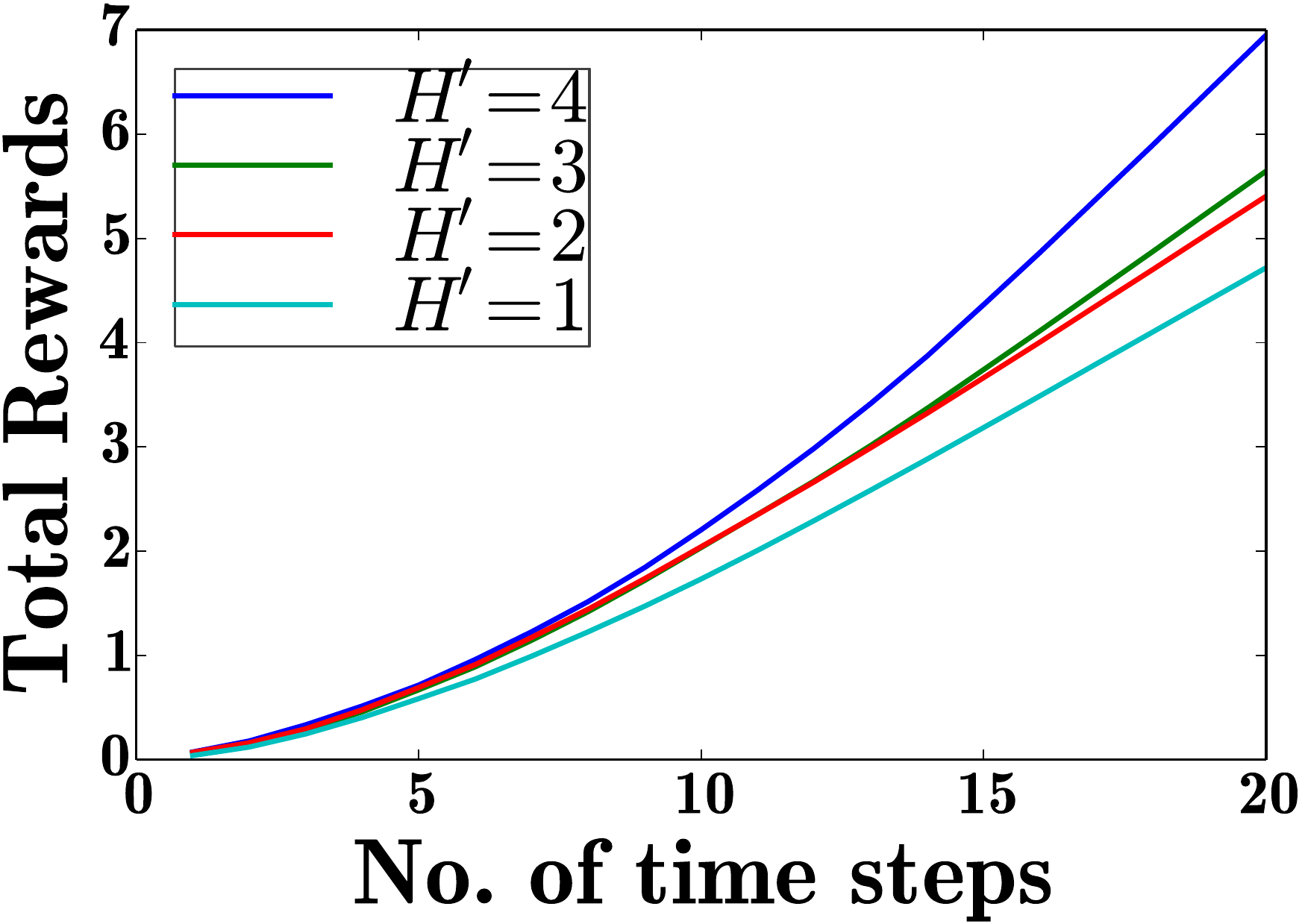} &  \hspace{-0mm}\includegraphics[width=2.77cm]{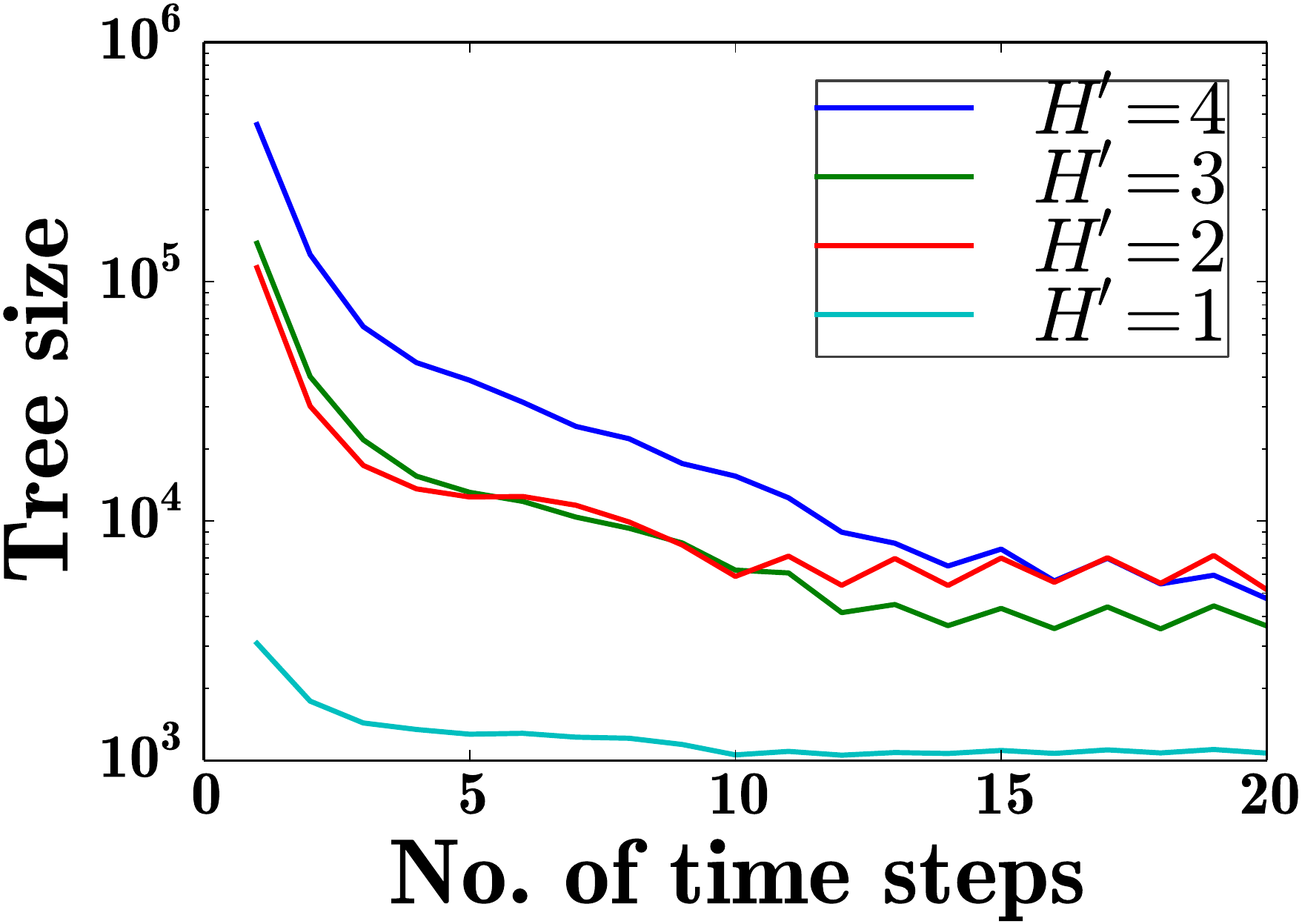}\vspace{-2mm}\\
\hspace{2mm}{\scriptsize (c)} & \hspace{3mm}{\scriptsize (d)}\vspace{-1mm}
\end{tabular}
}\vspace{-4mm}\\
{vs. no. of time steps for energy harvesting task.  The plot of $\epsilon^*=5$ uses our anytime variant with a maximum tree size of $5 \times 10^4$ nodes 
 while the plot of $\epsilon=250$ effectively assumes 
 maximum likelihood observations during planning like that of nonmyopic UCB \cite{RamosUAI14}.}\vspace{-4.9mm}
 %\vspace{-5mm}
\end{tabular}}
\end{table}
Fig.~\ref{fig:gpp_synthetic_log} shows results of performances of our $\epsilon$-GPP policy and its anytime variant averaged over $30$ independent realizations of the wind speed field. 
It can be observed that the gradients of the achieved total rewards (i.e., power production) 
%generally
increase over time, which indicate a higher obtained reward with an increasing number of  time steps as the robot can exploit the environment more effectively with the aid of exploration from previous time steps.
The gradients eventually stop increasing
%The increase in mean reward eventually levels off 
when the robot enters a perceived high-reward region. Further exploration is deemed unnecessary as it is unlikely to find another preferable location within $H'$ time steps; so, the robot remains near-stationary for the remaining time steps.
It can also be observed 
%from Figs.~\ref{fig:gpp_synthetic_log}b and~\ref{fig:gpp_synthetic_log}d 
that the incurred time is much higher in the first few time steps.
This is expected because the posterior variance $\sigma_{s_{t+1}|\Shist{t}}$ decreases with increasing time step $t$, thus requiring a decreasing deterministic sampling size $n$ to satisfy \eqref{eq:PartitionIneq}.

Initially, all $\epsilon$-GPP policies achieve similar total rewards as the robots begin from the same starting location. 
After some time, $\epsilon$-GPP policies with lower user-specified loss bound $\epsilon$ and longer online planning horizon $H'$ achieve considerably higher total rewards  
%But, their achieved mean rewards diverge after some time: Considerable benefits in  can be 
%(Figs.~\ref{fig:gpp_synthetic_log}a and~\ref{fig:gpp_synthetic_log}c) 
at the cost of more incurred time.
% (Figs.~\ref{fig:gpp_synthetic_log}b and~\ref{fig:gpp_synthetic_log}d).
%, respectively.
In particular, it can be observed that a robot assuming maximum likelihood observations during planning (i.e., $\epsilon=250$) like that of nonmyopic UCB or using a greedy policy (i.e., $H'=1$) performs poorly very quickly.
In the former case (Fig.~\ref{fig:gpp_synthetic_log}a), the gradient of its total rewards stops increasing quite early (i.e., from time step $9$ onwards), which indicates that its perceived local maximum is reached prematurely. 
%In general, a lower user-specified loss bound $\epsilon$ (e.g., $\epsilon=5.0$) allows the robot to partake in meaningful exploration and hence achieve better mean reward over all $20$ time steps 
%, as evidenced by $\epsilon=5.0$ being strictly better than all other $\epsilon$-GPP policies over all $20$ time steps.
%But, this requires an incurred time of more than $10$ times the search tree size.
% are required to achieve this performance improvement.
%A similar observation can also be made from Fig.~\ref{fig:gpp_synthetic_log}b, which demonstrates a similar trend for deeper search horizons.
%For example, $H=3$ performs slightly better than $H=2$ under the stated conditions, despite $\epsilon$ being chosen such that $H=2$ requires more resources. 
Interestingly, it can be observed from Fig.~\ref{fig:gpp_synthetic_log}d 
%and~\ref{fig:gpp_synthetic_log}d 
that the $\epsilon$-GPP policy with $H'=2$ and $\epsilon=0.06$ incurs more time than that with $H'=3$ and $\epsilon=0.8$ despite the latter achieving higher total rewards. This suggests trading off tighter loss bound $\epsilon$ for longer online planning horizon $H'$, especially when $\epsilon$ is too small that in turn requires a very large $n$ and consequently incurs significantly more time\cref{sucks}.\vspace{0.5mm}
%
%This suggests that under certain situations, it may be advantageous to accept poor value estimates in exchange for a deeper search horizon.
%We hypothesize that keeping $H$ constant while decreasing $\epsilon$ does not grant any significant benefit past a certain point.
%Further additional improvements are likely to require the adoption of a deeper search horizon.
%

\begin{figure}
%{r}{6.9cm}\vspace{-3mm}
\begin{tabular}{ccc}
\hspace{-2.3mm}
\includegraphics[width=2.7cm]{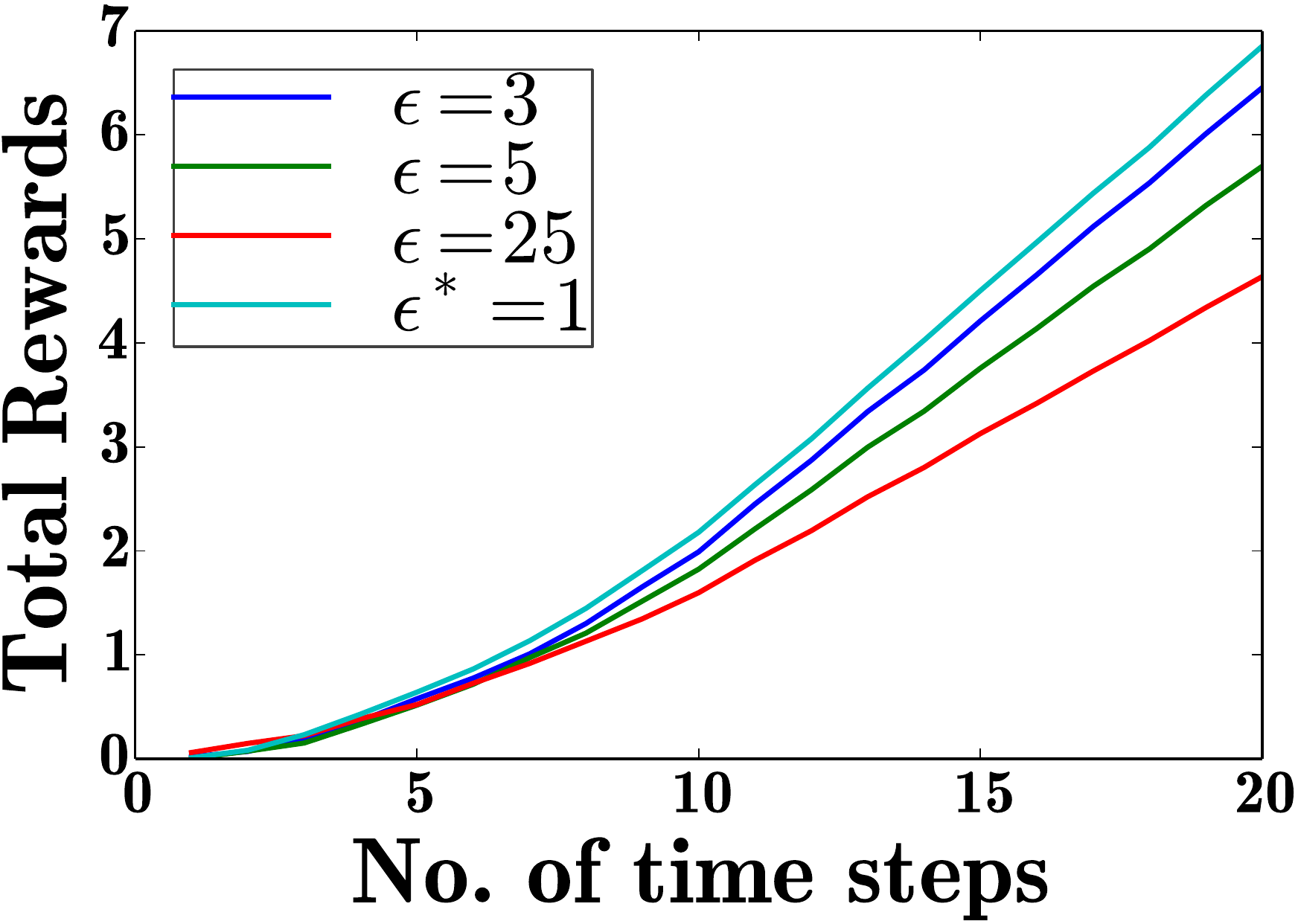} & \hspace{-4mm}\includegraphics[width=2.7cm]{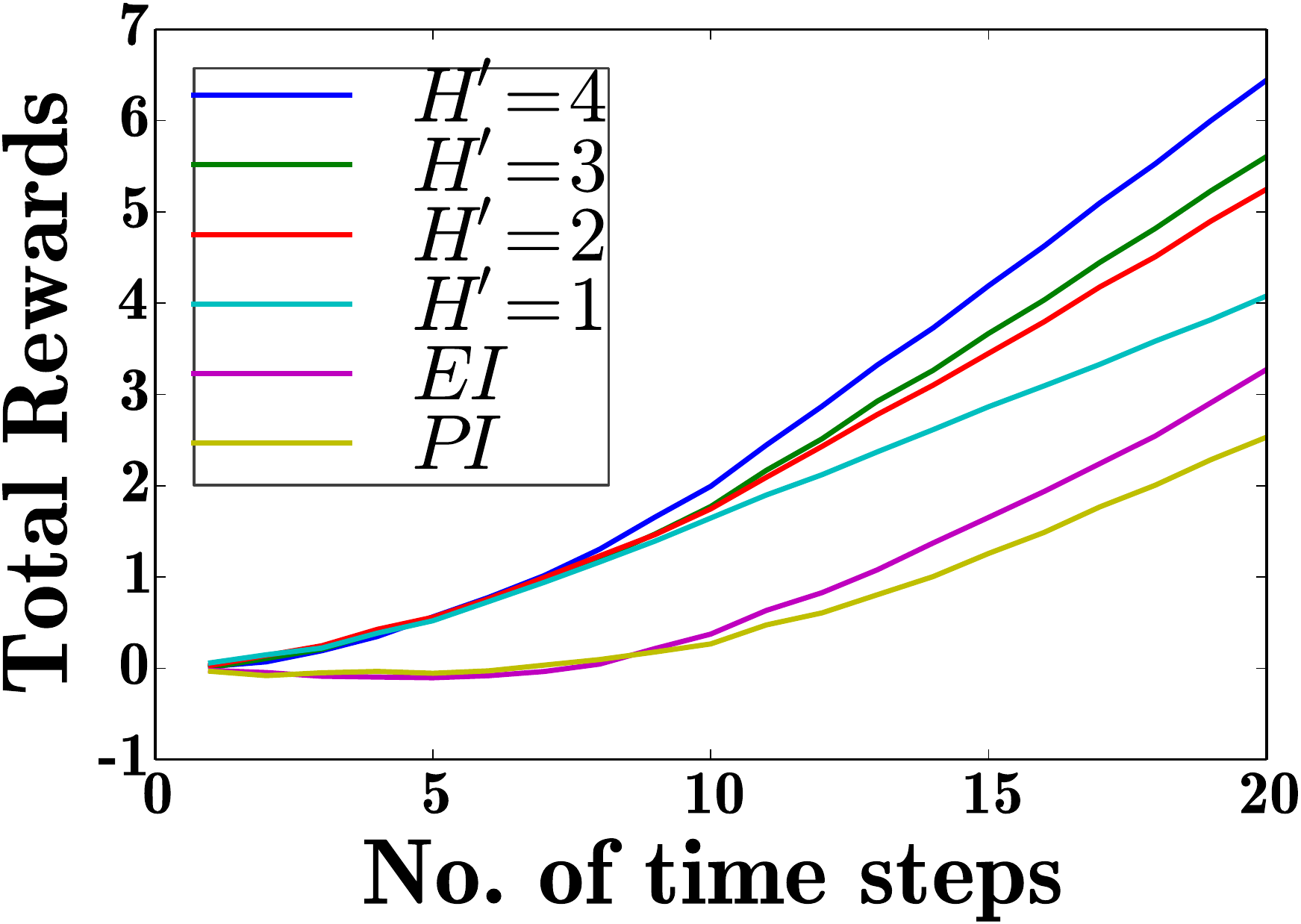} & \hspace{-4mm}\includegraphics[width=2.7cm]{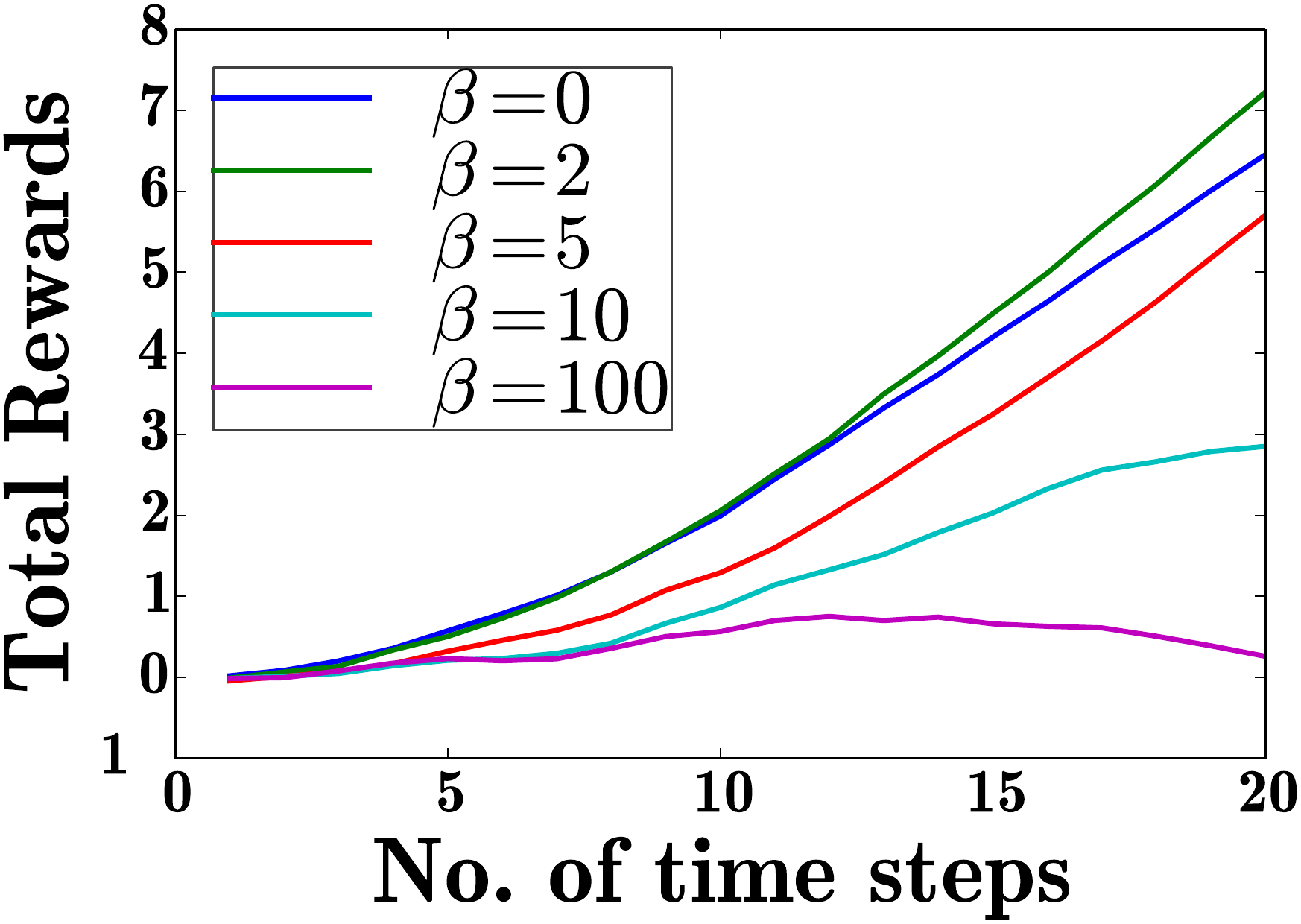}\vspace{-2.5mm}\\
\hspace{-2.5mm}{\scriptsize (a)} & \hspace{-4mm}{\scriptsize (b)} & \hspace{-4mm}{\scriptsize (c)}\vspace{-4.5mm}
\end{tabular}
\caption{Graphs of total normalized\cref{crass} rewards of $\epsilon$-GPP policies using UCB-based rewards with (a) $H'=4$, $\beta=0$, and varying $\epsilon$, (b) varying $H'=1,2,3,4$ (respectively, $\epsilon=0.002, 0.003, 0.4, 2$) and $\beta=0$, and (c) $H'=4$, $\epsilon=1$, and varying $\beta$ vs. no. of time steps for BO on real-world lg-K field.
The plot of $\epsilon^*=1$ uses our anytime variant with a maximum tree size of $3 \times 10^4$ nodes while the plot of $\epsilon=25$ effectively assumes maximum likelihood observations during planning like that of nonmyopic UCB.\vspace{-5mm}}
\label{fig:gpp_realdata}
\end{figure}

\noindent
{\bf BO on Real-World Log-Potassium Concentration Field.} 
An agricultural robot is tasked to find the peak lg-K measurement (i.e., possibly in an over-fertilized area) while exploring the Broom's Barn farm \cite{Webster01}. It is driven by the UCB-based reward function described under `BO' in Section~\ref{ch:rewardfunctions}.
Fig.~\ref{fig:gpp_realdata} shows results of performances of our $\epsilon$-GPP policy and its anytime variant, nonmyopic UCB (i.e., $\epsilon=25$), and greedy PI, EI, UCB (i.e., $H'=1$) averaged over $25$ randomly selected robot's initial starting location. 
It can be observed from Figs.~\ref{fig:gpp_realdata}a and~\ref{fig:gpp_realdata}b that the gradients of the achieved total normalized\footnote{To ease interpretation of the results, each reward is normalized by subtracting the prior mean from it.\label{crass}} rewards %achieved by our $\epsilon$-GPP policy and its anytime variant 
generally increase over time.
In particular, from Fig.~\ref{fig:gpp_realdata}a,  nonmyopic UCB assuming maximum likelihood observations during planning obtains much less total rewards than the other $\epsilon$-GPP policies and the anytime variant after $20$ time steps and 
finds a maximum lg-K measurement of $3.62$ that is at least $0.4\sigma_y$ worse after $20$ time steps.
%0.096296/0.238746727726267
The performance of the anytime variant is comparable to that of our best-performing $\epsilon$-GPP policy with $\epsilon= 3$.
From Fig.~\ref{fig:gpp_realdata}b, the greedy policy (i.e., $H'=1$) with $\beta=0$ performs much more poorly than its nonmyopic $\epsilon$-GPP counterparts
and finds a maximum lg-K measurement of $3.56$ that is lower than that of greedy PI and EI due to its lack of exploration. By increasing $H'$ to $2$-$4$, our $\epsilon$-GPP policies with $\beta=0$ outperform greedy PI and EI as they can naturally and jointly optimize the exploration-exploitation trade-off.
Interestingly, Fig.~\ref{fig:gpp_realdata}c shows that our $\epsilon$-GPP policy with $\beta = 2$ achieves the highest total rewards after $20$ time steps, which indicates the need of a slightly stronger exploration behavior than that with $\beta=0$.
%no weighted exploration term $\beta\sigma_{s_{t+1}|\Shist{t}}$ (i.e., $\beta=0$). 
This may be explained by a small length-scale (i.e., spatial correlation) of the lg-K field, thus requiring some exploration to find the peak measurement.
By increasing $H'$ beyond $4$ or with larger spatial correlation\if\myproof1 (Appendix~\ref{treesize1}), \fi\if\myproof0 \cite{AA16}, \fi
we expect a diminishing role of the $\beta\sigma_{s_{t+1}|\Shist{t}}$ term.
It can also be observed that aggressive exploration (i.e., $\beta\geq 10$) hurts the performance. Results of the tree size (i.e., incurred time) of our $\epsilon$-GPP policy and its anytime variant are in\if\myproof1 Appendix~\ref{treesize2}. \fi\if\myproof0 \cite{AA16}.\vspace{-2.6mm}\fi

\if\myproof1 Due to lack of space, we present additional experimental results for BO on simulated plankton density field in Appendix~\ref{treesize1} that yield similar observations to the above.\fi

\section{Conclusion}\vspace{-1mm}
This paper describes a novel nonmyopic adaptive $\epsilon$-GPP framework endowed with a general class of Lipschitz continuous reward functions that can unify some AL and BO criteria and be used for defining new tasks/problems. In particular, it can jointly and naturally optimize the exploration-exploitation trade-off. 
%To plan in real time, we propose an  anytime variant of $\epsilon$-GPP with performance guarantee and 
We theoretically guarantee the performances of our $\epsilon$-GPP policy and its anytime variant and
empirically demonstrate their effectiveness in BO and an energy harvesting task.
For our future work, we plan to scale up $\epsilon$-GPP and its anytime variant for big data using parallelization \cite{LowUAI13,LowAAAI15}, online learning \cite{Xu2014}, and stochastic variational inference \cite{NghiaICML15} and extend them to handle unknown hyperparameters \cite{NghiaICML14}.\vspace{1mm}
%,Andreas07 and develop sparse/stochastic sampling versions of them \cite{Asmuth2011} with performance guarantees based on concentration inequalities for comparison.\vspace{1mm}
%we plan to scale up $\epsilon$-GPP and its anytime variant for large data using online sparse GP models \cite{Xu14} and extend them to handle unknown hyperparameters \cite{NghiaICML14,Andreas07}.

\noindent
{\bf Acknowledgments.}
This work was supported by Singapore-MIT Alliance for Research and Technology Subaward Agreement No. $52$ R-$252$-$000$-$550$-$592$.\vspace{-2.6mm}

\bibliographystyle{aaai}
\bibliography{icml2015}

\if \myproof1
\clearpage
\onecolumn
\appendix
\section{Proof of Lemma~\ref{rewlc}}
\label{geeez2}
Since $h_{\sigma_{s_{t+1}|\Shist{t}}}$ and  $g_{\sigma_{s_{t+1}|\Shist{t}}}$ are Lipschitz continuous with the respective Lipschitz constants $\ell_1$ and $\ell_2(\sigma_{s_{t+1}|\Shist{t}})$ (Section~\ref{ch:rewardfunctions}),
$\mathbb{E}[ \rfn(Z_{t+1}, \Shist{t+1})|d_t, s_{t+1} ]$ is Lipschitz continuous in $\mu_{s_{t+1}|d_{t}}$ with Lipschitz constant $\ell_1 + \ell_2(\sigma_{s_{t+1}|\Shist{t}})$.
Then, by using the definition of GP posterior mean $\mu_{s_{t+1}|d_{t}}$
%\eqref{eq:predictive} 
and Cauchy-Schwarz inequality, Lemma~\ref{rewlc} results.
%\eqref{geez} results.
%
\section{Other Complex Reward Functions Represented by $R$}
\label{aeg}
When sampling a natural phenomenon, locations featuring a specific band of field measurements may be preferred.
Suppose that a robot is tasked to gather algal samples from an aquatic environment. 
It is well established that algal growth is faster in certain temperature range and survival is not possible at either extremes.
The temperature field can be modeled as a GP.
A rectangular reward function can be used to direct the robot to search among regions with the specified temperature range.
The rectangular reward function can be viewed as a difference of two unit step functions, each of which is of the form:
%$\rfn(z_{t+1}, \Shist{t+1}) = 
$\rfn_2(z_{t+1})$ gives a value of $1$ if $z_{t+1}>a$, and $0$ otherwise where $a$ is some user-defined constant. 
For such a step function, $g_{\sigma_{s_{t+1}|\Shist{t}}}(u)=1-\Phi((a-u)/\sigma_{s_{t+1}|\Shist{t}})$
%, $\ell_1=0$, 
and $\ell_2(\sigma_{s_{t+1}|\Shist{t}})=1/(\sqrt{2\pi}\sigma_{s_{t+1}|\Shist{t}})$. 

Another interesting reward function is the Gaussian distribution itself\footnote{Convolution between two Gaussian distributions yields another Gaussian which is Lipschitz continuous. An alternative formulation is to set $\rfn_1(z_{t+1})=\mathcal{N}(0, 1)$ since the reward function is itself Lipschitz continuous.} which may be represented using $\rfn_2(z_{t+1})=\mathcal{N}(0, 1)$, 
$g_{\sigma_{s_{t+1}|\Shist{t}}}(u)=(2\pi(1+\sigma_{s_{t+1}|\Shist{t}}^2))^{-1/2}\exp(-u^2/(2(1+\sigma_{s_{t+1}|\Shist{t}}^2)))$
%, $\ell_1=0$, 
and $\ell_2(\sigma_{s_{t+1}|\Shist{t}})=(\sqrt{2\pi}(1+\sigma_{s_{t+1}|\Shist{t}}^2))^{-1}\exp(-1/2)$.
%
%\section{Appendix}
\section{Proof of Theorem~\ref{th:LipschitzAll}}
\label{akaka}
We will give a proof by induction on $t$ that 
 \begin{equation}
 |V^*_t(d_t) - V^*_t(d_{t}')| \le L_{t}(\Shist{t}) \lVert\Zhist{t}-\Zhist{t}'\rVert\ .
 \label{whatthe}
 \end{equation}
When $t=H$,
$|V_{H}^{*}(d_H) - V_{H}^{*}(d_{H}')|  = 0 \le L_{H}(\Shist{H})\lVert\mathbf{z}_{H} - \mathbf{z}_{H}'\rVert$.
Supposing \eqref{whatthe} holds for $t+1$ (i.e., induction hypothesis), we will prove that it holds for $0 \leq t < H$.
%Inductive case where $0 \le t < H$. Suppose that we have proven the theorem till $t=T+1$.  We now prove the case for $t=T$.
Without loss of generality, assume that $V_t^{*}(d_t) \ge V_t^{*}(d_{t}')$.
Let $s^*_{t+1}\triangleq \pi^*(d_t)$ and $\Delta_{t+1} \triangleq \mu_{s_{t+1}^{*}| d_t} - \mu_{s_{t+1}^{*}| d_t'}$. Then, $|\Delta_{t+1}|\leq\alpha(\Shist{t} \oplus s_{t+1}^{*}) \lVert\Zhist{t}-\Zhist{t}'\rVert$.
%
% Pick any optimal action for $d_T$ and denote it as $s^*_{T+1}$.
% Define $\Delta \mu = \mu(s_{T+1}^{*}, d_T) - \mu(s_{T+1}^{*}, d_T') = \mathbf{B}(\Shist{T+1}^{*})(\Zhist{T} - \Zhist{T}')$, ie. the difference in posterior means between $d_T$ and $d_{T}'$ after moving to location $s^*_{T+1}$.
%\small
$$
\begin{array}{l}
\displaystyle V_t^{*}(d_t) - V_t^{*}(d_{t}') \vspace{1mm}\\ 
\displaystyle   \le Q_t^{*}(d_t, s^{*}_{t+1}) - Q_t^{*}(d_{t}', s^{*}_{t+1}) \vspace{1mm}\\
\displaystyle   = \mathbb{E}[ \rfn(Z_{t+1}, \Shist{t} \oplus s_{t+1}^{*})|d_t,s^{*}_{t+1} ] -  \mathbb{E}[ \rfn(Z_{t+1}, \Shist{t} \oplus s_{t+1}^{*})|d'_t,s^{*}_{t+1} ]\ +\vspace{1mm}\\
%h_{\sigma_{s_{t+1}^{*}|\Shist{t}}}(\mu_{s_{t+1}^{*}| d_t}) - h_{\sigma_{s_{t+1}^{*}|\Shist{t}}}(\mu_{s_{t+1}^{*}| d_t'})\ + \vspace{1mm}\\
%\quad g_{\sigma_{s_{t+1}^{*}|\Shist{t}}}(\mu_{s_{t+1}^{*}| d_t}) - g_{\sigma_{s_{t+1}^{*}|\Shist{t}}}(\mu_{s_{t+1}^{*}| d_t'})\ + \vspace{1mm}\\
\displaystyle \quad \int_{-\infty}^{\infty}p(z_{t+1}|d_t,s^{*}_{t+1})\ V_{t+1}^{*}(d_{t+1}) \ \text{d}z_{t+1} -  \int_{-\infty}^{\infty}p(z'_{t+1}|d'_t,s^{*}_{t+1})\ V_{t+1}^{*}(d_{t+1}') \ \text{d}z'_{t+1}\vspace{1mm}\\
\displaystyle   \le  |\Delta_{t+1}|\left(\ell_1 + \ell_2(\sigma_{s^{*}_{t+1}|\Shist{t}})\right) + \int_{-\infty}^{\infty}p(z_{t+1}|d_t,s^{*}_{t+1})L_{t+1}(\Shist{t} \oplus s_{t+1}^{*}) \lVert(\Zhist{t}-\Zhist{t}') \oplus \Delta_{t+1}\rVert\ \text{d}z_{t+1}\vspace{1mm}\\
\displaystyle   = |\Delta_{t+1}|\left(\ell_1 + \ell_2(\sigma_{s^{*}_{t+1}|\Shist{t}})\right) +  L_{t+1}(\Shist{t} \oplus s_{t+1}^{*}) \lVert(\Zhist{t}-\Zhist{t}') \oplus \Delta_{t+1}\rVert\vspace{1mm}\\
\displaystyle   \leq \alpha(\Shist{t} \oplus s_{t+1}^{*}) \lVert\Zhist{t}-\Zhist{t}'\rVert\left(\ell_1 + \ell_2(\sigma_{s^{*}_{t+1}|\Shist{t}})\right) + L_{t+1}(\Shist{t} \oplus s_{t+1}^{*})\sqrt{1+\alpha(\Shist{t} \oplus s_{t+1}^{*})^2} \lVert\Zhist{t}-\Zhist{t}'\rVert\vspace{1mm}\\
\displaystyle   =  \Big( \alpha(\Shist{t} \oplus s_{t+1}^{*}) \left(\ell_1 + \ell_2(\sigma_{s^{*}_{t+1}|\Shist{t}})\right) + L_{t+1}(\Shist{t} \oplus s_{t+1}^{*})\sqrt{1+\alpha(\Shist{t} \oplus s_{t+1}^{*})^2}\Big)\lVert\Zhist{t}-\Zhist{t}'\rVert\vspace{1mm}\\
\displaystyle \leq L_{t}(\Shist{t}) \lVert\Zhist{t}-\Zhist{t}'\rVert\ .
  \end{array}
$$
The first equality is due to \eqref{eq:OptimalValFunDef}.
The second inequality follows from Lemma~\ref{rewlc}, change of variable $z'_{t+1}\triangleq z_{t+1}-\Delta_{t+1}$, and the induction hypothesis.
The last inequality is due to definition of $L_{t}$ (Definition~\ref{def:ValFunGenLip}).
\pagebreak
\section{Illustration of Deterministic Sampling Strategy}
\label{dsampeg}
\begin{figure}[h]
\centering
% First draw the upper distribution.
    % Shade the critical region:
    
    \begin{tikzpicture}
      %\draw[->] (0,-0.5) -- (0,4.2) node[above] {$y$};
            
      \draw[scale=1,domain=-5:5,smooth,variable=\x,black, line width=3.0pt] plot ({\x},{4*exp(-\x*\x*0.5*0.25)}) (4.5, 1) node[above, black] {$p(z_{t+1}|d_t, s_{t+1})$};
      
      \fill[fill=red] (3,0) -- plot [domain=3:5] ({\x},{4*exp(-\x*\x*0.5*0.25)}) -- (5,0) -- cycle;
      
      \fill[fill=red] (-5,0) -- plot [domain=-5:-3] ({\x},{4*exp(-\x*\x*0.5*0.25)}) -- (-3,0) -- cycle;
            
      \fill[fill=gray] (-3,0) -- plot [domain=-3:3] ({\x},{4*exp(-\x*\x*0.5*0.25)}) -- (3,0) -- cycle;      
      
      \foreach \x [evaluate = \x as \xpp using 4*exp(-\x*\x*0.5*0.25)] in {-3,-1, 1, 3} 
      \draw[dashed, black, line width=1pt] (\x, 0) -- (\x,\xpp );

     \foreach \x [evaluate = \x as \xpp using 5+1-2.5*2.5/5+3*ln(1+2.5+\x)] in {-2, 0, 2}
     \draw[dotted, green, line width=1pt] (\x, 0) -- (\x,\xpp);
     
     \foreach \x [evaluate = \x as \xpp using 5+(1-(\x+5) * (\x+5)/5, evaluate = \x as \xppp using 4*exp(-\x*\x*0.5*0.25)] in {-3}
     \draw[dotted, green, line width=1pt] (\x, \xppp) -- (\x,\xpp);

     \foreach \x [evaluate = \x as \xpp using 5+1-2.5*2.5/5+3*ln(1+2.5+\x), evaluate = \x as \xppp using 4*exp(-\x*\x*0.5*0.25)] in {3}
     \draw[dotted, green, line width=1pt] (\x, \xppp) -- (\x,\xpp);
           
      \foreach \x in {-5, -4, -3, -2, 2, 	3, 4, 5}
       \draw (\x,0) -- (\x,-0.1) node[below] {$\x \sigma$};
      \draw (-1,0) -- ( -1,-0.1) node[below] {$-\sigma$};
      \draw ( 0,0) -- ( 0,-0.1) node[below] {$0$};
      \draw ( 1,0) -- ( 1,-0.1) node[below] {$\sigma$};
      
	  \draw[->, black, line width =1] (-5,5) -- (5.5,5) node[black,right] {$z_{t+1}$};
	    \draw (-3,5) -- ( -3,4.9) node[below] {$z^{0}$} (-4,0) node[above] {$w^0$};      
      \draw (-2,5) -- ( -2,4.9) node[below] {$z^{1}$} ( -2,1) node[above] {$w^{1}$}; 
      \draw (0,5) -- ( 0,4.9) node[below] {$z^{2}$} ( 0,1.7) node[above] {$w^{2}$}; 
      \draw (2,5) -- ( 2,4.9) node[below] {$z^{3}$} ( 2,1) node[above] {$w^{3}$}; 
	    \draw (3,5) -- ( 3,4.9) node[below] {$z^{4}$} (4,0) node[above] {$w^4$};
	                	 
	 \draw[scale=1,domain=-5:-2.5,smooth,variable=\x,brown, line width=3.0pt, line cap = round] plot ({\x},{5+(1-(\x+5) * (\x+5)/5});
	 
	 \draw[scale=1,domain=-2.5:3.5,smooth,variable=\x,brown, line width=3.0pt, line cap = round] plot ({\x},{5+1-2.5*2.5/5+3*ln(1+2.5+\x)}) -- (5,9) node[below, black] {$\begin{array}{l}R_1(z_{t+1})\ +\\
	 V_{t+1}^{*}(\langle \Shist{t+1}, \Zhist{t}\oplus z_{t+1}\rangle)\end{array}$};

     % The approximated partial value 
     \foreach \x [evaluate = \x as \xpp using 5+1-2.5*2.5/5+3*ln(1+2.5+\x)] in {-2, 0, 2}
     \draw[solid, blue, line width =3.0] (\x-1, \xpp) -- (\x+1,\xpp);
     
     \foreach \x [evaluate = \x as \xpp using 5+(1-(\x+5) * (\x+5)/5] in {-3}
     \draw[solid, blue, line width =3.0] (-5, \xpp) -- (-3,\xpp);
     
     \foreach \x [evaluate = \x as \xpp using 5+1-2.5*2.5/5+3*ln(1+2.5+\x)] in {3}
     \draw[solid, blue, line width =3.0] (3, \xpp) -- (5,\xpp)          node[right, black] {};
     
     % vertical connectors
     \foreach \x [evaluate = \x as \xpp using 5+1-2.5*2.5/5+3*ln(1+2.5+\x), evaluate = \x as \xppp using 5+1-2.5*2.5/5+3*ln(1+2.5+(\x+2)] in {-2, 0}
     \draw[solid, blue, line width =3.0, line cap = rect] (\x+1, \xpp) -- (\x+1,\xppp);
               
     \foreach \x [evaluate = \x as \xpp using 5+(1-(\x+5) * (\x+5)/5, evaluate = \x as \xppp using 5+1-2.5*2.5/5+3*ln(1+2.5+(\x+1)] in {-3}
     \draw[solid, blue, line width =3.0, line cap = rect] (\x, \xpp) -- (\x,\xppp);
          
     \foreach \x [evaluate = \x as \xpp using 5+1-2.5*2.5/5+3*ln(1+2.5+(\x), evaluate = \x as \xppp using 5+1-2.5*2.5/5+3*ln(1+2.5+(\x-1)] in {3}
     \draw[solid, blue, line width =3.0, line cap = rect] (\x, \xpp) -- (\x,\xppp);

	% labels for zetas
	\foreach \x [evaluate = \x as \xpp using int(\x/2+2)] in {-2, 0, 2}
	     \draw[<->, gray, line width =1.5, line cap = rect] (\x-1, -0.7) -- node[black,below] {$\zeta_{\xpp}$} ++ (2,0);

	\draw[->, red, line width =1.5] (-5, -0.7) -- node[black,below] {$\zeta_0$} ++ (2,0);
	
	\draw[<-, red, line width =1.5] (3.0, -0.7) -- node[black,below] {$\zeta_4$} ++ (2,0);
	
	% tau
	\draw[<->, black, line width =1.5, line cap = rect] (0, -1.5) -- node[black,below] {$\tau\sigma$} ++ (3,0);
        \draw[->, black, line width =1] (-5,0) -- (5.5,0) node[black,right] {$z_{t+1}$};
   
    \end{tikzpicture}
    \caption{Example of a possible partition with $\mu_{s_{t+1}|d_t} = 0$, $\tau=3$, and $n=5$ where we let $\sigma\triangleq\sigma_{s_{t+1}|\Shist{t}}$ for notational simplicity. 
At the bottom, the measurement space of $p(z_{t+1}|d_t, s_{t+1})$ is partitioned into $5$ intervals $\zeta_0$, $\zeta_1$, $\zeta_2$, $\zeta_3$, and $\zeta_{4}$ such that intervals $\zeta_1$, $\zeta_2$, and $\zeta_{3}$ are equally spaced within the bounded gray region $[- 3\sigma, 3\sigma]$ while intervals $\zeta_0$ and $\zeta_{4}$ span the two infinitely long red tails.
The $5$ sample measurements $z^{0}$, $z^{1}$, $z^{2}$, $z^{3}$, and $z^{4}$ are then selected by setting $z^0$ as the upper limit of red interval $\zeta_0$, $z^{4}$
as the lower limit of red interval $\zeta_{4}$,
and $z^1$, $z^2$, and $z^{3}$ as the centers of the respective gray intervals $\zeta_1$, $\zeta_2$, and $\zeta_{3}$.
The weights $w^{0}$, $w^{1}$, $w^{2}$, $w^{3}$, and $w^{4}$ for the corresponding sample measurements $z^{0}$, $z^{1}$, $z^{2}$, $z^{3}$, and $z^{4}$ are defined as the areas under their respective intervals $\zeta_0$, $\zeta_1$, $\zeta_2$, $\zeta_3$, and $\zeta_{4}$ of the Gaussian predictive distribution $p(z_{t+1}|d_t, s_{t+1})$.
%The $3$ inner partitions $\zeta_1, \zeta_2, \zeta_3$ are equally spaced. Tails partitions $\zeta_0, \zeta_4$ are infinitely long. For non-tail partitions $\zeta_1, \zeta_2, \zeta_3$, the representative sample observations $z^1$, $z^2$ and $z^3$ are given by their mid-points. The weights are given by the region under the distribution curve, bounded by $\zeta_i$ (red regions represent tail probabilities).
At the top, the brown curve (i.e., Lipschitz continuous) denotes the sum of $R_1(z_{t+1})$ and the expected future rewards $V_{t+1}^{*}(\langle \Shist{t+1}, \Zhist{t}\oplus z_{t+1}\rangle)$ achieved by the GPP policy. 
The latter is unknown in practice, as explained in Section~\ref{gppfram}.
%The latter cannot be derived exactly due to an uncountable set of candidate measurements, as explained in Section~\ref{gppfram}. 
The blue curve shows the rectangular approximations using the sample measurements $z^{0}$, $z^{1}$, $z^{2}$, $z^{3}$, and $z^{4}$. Note that, in reality, $V^*_{t+1}$ cannot be accurately evaluated -- there will be additional errors accruing from future time steps that are not reflected in this figure but will be formally accounted for in the theoretical performance guarantee of our $\epsilon$-GPP policy, as shown in the proof of Theorem~\ref{th:MultistageError} in Appendix~\ref{jordan}.
}
    \label{fig:ds_partition}
\end{figure}
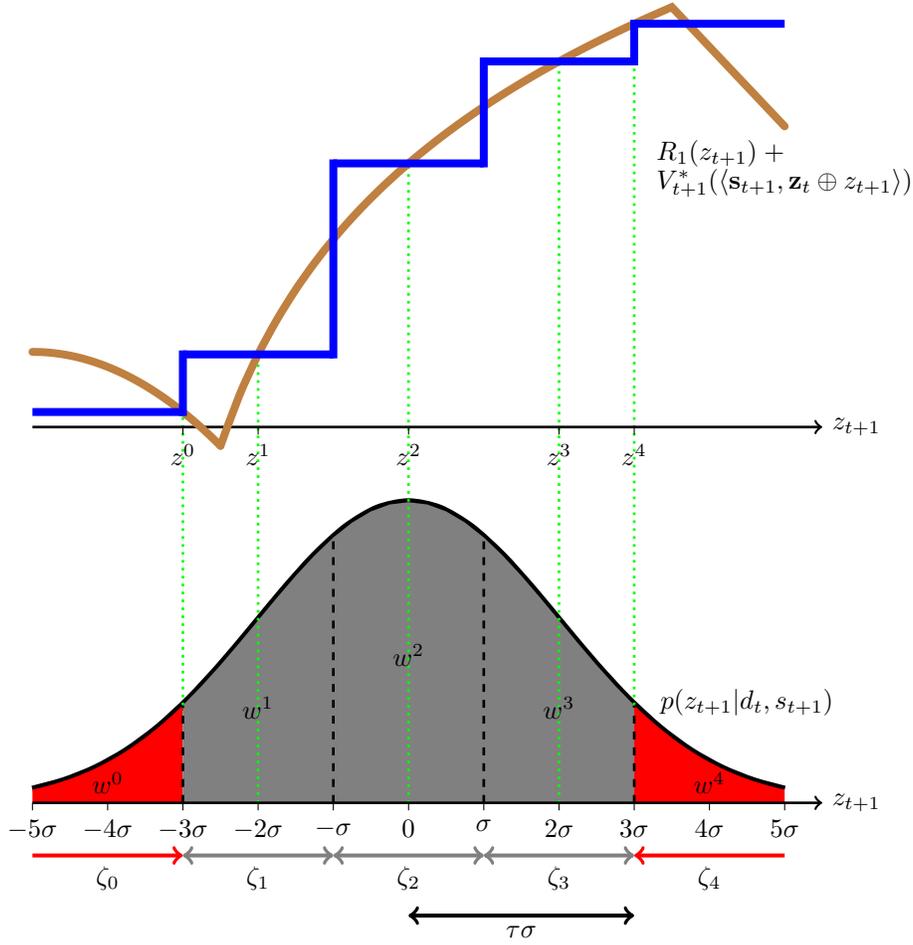
\section{Proof of Theorem~\ref{th:MultistageError}}
\label{jordan}
There are two sources of error arising in using $V_t^{\epsilon}$ in \eqref{eq:EpsilonValFunDef} to approximate $V_t^{*}$ in \eqref{eq:OptimalValFunDef}: 
(a) In our deterministic sampling strategy (Section~\ref{ch:SamplingStrategy}), only a finite number of sample measurements is selected for approximating the stage-wise expectation terms in \eqref{eq:OptimalValFunDef}, and (b) computing $V^{\epsilon}_t$ does not involve utilizing the values of $V^{\ast}_{t+1}$ but that of its approximation $V^{\epsilon}_{t+1}$ instead.
To facilitate capturing the error due to finite deterministic sampling described in (a), the following intermediate function is introduced:
 %$\tilde{U}_t^{A*}(d_t, s_{t+1})$ to be the value function supposing finite samples are taken at the current stage, but use the optimal value functions thereafter.
%That is, 
%\small
\begin{equation}
  U_t^{*}(d_t, s_{t+1}) \triangleq \displaystyle g_{\sigma_{s_{t+1}|\Shist{t}}}\hspace{-1mm}\left(\mu_{s_{t+1}|d_{t}}\right) +\rfn_3(\Shist{t+1}) +  \sum^{n-1}_{i=0} w^i \hspace{-1mm}\left(R_1(z^{i}) + V_{t+1}^{*}(\langle \Shist{t+1}, \Zhist{t}\oplus z^{i}\rangle)\right)
\label{ustar}
\end{equation}
for stages $t = 0,\ldots,H-1$ where $V_H^{*}(d_H)\triangleq 0$. 
%\begin{equation*}
% \begin{aligned}
%  \tilde{U}_t^{A*}(d_t, s_{t+1}) = \rfn_{3}(\Shist{t+1}) + g_{\sigma(\Shist{t+1})}(\mu({s_{t+1}}| d_t))\\ 
%  \phantom{=} + \sum_{j=0}^{m-1}w^{i} \Big(\rfn_1(z^{i}) + V_{t+1}^{*}(\Shist{t+1}, \Zhist{t} \oplus z^{i})\Big)
%  \end{aligned}
%\end{equation*}
%\normalsize
The following lemma bounds the error between ${U}_t^{*}$ and $Q_t^{*}$, which can then be used to bound the error between $V_t^{\epsilon}$ and $V_t^{*}$:
\begin{lemma} \label{th:SingleStageError}
% Selecting measurements $z^{i}$ and weights $w^{i}$ in accordance to the sampling strategy described in Section~\ref{ch:SamplingStrategy} gives, for all $0 \le t < H$:
For $t=0,\ldots,H-1$,
% \begin{equation*} \label{hideko}
  $|{U}_t^{*}(d_t, s_{t+1}) - Q_t^{*}(d_t, s_{t+1})| \le \lambda$.
% \end{equation*}
\end{lemma}
%\begin{proof}
\textit{Proof}.
When $n>2$, define $\displaystyle\bar{z} \triangleq \frac{z^{n-1}-z^{0}}{n-2}$. Then, 
%For convenience, we write $\zeta^{i}$ to be the integration limits as above.
$\zeta_{0}=[-\infty, z^{0}]$, $\zeta_{n-1}=[z^{n-1}, \infty]$, and $\zeta_{i}=[z^{0}+(i-1)\bar{z}, z^{0}+i\bar{z}]$ for $i=1,\ldots,n-2$.
For $i=0,\ldots,n-1$,
$$
 w^{i} = \int\limits_{z_{t+1}\in\zeta_{i}} p(z_{t+1}|d_t, s_{t+1})\ \text{d}z_{t+1}\ .
 $$
%
%\begin{equation*}
% \begin{aligned}
% w^{i} &= 
% \begin{cases}
%  \int\limits_{-\infty}^{z^{i}} p(z_{t+1}|d_t, s_{t+1}) dz_{t+1} \quad &j = 0\\
 % \int\limits_{z^{i}}^{\infty} p(z_{t+1}|d_t, s_{t+1}) dz_{t+1} \quad &j = m-1\\
%  \int\limits_{z^{(0)}+(j-1)\bar{z}}^{z^{(0)}+j\bar{z}} p(z_{t+1}|d_t, s_{t+1}) dz_{t+1}  \quad &0 < j < m-1\\
% \end{cases}
%\end{aligned}
%\end{equation*}
Observe that the integration limits form a partition (at least in the sense of a Riemannian integral) of $\mathbb{R}$, thus implying $\sum^{n-1}_{i=0} w^{i}=1$.
%Returning to the original error term,
%\small
\begin{equation} \label{eq:StepwiseProofKey}
\hspace{-1.9mm}
 \begin{array}{l}
\displaystyle |{U}_t^{*}(d_t, s_{t+1}) - Q_t^{*}(d_t, s_{t+1})| \\
\displaystyle   = \left|\sum_{i=0}^{n-1}\int\limits_{\zeta_{i}} p(z_{t+1}|d_t, s_{t+1}) \left(V^*_{t+1}(\langle \Shist{t+1}, \Zhist{t} \oplus z^{i} \rangle) + \rfn_1(z^{i})- V^*_{t+1}(\langle \Shist{t+1}, \Zhist{t} \oplus z_{t+1} \rangle)-\rfn_1(z_{t+1}) \right) \text{d}z_{t+1}\right| \\
\displaystyle   \le \sum_{i=0}^{n-1}\int\limits_{\zeta_{i}} p(z_{t+1}|d_t, s_{t+1}) \left|V^*_{t+1}(\langle \Shist{t+1}, \Zhist{t} \oplus z^{i} \rangle) + \rfn_1(z^{i})- V^*_{t+1}(\langle \Shist{t+1}, \Zhist{t} \oplus z_{t+1} \rangle)-\rfn_1(z_{t+1}) \right|\ \text{d}z_{t+1} \\
\displaystyle    \le (\ell_1 + L_{t+1}(\Shist{t+1}))\sum_{i=0}^{n-1}\int\limits_{\zeta_{i}}p(z_{t+1}|d_t, s_{t+1})\left|z^{i}-z_{t+1}\right| \text{d}z_{t+1}\ .
 \end{array}
\end{equation}
%\normalsize
%The first line is obtained by direct substitution. 
The last inequality follows from Corollary~\ref{th:LipschitzSingle} and the fact that $\rfn_1$ is Lipschitz continuous with Lipschitz constant $\ell_1$ (Section~\ref{ch:rewardfunctions}).
We now focus on the integral within the summation. When $i=0$,
%\small
\begin{equation} \label{eq:StepwiseProofTail}
 \begin{array}{l}
\displaystyle\int\limits_{\zeta_{0}}p(z_{t+1}|d_t, s_{t+1})\left|z^{0}-z_{t+1}\right|\ \text{d}z_{t+1}\\
\displaystyle  = \frac{\sigma_{s_{t+1}|\Shist{t}}}{\sqrt{2\pi}} \int_{-\infty}^{-\sdk}\exp\hspace{-1mm}\left(-\frac{y^2}{2}\right)(-\sdk-y)\ \text{d}y \\
\displaystyle  = \sigma_{s_{t+1}|\Shist{t}} \left( \frac{1}{\sqrt{2\pi}} \exp\hspace{-1mm}\left(-\frac{\sdk^2}{2}\right) - \sdk\Phi(-\sdk) \right)\\
\displaystyle  =\frac{1}{2} \sigma_{s_{t+1}|\Shist{t}} \kappa(\sdk)\ .%\\
%  = &\sigma(\Shist{t+1})\kappa(\sdk)
 \end{array}
\end{equation}
%\normalsize
The first equality can be obtained by plugging in the density function of a Gaussian and normalizing. 
%The rest of the lines follow by algebra.
A similar result can be obtained when $i=n-1$. When $n>2$,
%\small
\begin{equation} \label{eq:StepwiseProofInner}
 \begin{array}{l}
\displaystyle  \sum_{i=1}^{n-2}\int\limits_{\zeta_{i}}p(z_{t+1}|d_t, s_{t+1})\left|z^{i}-z_{t+1}\right|\text{d}z_{t+1}\\
\displaystyle  \le \frac{\bar{z}}{2}\sum_{i=1}^{n-2}\int\limits_{\zeta_{i}}p(z_{t+1}|d_t, s_{t+1})\ \text{d}z_{t+1}\\
\displaystyle  = \frac{\sdk\sigma_{s_{t+1}|\Shist{t}}}{n-2}\int_{\bigcup\limits_{i=1}^{n-2}\zeta_{i}}p(z_{t+1}|d_t, s_{t+1})\ \text{d}z_{t+1}\\
\displaystyle  = \frac{2\sdk\sigma_{s_{t+1}|\Shist{t}}}{n-2}\left(\frac{1}{2}-\Phi(-\sdk)\right)\\
\displaystyle  = \sigma_{s_{t+1}|\Shist{t}}\eta(n,\sdk)\ .%\\
%  =& 2\sigma(\Shist{t+1})\frac{\sdk}{n}\big(\frac{1}{2}-\Phi(-\sdk)\big)
 \end{array}
\end{equation}
% \normalsize
The inequality is obtained by observing that $z^{i}$ is the center of $\zeta_{i}$ for $i =1, \ldots, n-2$ and that each interval $\zeta_{i}$ for $i =1, \ldots, n-2$ is of equal length $\bar{z}$.
The first equality is obtained by combining the definitions of $\bar{z}$, $z^{0}$, and $z^{n-1}$.
%The rest of the lines follow by substitution and algebra.
Plugging \eqref{eq:StepwiseProofTail} and \eqref{eq:StepwiseProofInner} into \eqref{eq:StepwiseProofKey} gives
$$
 |{U}_t^{*}(d_t, s_{t+1}) - Q_t^{*}(d_t, s_{t+1})| \hspace{-0.5mm}\le\hspace{-0.5mm} \Lambda(n,\sdk)\sigma_{s_{t+1}|\Shist{t}}(\ell_1 + L_{t+1}(\Shist{t+1}))
$$
A similar derivation shows that the result is also true when $n=2$. 
Using~\eqref{eq:PartitionIneq}, Lemma~\ref{th:SingleStageError} results.$\quad_\Box$\vspace{1mm}

\noindent
\textit{Main Proof}.
We will give a proof by induction on $t$ that
\begin{equation}\label{whack}
|V_t^{\epsilon}(d_t) - V_t^{*}(d_t)| \le \lambda (H-t)\ .
\end{equation}
The base case of $t=H$ is trivially true.
 Supposing \eqref{whack} holds for $t+1$ (i.e., induction hypothesis),
 we will prove that it holds for $0 \leq t < H$.
For all $s_{t+1}\in\Adom(s_{t})$,
 \begin{equation}\label{eq:MultistageErrorTilde}
% \hspace{-1.9mm}
 \begin{array}{l}
\displaystyle |{Q}_t^{\epsilon}(d_t, s_{t+1}) - Q_t^{*}(d_t, s_{t+1})|\vspace{1mm}\\
\displaystyle  \le |{Q}_t^{\epsilon}(d_t, s_{t+1})\hspace{-0.5mm} - \hspace{-0.5mm}{U}_t^{*}(d_t, s_{t+1})|  + \underbrace{|{U}_t^{*}(d_t, s_{t+1})\hspace{-0.5mm} - \hspace{-0.5mm}Q_t^{*}(d_t, s_{t+1})|}_{\le \lambda}\\
 \displaystyle \le \lambda + \sum_{i=0}^{n-1}w^{i} \left|{V}_{t+1}^{\epsilon}(\Shist{t+1}, \Zhist{t} \oplus z^{i})- V_{t+1}^{*}(\Shist{t+1}, \Zhist{t} \oplus z^{i})\right|\\
  \le \lambda(H-t)\ .
  \end{array}
 \end{equation}
 The second inequality follows from Lemma~\ref{th:SingleStageError} and the definitions of ${Q}_t^{\epsilon}$ \eqref{eq:EpsilonValFunDef} and ${U}_t^{*}$.
 The last inequality follows from the induction hypothesis and the fact that $\sum^{n-1}_{i=0} w^{i} =1$. Then,
$$
\hspace{-1.9mm}
  \begin{array}{l}
\displaystyle   |{V}_t^{\epsilon} - V_t^{*}(d_t)|\\
\displaystyle   \le \left|\max_{s_{t+1} \in \Adom(s_t)} {Q}_t^{\epsilon}(d_t, s_{t+1}) - \max_{s_{t+1}' \in \Adom(s_t)} Q_t^{*}(d_t, s_{t+1}')\right|   \\
\displaystyle   \le \max_{s_{t+1} \in \Adom(s_t)} \left| {Q}_t^{\epsilon}(d_t, s_{t+1}) - Q_t^{*}(d_t, s_{t+1})\right|\\
\displaystyle   \le \lambda(H-t)\ .
  \end{array}
$$
 The final inequality is due to \eqref{eq:MultistageErrorTilde}. 
 %Thus, Theorem~\ref{th:MultistageError} holds by induction.
%
\section{Analytically Computing a Feasible Choice of $\tau$ and $n$}
\label{yayat}
Set 
\begin{equation}
n=2+\sdk\sqrt\frac{{\pi}}{{2}}\exp\hspace{-1mm}\left(\frac{\sdk^2}{2}\right) .
\label{toure}
\end{equation}
Then,
$$
\begin{array}{l}
\displaystyle \Lambda(n,\sdk)\\
\displaystyle= \kappa(\sdk)+\eta(n,\sdk)\vspace{1mm}\\
\displaystyle \leq \sqrt{\frac{2}{\pi}} \exp\hspace{-1mm}\left(-\frac{\sdk^2}{2}\right) + \frac{\sdk}{n-2}\vspace{1mm}\\
\displaystyle= 2\sqrt{\frac{2}{\pi}} \exp\hspace{-1mm}\left(-\frac{\sdk^2}{2}\right).\end{array}
$$
Using \eqref{eq:PartitionIneq}, the feasible values of $\tau$ satisfy 
$$2\sqrt{\frac{2}{\pi}} \exp\hspace{-1mm}\left(-\frac{\sdk^2}{2}\right) \leq \frac{\lambda}{\sigma_{s_{t+1}|\Shist{t}}(\ell_1 + L_{t+1}(\Shist{t+1}))}\ .$$
So, a feasible choice of $\tau$ can be obtained from  
$$2\sqrt{\frac{2}{\pi}} \exp\hspace{-1mm}\left(-\frac{\sdk^2}{2}\right) = \frac{\lambda}{\sigma_{s_{t+1}|\Shist{t}}(\ell_1 + L_{t+1}(\Shist{t+1}))}\ ,$$
which gives 
%\begin{equation} 
$$
\sdk = \sqrt{-2\log\hspace{-1mm}\left(\sqrt{\frac{\pi}{2}}\frac{\lambda}{2\sigma_{s_{t+1}|\Shist{t}}(\ell_1 + L_{t+1}(\Shist{t+1}))}\right)}\ .
$$
%\label{checkc}
%\end{equation}
By substituting this value of $\sdk$ into \eqref{toure}, the value of $n$ can be obtained. If $n$ is not an integer, then take the ceiling of $n$.
%Note that $\sdk$ and $n$ can be expressed in terms of $\epsilon$ by substituting $\lambda=\epsilon/( H(H+1))$ into \eqref{toure} and \eqref{checkc}.

If the deterministic sampling size $n$ is to be constrained (i.e., less than some user-defined constant $\widehat{n}$) to preserve the time efficiency of solving for the $\epsilon$-GPP policy, then a numerical alternative is available:
%The value of $n$ should be sufficiently small (say less than some pre-defined constant $\widehat{n}$) to preserve time efficiency.
For all $n<\widehat{n}$, $\min_{\sdk}\Lambda(n, \sdk)$ can be computed numerically offline.
%This can be computed for all $n<\hat{n}$ offline. 
Choosing the values of $\sdk$ and $n$ to satisfy \eqref{eq:PartitionIneq} simply involves finding $n^*<\widehat{n}$ such that $\min_{\sdk}\Lambda(n^*, \sdk) \le \lambda/(\sigma_{s_{t+1}|\Shist{t}}(\ell_1 + L_{t+1}(\Shist{t+1})))$.
%Alternatively, 
%However, for practical purposes, we know that $n$ should not be too large - $n$ should be smaller than some predefined constant $\hat{n}$.
%
\section{Proof of Theorem~\ref{th:PolicyLoss}}
\label{hender}
%Theorem~\ref{th:MultistageError} bounds the error in our approximate value functions. Intuitively, the policy loss incurred should be no greater than twice the sum of these errors, summed over all $H$ stages.
%This is reflected in the following lemma.
%\begin{lemma} \label{th:PolicyLossStagewise}
%For all $0 \le t \le H$ and all reachable $d_t$, we have:
We will give a proof by induction on $t$ that
 \begin{equation}\label{sad}
  V_t^*(d_t)-V_t^{\pi^{\epsilon}}(d_t) \le \lambda (H-t)(H-t+1)\ .
 \end{equation}
%\end{lemma}
%
The base case of $t=H$ is trivially true.
Supposing \eqref{sad} holds for $t+1$ (i.e., induction hypothesis),  we will prove that it holds for $0 \leq t < H$.
For all $s_{t+1}\in\Adom(s_{t})$,
 \begin{equation}
 \label{eq:FullMultistageErrorTilde}
 \hspace{-1.9mm}
  \begin{array}{l}
\displaystyle   |Q_t^{*}(d_t, s_{t+1})-Q_t^{\pi^{\epsilon}}(d_t, s_{t+1})|\\
 \displaystyle  \le\int\limits_{-\infty}^{\infty}p(z_{t+1}|d_t, s_{t+1})\left|V_{t+1}^{*}(d_{t+1})-V_{t+1}^{\pi^{\epsilon}}(d_{t+1})\right| \text{d}z_{t+1}\\
 \displaystyle  \le \lambda(H-t-1)(H-t)\ .
  \end{array}
 \end{equation}
 The first inequality follows from definitions of $Q_t^{*}(d_t, s_{t+1})$ \eqref{eq:OptimalValFunDef} and $Q_t^{\pi^{\epsilon}}(d_t, s_{t+1})$.
 % \eqref{eq:ValFunDef}.
The last inequality follows from the induction hypothesis. 
 \begin{equation*}
  \begin{array}{l}
\displaystyle    V_t^*(d_t)-V_t^{\pi^{\epsilon}}(d_t)\\
\displaystyle    \le |V_t^{\pi^{\epsilon}}(d_t)-V_t^{{\epsilon}}(d_t)|+\underbrace{|V_t^{{\epsilon}}(d_t)-V_t^*(d_t)|}_{\le \lambda(H-t)}\vspace{1mm}\\
\displaystyle    \le \lambda(H-t) + |Q_t^{\pi^{\epsilon}}(d_t, \pi^{\epsilon}(d_t))-{Q}_t^{\epsilon}(d_t, \pi^{\epsilon}(d_t))|\vspace{1mm}\\
\displaystyle    \le \lambda(H-t) + \underbrace{|Q_t^{\pi^{\epsilon}}(d_t, \pi^{\epsilon}(d_t))-Q_t^*(d_t, \pi^{\epsilon}(d_t))|}_{\le \lambda(H-t-1)(H-t)} +  \underbrace{|Q_t^*(d_t, \pi^{\epsilon}(d_t))-{Q}_t^{\epsilon}(d_t, \pi^{\epsilon}(d_t))|}_{\le \lambda(H-t)}\vspace{1mm}\\    
\displaystyle    \le \lambda(H-t)(H-t+1)\ .
  \end{array}
 \end{equation*}
%The first is from the triangle inequality.
The second inequality requires Theorem~\ref{th:MultistageError}.
% and the claim that the policy $A$ outputs action $s^{A}_{T+1}$, which was induced by the values produced by $U$ and $\tilde{U}$.
%The third line is from the triangle inequality. 
The last inequality is due to~\eqref{eq:MultistageErrorTilde} and~\eqref{eq:FullMultistageErrorTilde}.
By substituting $t=0$ and $\lambda=\epsilon/(H(H+1))$, Theorem~\ref{th:PolicyLoss} results.

\section{Anytime $\epsilon$-GPP}
\label{aegpp}
Algorithm~\ref{alg:ds-anytime} gives the pseudocode of our anytime $\epsilon$-GPP. We discuss a few important issues:
\begin{algorithm} 
\caption{Anytime $\epsilon$-GPP} \label{alg:ds-anytime}
 \begin{algorithmic}[1] 
 \algrenewcommand\algorithmicindent{0.7em}%
% \begin{comment}
%  \Function {Compute$U$}{$t, d_t, \lambda$}
%  \If {$t=H$} {\Return $\langle 0,null \rangle$} \EndIf
%   \ForAll{$s_{t+1}$ in $\Adom(s_t)$}
%    \State \Call{Compute$\tilde{U}$}{$t, d_t, s_{t+1}, \lambda$}
%   \EndFor
%   \Return $\langle \text{best value}, \text{best action} \rangle$
%  \EndFunction
%  \Function {Compute$\tilde{U}$}{$t, d_t, s_{t+1}, \lambda)$}
%   \State \Call{FetchSamplesAndWeights}{$t, d_t, s_{t+1}, \lambda$} \label{algline:dsSelectPartitions}
%   \ForAll{$z^{(j)}$ in FetchedSamples}
%    \State \Call{Compute$\tilde{U}$}{$t+1, \langle \Shist{t} \oplus s_{t+1}, \Zhist{t} \oplus z^{(j)}\rangle, \lambda$}
%   \EndFor
%   \Return sum of values weighted by $w^{(j)}$
%  \EndFunction
%  \end{comment}
  \Function {ExpandTree}{$t, d_t, \lambda$}
   \If {$t=H$} \Return $\langle 0, 0 \rangle$ \EndIf
   \ForAll{$s_{t+1}$ in $\Adom(s_t)$}
    \State $\{ \langle z^{i} , w^{i} \rangle \}_{i=0,\ldots,n-1} \leftarrow$ \Call{GenerateSamples}{$t, d_t,s_{t+1}, \lambda$}\Comment{Deterministic sampling}
    \ForAll{$z^{i}$}
%     \State $d^{i}_{t+1} \leftarrow \langle \Shist{t+1}, \Zhist{t} \oplus z^{i} \rangle$
     \State $\obar{V}_{t+1}^{*}(\langle \Shist{t+1}, \Zhist{t} \oplus z^{i} \rangle) \leftarrow \infty, \ubar{V}_{t+1}^{*}(\langle \Shist{t+1}, \Zhist{t} \oplus z^{i} \rangle) \leftarrow -\infty$
    \EndFor
    \State $\bar{i}\leftarrow \lfloor n/2 \rfloor$\label{algline:DefaultMeasurement}
    \State $\langle \obar{V}_{t+1}^{*}(\langle \Shist{t+1}, \Zhist{t} \oplus z^{\bar{i}} \rangle),\ubar{V}_{t+1}^{*}(\langle \Shist{t+1}, \Zhist{t} \oplus z^{\bar{i}} \rangle)\rangle \leftarrow$ \Call{ExpandTree}{$t+1, \langle \Shist{t+1}, \Zhist{t} \oplus z^{\bar{i}} \rangle, \lambda$}
    \State \Call{RefineBounds}{$t, d_t,s_{t+1},  {\bar{i}}, \lambda$}
    \State $r \leftarrow g_{\sigma_{s_{t+1}|\Shist{t}}}(\mu_{s_{t+1}|d_{t}}) + \rfn_3(\Shist{t+1})$
    \State $\obar{Q}_{t}^{*}(d_t, s_{t+1}) \leftarrow r + \left(\sum^{n-1}_{i=0} w^{i}( \rfn_1(z^{i}) + \obar{V}_{t+1}^{*}(\langle \Shist{t+1}, \Zhist{t} \oplus z^{i} \rangle))\right) +\lambda$\label{Q1}
    \State $\ubar{Q}_{t}^{*}(d_t, s_{t+1}) \leftarrow r + \left(\sum^{n-1}_{i=0} w^{i}( \rfn_1(z^{i}) + \ubar{V}_{t+1}^{*}(\langle \Shist{t+1}, \Zhist{t} \oplus z^{i} \rangle))\right) - \lambda$\label{Q2}
   \EndFor
   \State $\obar{V}_{t}^{*}(d_t)=\max_{s_{t+1} \in \Adom(s_t)}\obar{Q}_{t}^{*}(d_t, s_{t+1})$
   \State $\ubar{V}_{t}^{*}(d_t)=\max_{s_{t+1} \in \Adom(s_t)}\ubar{Q}_{t}^{*}(d_t, s_{t+1})$
   \State \Return $\langle \obar{V}_{t}^{*}(d_t), \ubar{V}_{t}^{*}(d_t) \rangle$
  \EndFunction
  \Function {RefineBounds}{$t, d_t, s_{t+1}, k, \lambda$}
%          \State $d' \leftarrow \langle \Shist{t+1}, \Zhist{t} \oplus z^k \rangle$ 
    \State $\{ \langle z^{i} , w^{i} \rangle \}_{i=0,\ldots,n-1} \leftarrow$ \Call{RetrieveSamples}{$t, d_t,s_{t+1}, \lambda$} 
    \ForAll{${i} \neq k$}
%      \State $d^i_{t+1} \leftarrow \langle \Shist{t+1}, \Zhist{t} \oplus z^{i} \rangle$
      \State $\Delta z \leftarrow |z^{i}-z^k|$, $b \leftarrow L_{t+1}(\Shist{t+1})\Delta z$ 
      \State $\obar{V}_{t+1}^{*}(\langle \Shist{t+1}, \Zhist{t} \oplus z^{i} \rangle) \leftarrow \min(\obar{V}_{t+1}^{*}(\langle \Shist{t+1}, \Zhist{t} \oplus z^{i} \rangle), \obar{V}_{t+1}^{*}(\langle \Shist{t+1}, \Zhist{t} \oplus z^{k} \rangle) + b)$\label{lc1}
      \State $\ubar{V}_{t+1}^{*}(\langle \Shist{t+1}, \Zhist{t} \oplus z^{i} \rangle) \leftarrow \max(\ubar{V}_{t+1}^{*}(\langle \Shist{t+1}, \Zhist{t} \oplus z^{i} \rangle), \ubar{V}_{t+1}^{*}(\langle \Shist{t+1}, \Zhist{t} \oplus z^{k} \rangle) - b)$\label{lc2}
    \EndFor
  \EndFunction
  \Function {ConstructTree}{$t, d_t, \lambda$}
   \If {$d_{t}$ has been explored}
    \State $s_{t+1} \leftarrow \argmax_{s'_{t+1} \in \Adom(s_t)}\obar{Q}_{t}^{*}(d_t, s'_{t+1})$
        \State $\{ \langle z^{i} , w^{i} \rangle \}_{i=0,\ldots,n-1} \leftarrow$ \Call{RetrieveSamples}{$t, d_t,s_{t+1}, \lambda$} 
 %       \ForAll{$z^{i}$}
 %    \State $d^{i}_{t+1} \leftarrow \langle \Shist{t+1}, \Zhist{t} \oplus z^{i} \rangle$    
  %   \EndFor
    \State $i^* \leftarrow \argmax_{i\in\{0,\ldots,n-1\}} w^i \left(\obar{V}_{t+1}^{*}(\langle \Shist{t+1}, \Zhist{t} \oplus z^{i} \rangle)- \ubar{V}_{t+1}^{*}(\langle \Shist{t+1}, \Zhist{t} \oplus z^{i} \rangle)\right)$
    \State $\langle \obar{V}_{t+1}^{*}\hspace{-0.5mm}(\hspace{-0.5mm}\langle \Shist{t+1}, \Zhist{t} \oplus z^{i^*}\hspace{-0.3mm} \rangle\hspace{-0.5mm}),\ubar{V}_{t+1}^{*}\hspace{-0.5mm}(\hspace{-0.5mm}\langle \Shist{t+1}, \Zhist{t} \oplus z^{i^*} \hspace{-0.3mm}\rangle\hspace{-0.5mm})\rangle\hspace{-0.18mm}\leftarrow$ \Call{ConstructTree}{$t+1,\langle \Shist{t+1}, \Zhist{t} \oplus z^{i^*}\hspace{-0.3mm} \rangle,\lambda$}
    \State \Call{RefineBounds}{$t, d_t, s_{t+1}, i^*, \lambda$}
    \State $r \leftarrow g_{\sigma_{s_{t+1}|\Shist{t}}}(\mu_{s_{t+1}|d_{t}}) + \rfn_3(\Shist{t+1})$
    \State $\obar{Q}_{t}^{*}(d_t, s_{t+1}) \leftarrow r + \left(\sum^{n-1}_{i=0} w^{i}( \rfn_1(z^{i}) + \obar{V}_{t+1}^{*}( \langle \Shist{t+1}, \Zhist{t} \oplus z^{i} \rangle))\right) + \lambda$\label{Q3}
    \State $\ubar{Q}_{t}^{*}(d_t, s_{t+1}) \leftarrow r + \left(\sum^{n-1}_{i=0} w^{i}( \rfn_1(z^{i}) + \ubar{V}_{t+1}^{*}(\langle \Shist{t+1}, \Zhist{t} \oplus z^{i} \rangle))\right) - \lambda$\label{Q4}
    \State $\obar{V}_{t}^{*}(d_t)\leftarrow\max_{s'_{t+1} \in \Adom(s_t)}\obar{Q}_{t}^{*}(d_t, s'_{t+1})$
    \State $\ubar{V}_{t}^{*}(d_t)\leftarrow\max_{s'_{t+1} \in \Adom(s_t)}\ubar{Q}_{t}^{*}(d_t, s'_{t+1})$
    \State \Return $\langle \obar{V}_{t}^{*}(d_t), \ubar{V}_{t}^{*}(d_t) \rangle$
   \Else
    \State \Return \Call{ExpandTree}{$t,d_{t},\lambda$}
   \EndIf
  \EndFunction
  %\Function {Algorithm~\ref{alg:ds-anytime}}{$d_0$, $\epsilon$}
  \Function {Anytime-$\epsilon$-GPP}{$d_0$, $\epsilon$}
   \ForAll{reachable $\Shist{t+1}$ from $\Shist{0}$}
    % \State Precompute $\sigma(\Shist{t}), \mathbf{B}(\Shist{t}), \xi(\Shist{t})$ 
    \State precompute $\sigma_{s_{t+1}|\Shist{t}}$ and $L_{t+1}(\Shist{t+1})$ for all $\Shist{t+1}$ and $t=0,\ldots,H-1$
   \EndFor
   \State $\lambda \leftarrow \epsilon/(H(H+1))$ 
   \While {resources permit}
    \State $\langle \obar{V}_{0}^{*}(d_0), \ubar{V}_{0}^{*}(d_0) \rangle\leftarrow$ \Call{ConstructTree}{$0, d_0, \lambda$}
   \EndWhile
   \State $\alpha\leftarrow \obar{V}_{0}^{*}(d_0) - \ubar{V}_{0}^{*}(d_0)$
   %\State \Return $\argmax\limits_{s_{1} \in \Adom(s_0)}\big(\obar{Q_{0}^{*}}(d_0, s_{1}) + \ubar{Q_{0}^{*}}(d_0, s_{1})\big)/2$ \label{algline:MCTSrealaction}
   \State \Return ${\pi}^{\alpha\epsilon}(d_0)\leftarrow\argmax_{s_{1} \in \Adom(s_0)}\ubar{Q}_{0}^{*}(d_0,s_{1})$
  \EndFunction
  %\STATE operation1 
 \end{algorithmic}
\end{algorithm}
\squishlisttwo
\item The \textsc{ConstructTree} function needs to ensure that the chosen branch is not already fully expanded; we have omitted this for brevity.
This implies that the ``rank'' of the tree should be maintained while performing expansions -- this allows us to quickly determine if a node is already fully expanded.
\item The result below proves that $\obar{V}_{t}^{*}(d_t)$ and $\ubar{V}_{t}^{*}(d_t)$ are upper and lower heuristic bounds for ${V}_{t}^{*}(d_t)$, respectively:\vspace{1mm}
\begin{theorem}
For $t=0,\ldots,H$, 
\begin{equation}
\ubar{V}_t^{*}(d_t)\leq V_t^{*}(d_t)\leq \obar{V}_t^{*}(d_t)\ .\vspace{1mm}
\label{updown}
\end{equation}
\label{updownt}
\end{theorem}
\textit{Proof}. We will give a proof by induction on $t$. The base case of $t=H$ is trivially true.  Supposing \eqref{updown} holds for $t+1$ (i.e., induction hypothesis),
 we will prove that it holds for $0 \leq t < H$. For all $s_{t+1}\in\Adom(s_{t})$,
$$
\begin{array}{l}
Q_t^{*}(d_t, s_{t+1})\\
 \leq U_t^{*}(d_t, s_{t+1}) + \lambda\\
 \leq\displaystyle g_{\sigma_{s_{t+1}|\Shist{t}}}(\mu_{s_{t+1}|d_{t}}) + \rfn_3(\Shist{t+1}) + \left(\sum^{n-1}_{i=0} w^{i}( \rfn_1(z^{i}) + \obar{V}_{t+1}^{*}(\langle \Shist{t+1}, \Zhist{t} \oplus z^{i} \rangle))\right) +\lambda\\
= \obar{Q}_t^{*}(d_t, s_{t+1})\ .
 \end{array}
$$
The first inequality is due to Lemma~\ref{th:SingleStageError} while the second inequality follows from the definition of ${U}_t^{*}$ in \eqref{ustar} and induction hypothesis.
The equality is by definition of $\obar{Q}_t^{*}(d_t, s_{t+1})$ in lines~\ref{Q1} and~\ref{Q3} of Algorithm~\ref{alg:ds-anytime}. It follows that $V_t^{*}(d_t)\leq \obar{V}_t^{*}(d_t)$.
Similarly, for all $s_{t+1}\in\Adom(s_{t})$,
\begin{equation}
\begin{array}{l}
Q_t^{*}(d_t, s_{t+1})\\
 \geq U_t^{*}(d_t, s_{t+1}) - \lambda\\
 \geq\displaystyle g_{\sigma_{s_{t+1}|\Shist{t}}}(\mu_{s_{t+1}|d_{t}}) + \rfn_3(\Shist{t+1}) + \left(\sum^{n-1}_{i=0} w^{i}( \rfn_1(z^{i}) + \ubar{V}_{t+1}^{*}(\langle \Shist{t+1}, \Zhist{t} \oplus z^{i} \rangle))\right) -\lambda\\
= \ubar{Q}_t^{*}(d_t, s_{t+1})\ .
 \end{array}
 \label{crab}
\end{equation}
The first inequality is due to Lemma~\ref{th:SingleStageError} while the second inequality follows from the definition of ${U}_t^{*}$ in \eqref{ustar} and induction hypothesis.
The equality is by definition of $\ubar{Q}_t^{*}(d_t, s_{t+1})$ in lines~\ref{Q2} and~\ref{Q4} of Algorithm~\ref{alg:ds-anytime}. It follows that $V_t^{*}(d_t)\geq \ubar{V}_t^{*}(d_t)$.$\quad_\Box$\vspace{1mm}
\item The result below exploits the Lipschitz continuity of $V^*_t$  (Corollary~\ref{th:LipschitzSingle}) for producing informed upper and lower heuristic bounds in lines~\ref{lc1} and~\ref{lc2} of Algorithm~\ref{alg:ds-anytime}, respectively:\vspace{1mm}
\begin{theorem}
For $t=0,\ldots,H$ and $i,k=0,\ldots, n-1$, 
$$
\ubar{V}_{t}^{*}(\langle \Shist{t}, \Zhist{t-1} \oplus z^{k} \rangle) - L_{t}(\Shist{t})|z^{i}-z^k|\leq {V}_{t}^{*}(\langle \Shist{t}, \Zhist{t-1} \oplus z^{i} \rangle)\leq \obar{V}_{t}^{*}(\langle \Shist{t}, \Zhist{t-1} \oplus z^{k} \rangle) + L_{t}(\Shist{t})|z^{i}-z^k| .\vspace{1mm}
$$
\end{theorem}
\textit{Proof}. 
$$
\begin{array}{l}
{V}_{t}^{*}(\langle \Shist{t}, \Zhist{t-1} \oplus z^{i} \rangle)\\
\leq{V}_{t}^{*}(\langle \Shist{t}, \Zhist{t-1} \oplus z^{k} \rangle) + L_{t}(\Shist{t})|z^{i}-z^k|\\
\leq\obar{V}_{t}^{*}(\langle \Shist{t}, \Zhist{t-1} \oplus z^{k} \rangle) + L_{t}(\Shist{t})|z^{i}-z^k|\ .
\end{array}
$$
The first inequality is due to Corollary~\ref{th:LipschitzSingle} while the second inequality follows from Theorem~\ref{updownt}. Similarly,
$$
\begin{array}{l}
{V}_{t}^{*}(\langle \Shist{t}, \Zhist{t-1} \oplus z^{i} \rangle)\\
\geq{V}_{t}^{*}(\langle \Shist{t}, \Zhist{t-1} \oplus z^{k} \rangle) - L_{t}(\Shist{t})|z^{i}-z^k|\\
\geq\ubar{V}_{t}^{*}(\langle \Shist{t}, \Zhist{t-1} \oplus z^{k} \rangle) - L_{t}(\Shist{t})|z^{i}-z^k|\ .
\end{array}
$$
The first inequality is due to Corollary~\ref{th:LipschitzSingle} while the second inequality follows from Theorem~\ref{updownt}.$\quad_\Box$\vspace{1mm}
\item Finally, the performance loss of our anytime $\epsilon$-GPP policy relative to that of GPP policy $\pi^*$ can be bounded:\vspace{1mm}
\begin{theorem}
Suppose that the user-specified loss bound $\epsilon>0$ is given,  our anytime $\epsilon$-GPP is terminated at $\alpha\triangleq\obar{V}_{0}^{*}(d_0)- \ubar{V}_{0}^{*}(d_0)$, and our anytime $\epsilon$-GPP policy is defined as ${\pi}^{\alpha\epsilon}(d_0)\triangleq\argmax_{s_{1} \in \Adom(s_0)}\ubar{Q}_{0}^{*}(d_0, s_{1})$. Then, 
$V_0^*(d_0)-V_0^{\pi^{\alpha\epsilon}}(d_0)\leq\alpha H$ by setting and substituting $\lambda=\epsilon/( H(H+1))$ into the choice of $\tau$ and $n$ stated in Remark $2$ in Section~\ref{ch:SamplingStrategy}.\vspace{1mm}
\label{fast}
\end{theorem}
\textit{Proof}. It follows that ${V}_{0}^{*}(d_0)- \ubar{V}_{0}^{*}(d_0)\leq\alpha$.
In general, given that the length of planning horizon is reduced to $H-t$ for $t=0,\ldots,H-1$, this inequality is equivalent to 
\begin{equation}
{V}_{t}^{*}(d_t)- \ubar{V}_{t}^{*}(d_t)\leq\alpha
\label{ff2}
\end{equation}
by increasing/shifting the indices of ${V}_{0}^{*}(d_0)$ and $\ubar{V}_{0}^{*}$ above from $0$ to $t$ so that these value functions start at stage $t$ instead.

We will give a proof by induction on $t$ that 
\begin{equation}
V_t^*(d_t)-V_t^{\pi^{\alpha\epsilon}}(d_t)\leq\alpha (H-t)\ .
\label{furious}
\end{equation}
The base case of $t=H$ is trivially true.
Supposing \eqref{furious} holds for $t+1$ (i.e., induction hypothesis),  we will prove that it holds for $0 \leq t < H$.
For all $s_{t+1}\in\Adom(s_{t})$,
 \begin{equation}
 \label{eq:Gosh}
 \hspace{-1.9mm}
  \begin{array}{l}
\displaystyle   |Q_t^{*}(d_t, s_{t+1})-Q_t^{\pi^{\alpha\epsilon}}(d_t, s_{t+1})|\\
 \displaystyle  \le\int\limits_{-\infty}^{\infty}p(z_{t+1}|d_t, s_{t+1})\left|V_{t+1}^{*}(d_{t+1})-V_{t+1}^{\pi^{\alpha\epsilon}}(d_{t+1})\right| \text{d}z_{t+1}\\
 \displaystyle  \le \alpha(H-t-1)\ .
  \end{array}
 \end{equation}
 The first inequality follows from definitions of $Q_t^{*}(d_t, s_{t+1})$ \eqref{eq:OptimalValFunDef} and $Q_t^{\pi^{\alpha\epsilon}}(d_t, s_{t+1})$.
 % \eqref{eq:ValFunDef}.
The last inequality follows from the induction hypothesis. 
$$
\begin{array}{l}
V_t^*(d_t)-V_t^{\pi^{\alpha\epsilon}}(d_t)\\
= V_t^*(d_t)- \ubar{V}_{t}^{*}(d_t) + \ubar{V}_{t}^{*}(d_t) -V_t^{\pi^{\alpha\epsilon}}(d_t)\\
\leq \alpha + \ubar{Q}_{t}^{*}(d_t, {\pi}^{\alpha\epsilon}(d_t)) - {Q}_{t}^{\pi^{\alpha\epsilon}}(d_t, {\pi}^{\alpha\epsilon}(d_t))\\
\leq \alpha + \underbrace{\ubar{Q}_{t}^{*}(d_t, {\pi}^{\alpha\epsilon}(d_t)) - {Q}_{t}^{*}(d_t, {\pi}^{\alpha\epsilon}(d_t))}_{\leq 0} + \underbrace{{Q}_{t}^{*}(d_t, {\pi}^{\alpha\epsilon}(d_t)) - {Q}_{t}^{\pi^{\alpha\epsilon}}(d_t, {\pi}^{\alpha\epsilon}(d_t))}_{\leq\alpha(H-t-1)}\\
\leq\alpha(H-t)\ .
\end{array}
$$
The first inequality is due to \eqref{ff2}.
The second inequality follows from~\eqref{crab} and~\eqref{eq:Gosh}. By substituting $t=0$, Theorem~\ref{fast} results.$\quad_\Box$
\squishend
%Lastly, when recursively constructing subtrees (line~\ref{algline:DefaultMeasurement}), the center-most measurement was selected, that is, $z^{+}$ is selected to be the median of all candidates $z^{i}$.
%
%\emph{Remark}. To reduce computational requirements, the function \textsc{Preprocess} computes $\sigma_{s_{t+1}|\Shist{t}}$, $\alpha(\Shist{t+1})$ and $\Sigma_{s_{t+1}\Shist{t}}\Gamma_{\Shist{t}\Shist{t}}^{-1}$ for all possible $\Shist{t+1}$.
%
%
\pagebreak
\section{Bayesian Optimization (BO) on Real-World Log-Potassium Concentration Field}
\label{treesize2}
\begin{figure}[h]
\centering
%\vspace{-3mm}
\begin{tabular}{cc}
\includegraphics[width=4.5cm]{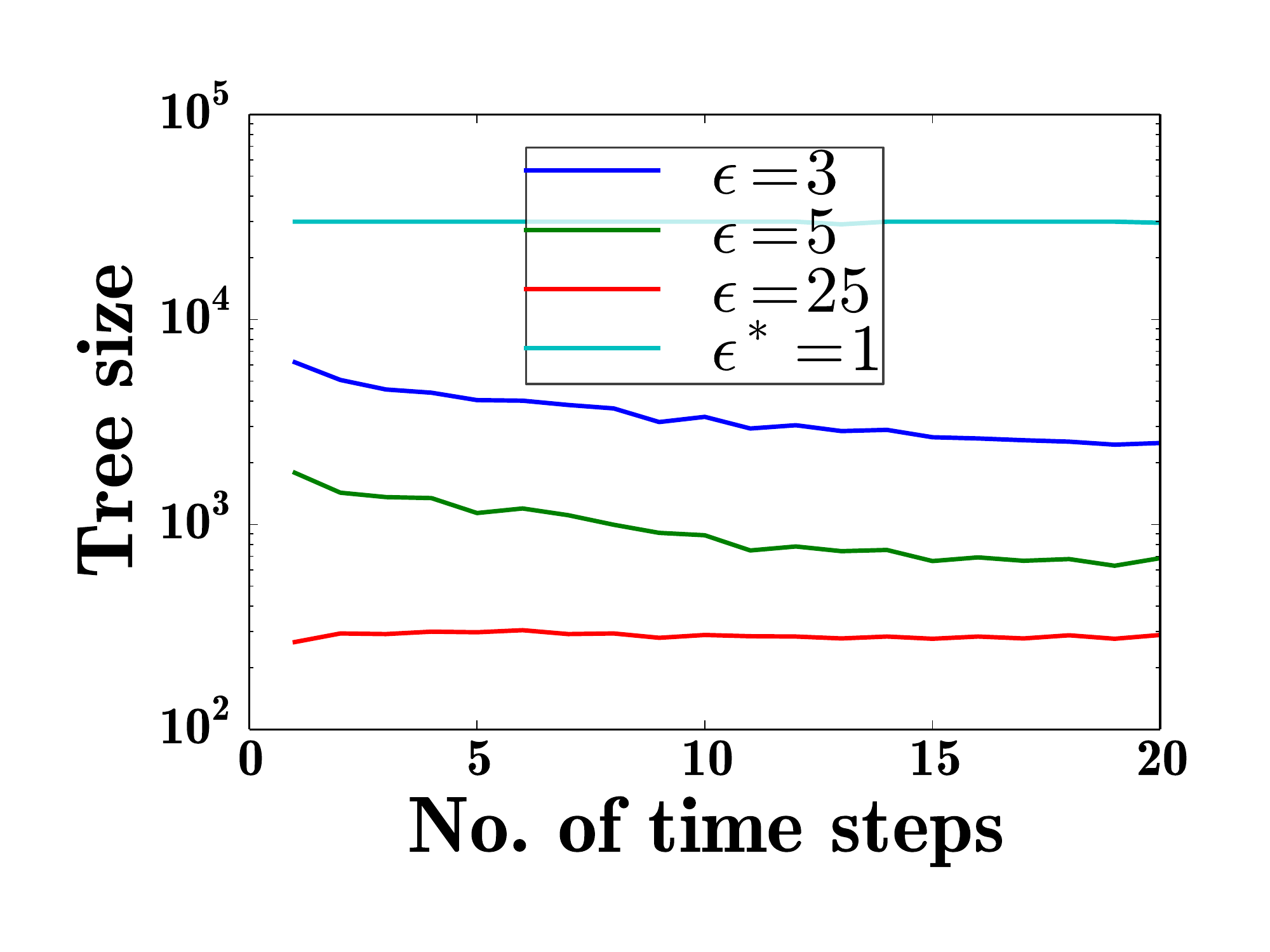} & \includegraphics[width=4.5cm]{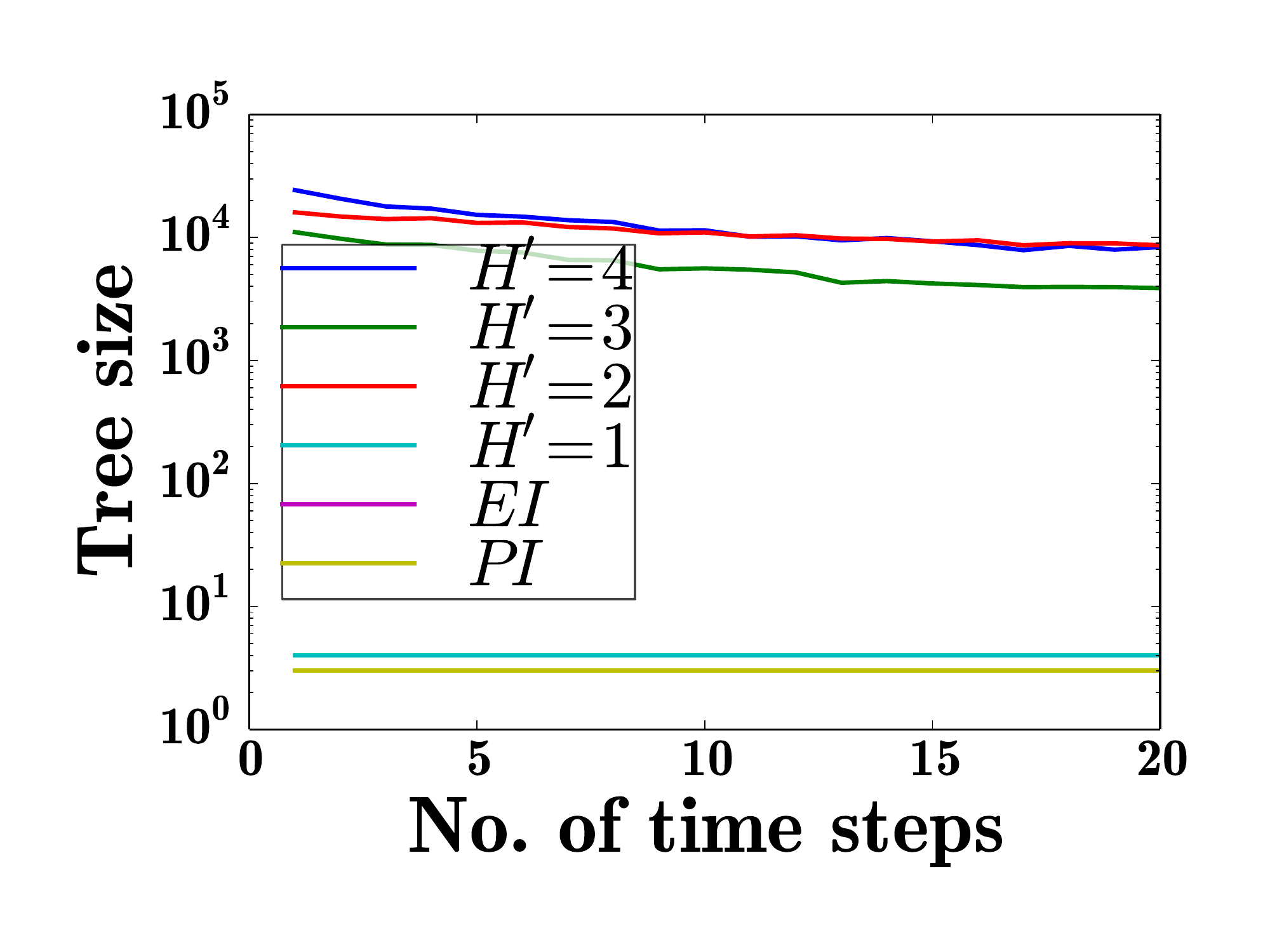}\vspace{-1mm}\\
{\scriptsize (a)} & {\scriptsize (b)} 
\end{tabular}%\vspace{-4mm}
% Figure spacing
\caption{Graphs of tree size of $\epsilon$-GPP policies using UCB-based rewards with (a) $H'=4$, $\beta=0$, and varying $\epsilon$ and (b) varying $H'=1,2,3,4$ (respectively, $\epsilon=0.002, 0.003, 0.4, 2$) and $\beta=0$ vs. no. of time steps for BO on real-world lg-K field.
The plot of $\epsilon^*=1$ utilizes our anytime variant with a maximum tree size of $3 \times 10^4$ nodes while the plot of $\epsilon=25$ effectively implements the nonmyopic UCB \cite{RamosUAI14} assuming maximum likelihood observations during planning.}
%%%%%
%\cite{RamosUAI14}
\label{fig:gpp_realdata2}
\vspace{-0mm}
\end{figure}

\begin{figure}[h]
\centering
\includegraphics{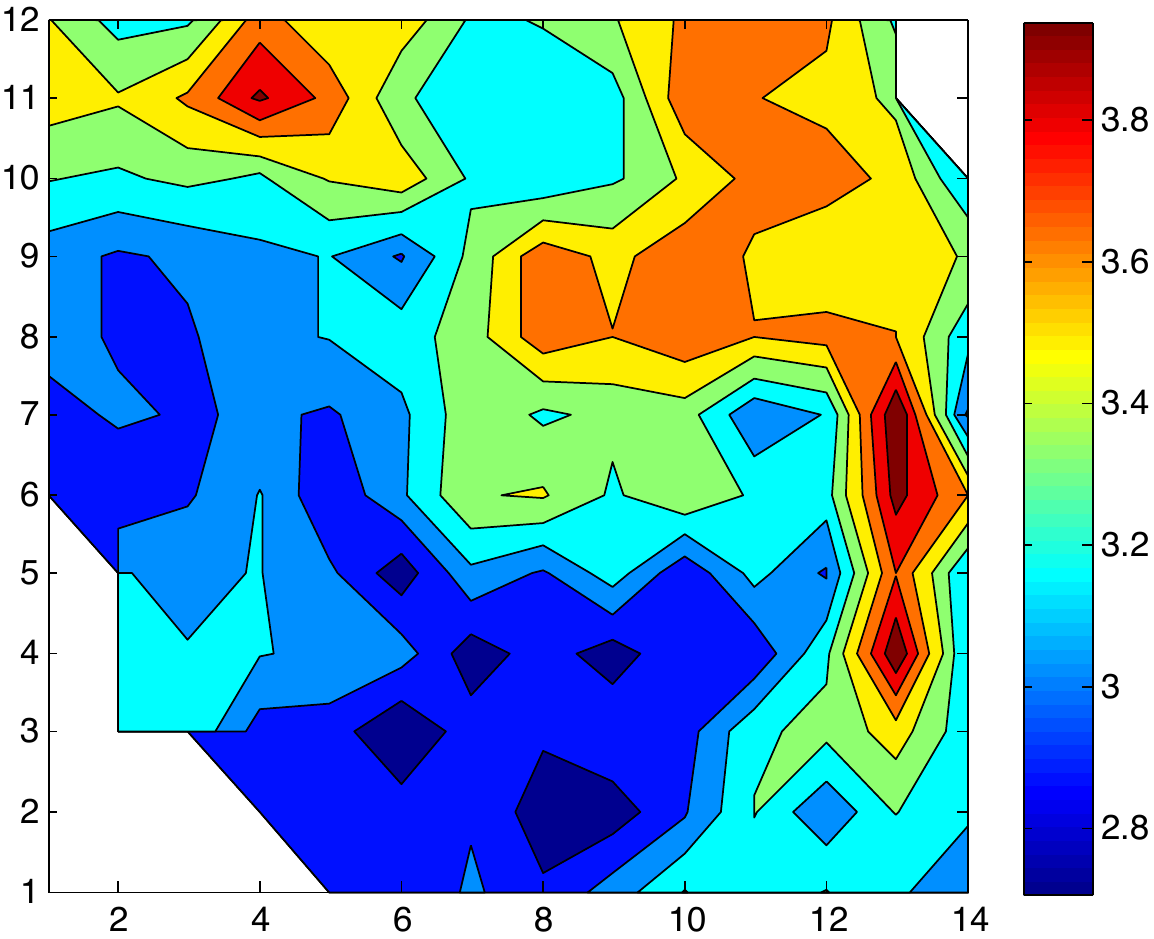}
\caption{Real-world log-potassium (lg-K) concentration (mg~l$^{-1}$) field of Broom's Barn farm.}
%%%%%
% \cite{Webster01}
\end{figure}

\section{Bayesian Optimization (BO) on Simulated Plankton Density Field}
\label{treesize1}
%\noindent
%{\bf BO on Simulated Plankton Density Field.}
%
A robotic boat equipped with a fluorometer is tasked to find the peak chl-a measurement (i.e., possibly in an algal bloom) while exploring a lake 
\cite{LowSPIE09}. It is driven by the UCB-based reward function described under `Bayesian optimization' in Section~\ref{ch:rewardfunctions}.
% In a linear reward model, the robot receives a periodic reward equivalent to its measurement made at its current time step. That is, $\rfn(z_{t+1}, \Shist{t+1}) = z_{t+1}$. Note that $\rfn_1(z_{t+1})=\rfn_3(\Shist{t+1})=0$. Some algebraic manipulation reveals that $g_{\sigma_{s_{t+1}|\Shist{t}}}(u)=u$, $\ell_1=0$, $\ell_2(\sigma_{s_{t+1}|\Shist{t}})=1$, that is, the expected reward per time step is equivalent to the robot's posterior mean.
%The length-scale used in this experiment was $l_1^2=l_2^2=0.04$.
%The signal and noise variances are $\sigma_{n}^2=10^{-5}$ and $\sigma_{y}^2=1.0$.
%

Fig.~\ref{fig:gpp_synthetic_linear} shows results of performances of our $\epsilon$-GPP policy and its anytime variant, nonmyopic UCB (i.e., $\epsilon=250$), and greedy PI, EI, UCB (i.e., $H'=1$) averaged over $30$ independent realizations of the chl-a field. 
% \cite{Brochu10,Lizotte07,Srinivas10} 
Similar to that of the energy harvesting task, the gradients of their achieved total rewards increase over time.
In particular, it can be observed from 
%Fig.~\ref{fig:gpp_synthetic_linear}, we can observe some benefits in wider tree expansions as well as the effectiveness of our anytime algorithm.
%It can be seen from
Fig.~\ref{fig:gpp_synthetic_linear}a that  nonmyopic UCB assuming maximum likelihood observations during planning obtains much less total rewards than the other $\epsilon$-GPP policies and the anytime variant and finds a maximum chl-a measurement of $1.25$ that is at least $0.26\sigma_y$ worse after $20$ time steps. The anytime variant performs reasonably well, albeit slightly worse than our $\epsilon$-GPP policy with $\epsilon=25.0$.
From Fig.~\ref{fig:gpp_synthetic_linear}b, the greedy policy (i.e., $H'=1$) performs much more poorly than its nonmyopic counterparts and finds a maximum chl-a measurement of $1.28$ that is at least $0.22\sigma_y$ worse after $20$ time steps. Such a greedy policy with $\beta=0$ is also worse than greedy PI and EI due to its lack of exploration. However, by increasing $H'$ to $3$ or $4$, our $\epsilon$-GPP policies with $\beta=0$ outperform greedy PI and EI as they can naturally and jointly optimize the exploration-exploitation trade-off. To see this,
Figs.~\ref{fig:gpp_synthetic_linear}c to~\ref{fig:gpp_synthetic_linear}f reveal that the  weighted exploration term $\beta\sigma_{s_{t+1}|\Shist{t}}$ in the UCB-based reward function becomes redundant when $H'\geq 3$; in fact, such nonmyopic $\epsilon$-GPP policies with $\beta=0$ achieve the highest total rewards among all $\epsilon$-GPP policies with $H'=1,2,3,4$ and $\beta=0,1,10,20$ and their performances degrade with stronger exploration behavior (i.e., $\beta > 0$).
%with non-myopicity (i.e., $H'\geq 3$) since $\epsilon$-GPP  policies with $\beta=0$ achieve the highest mean reward; an aggressive exploration behavior (i.e., $\beta > 0$) hurts their performances. 
Results of the tree size (i.e., incurred time) of our $\epsilon$-GPP policy and its anytime variant are similar to that of the energy harvesting task and shown in Fig.~\ref{fig:gpp_synthetic_linear2}.\vspace{0.5mm}
%
%the maximum-likelihood approach fares around $0.2\sigma_y$ worse than its other full-tree counterparts after $10$ time steps, which demonstrates the importance of obtaining accurate value estimates.
%As with logarithmic rewards, branch and bound achieves respectable performance, although it performs slightly worse than $\epsilon=25.0$.
%Varying the horizon $H$ did not seem to grant statistically significant benefits beyond the greedy instance ($H=1$ in Fig.~\ref{fig:gpp_synthetic_linear}).
%Nonetheless, the greedy policy performs significantly poorer than its counterparts, once again demonstrating the advantage of non-myopic planning.
\begin{figure}[h]
%\vspace{-1mm}
\begin{tabular}{ccc}
\hspace{-1mm}\includegraphics[width=4.5cm]{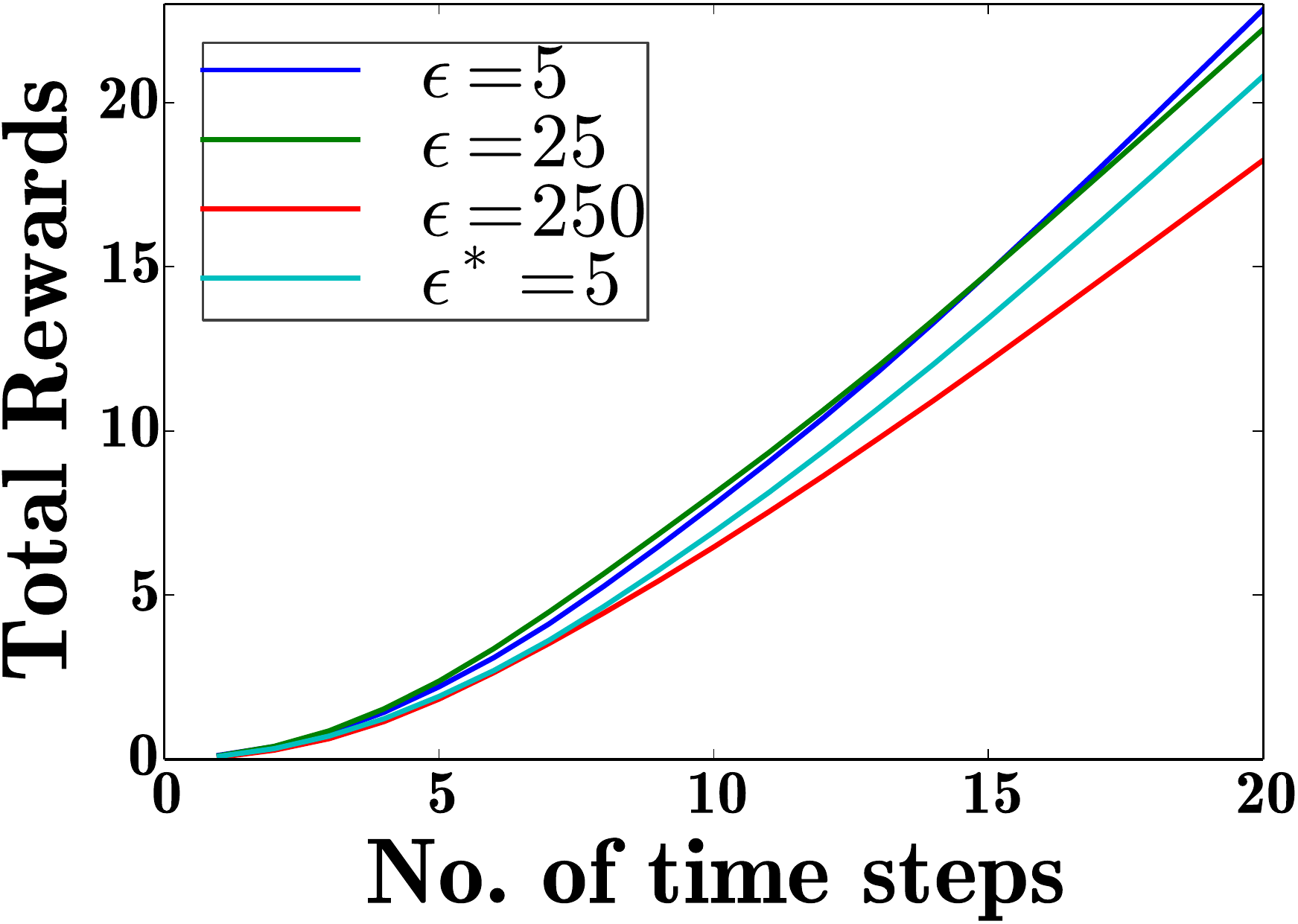} & \hspace{-3mm}\includegraphics[width=4.5cm]{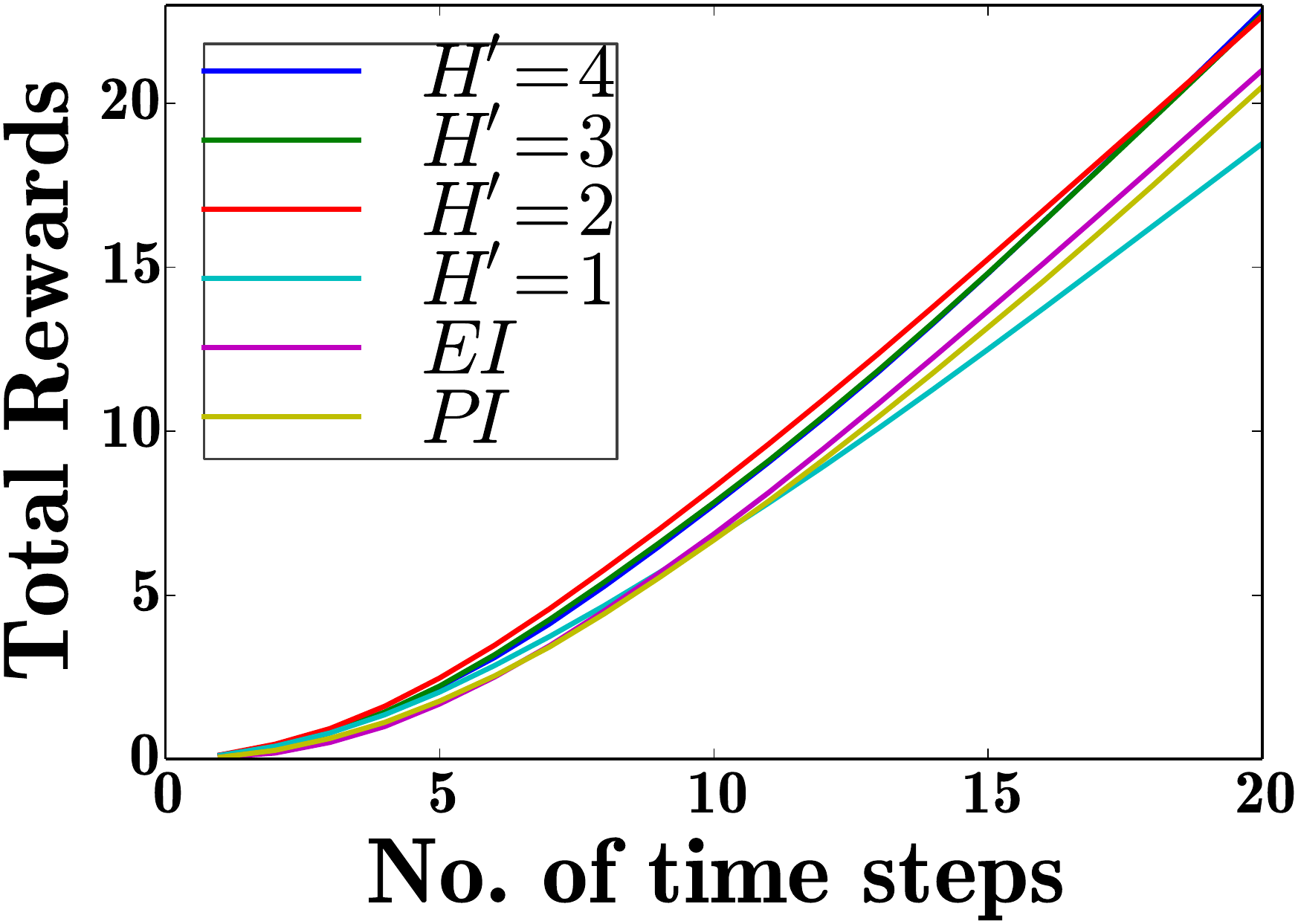} & 
\hspace{-3mm}\includegraphics[width=4.5cm]{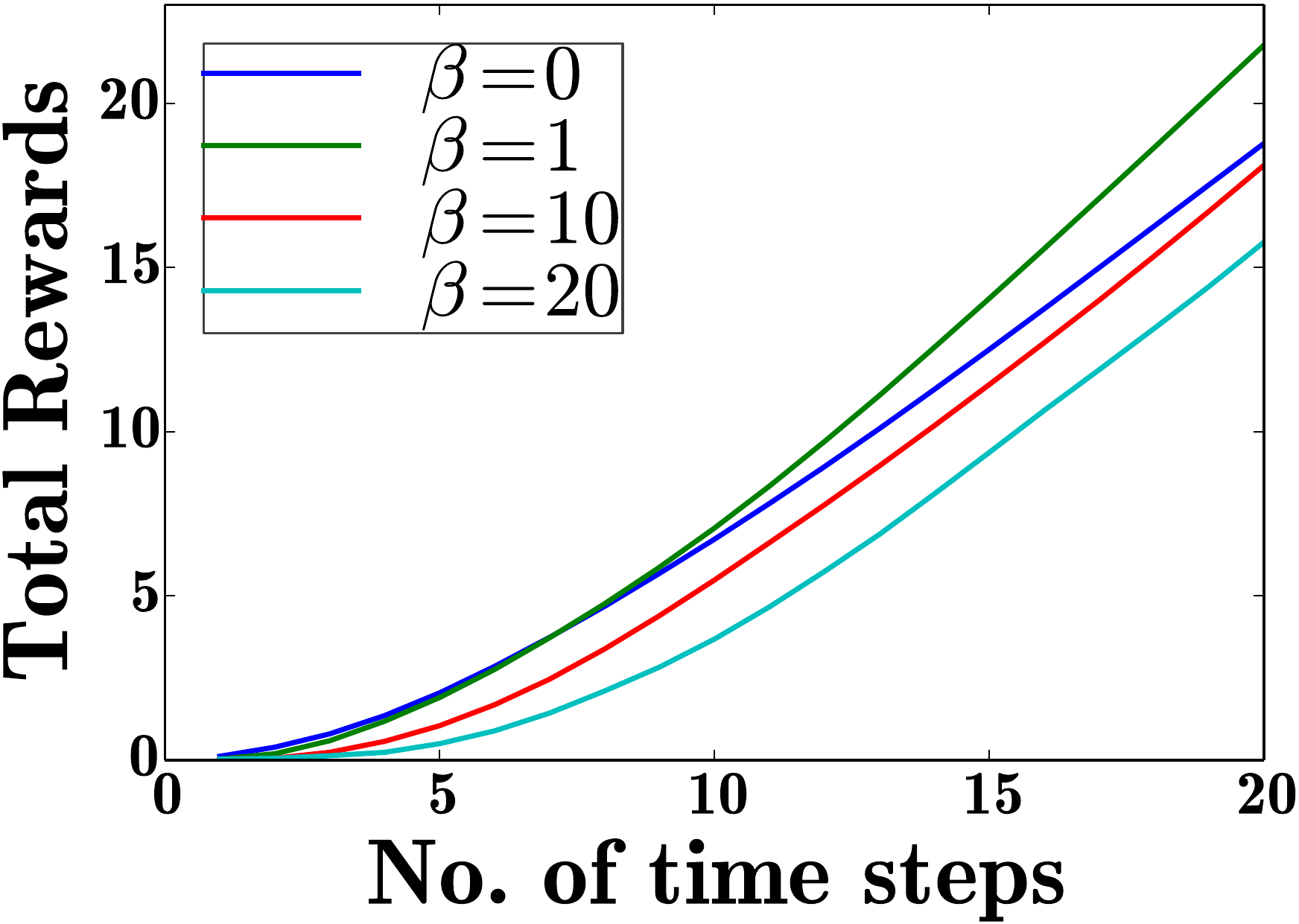}\vspace{-2mm}\\
\hspace{-0mm}{\scriptsize (a)} & \hspace{-4mm}{\scriptsize (b)} & \hspace{-4mm}{\scriptsize (c)}\vspace{-0mm}\\
\hspace{-1mm}\includegraphics[width=4.5cm]{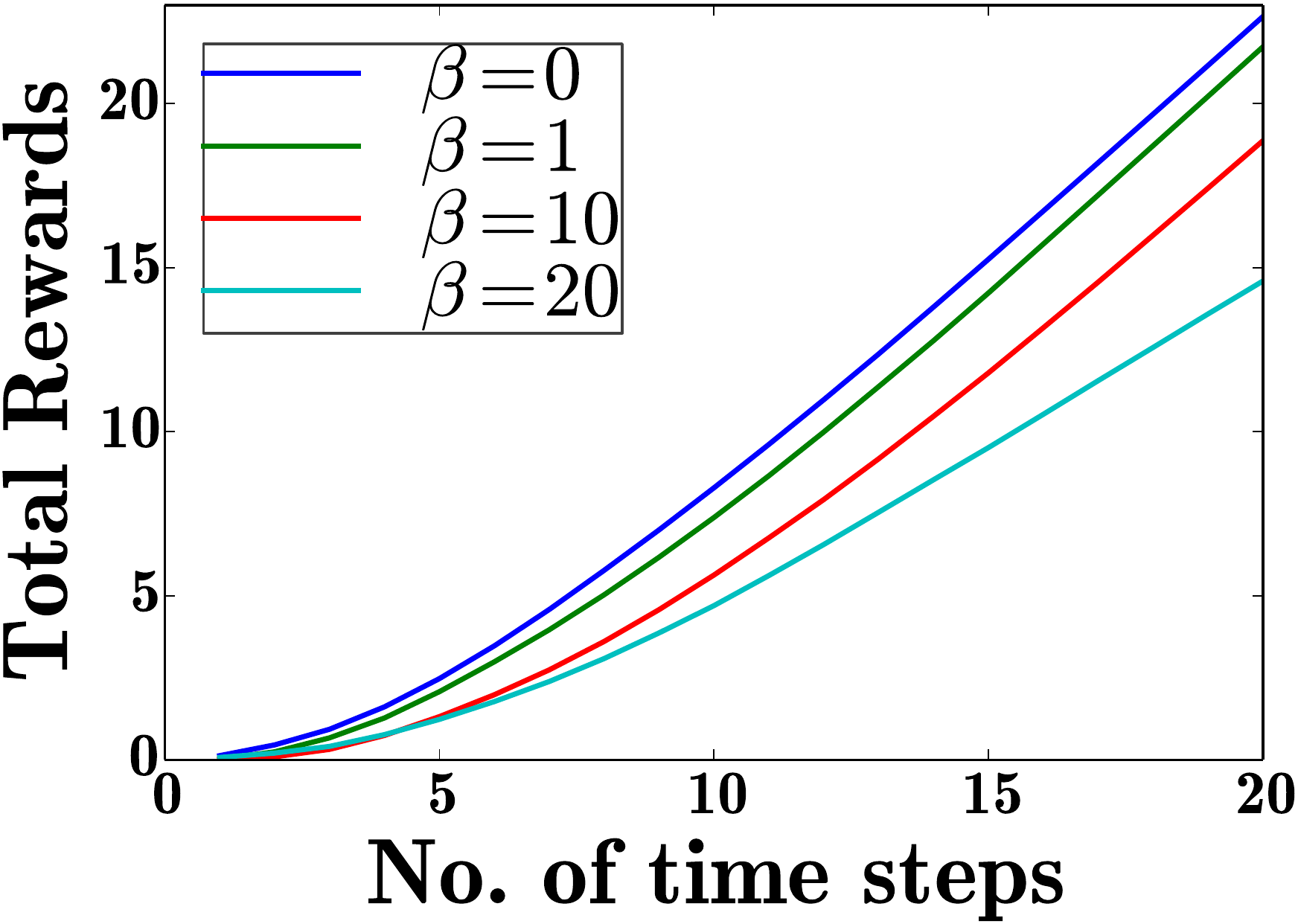} &
\hspace{-3mm}\includegraphics[width=4.5cm]{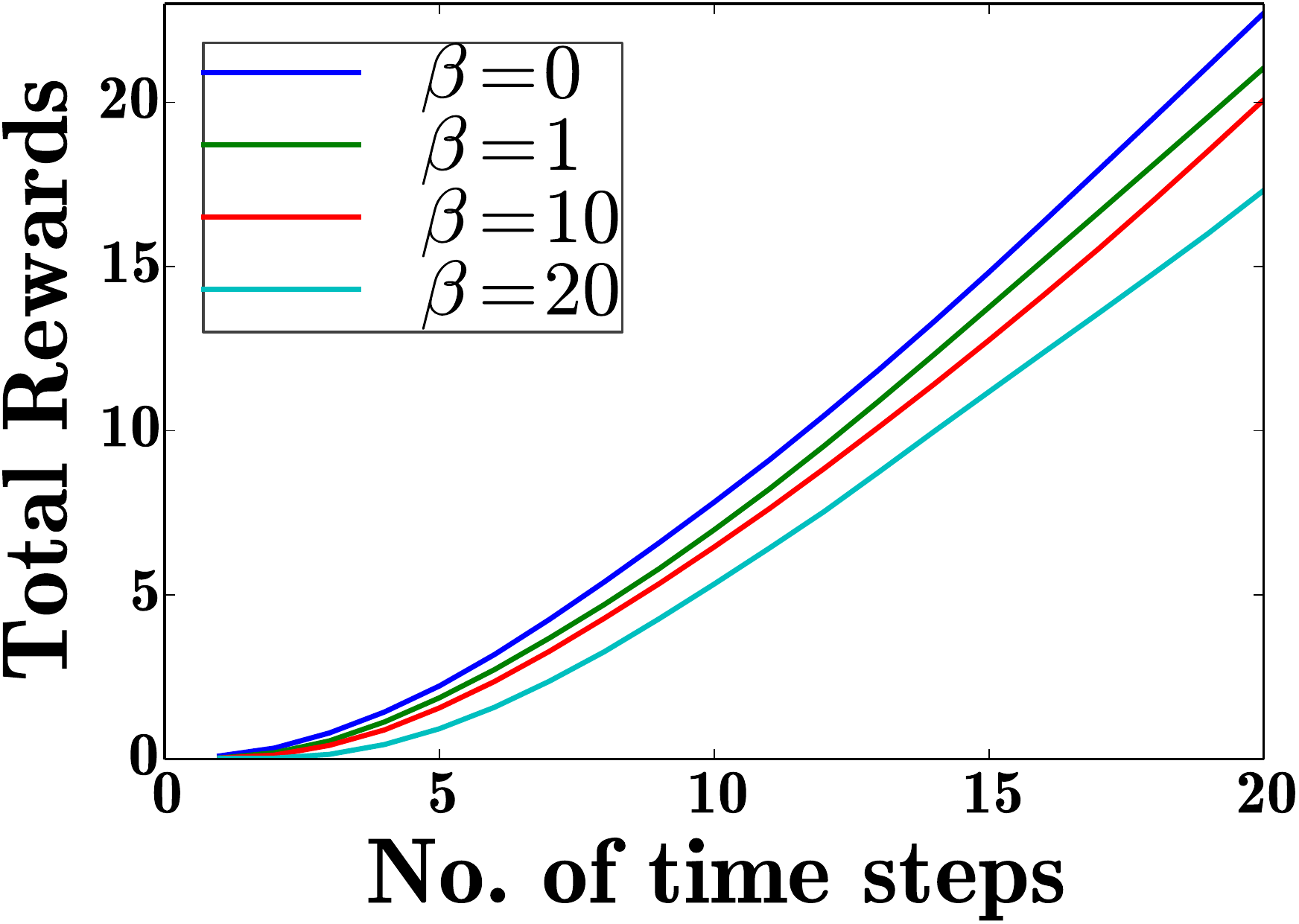} &
\hspace{-3mm}\includegraphics[width=4.5cm]{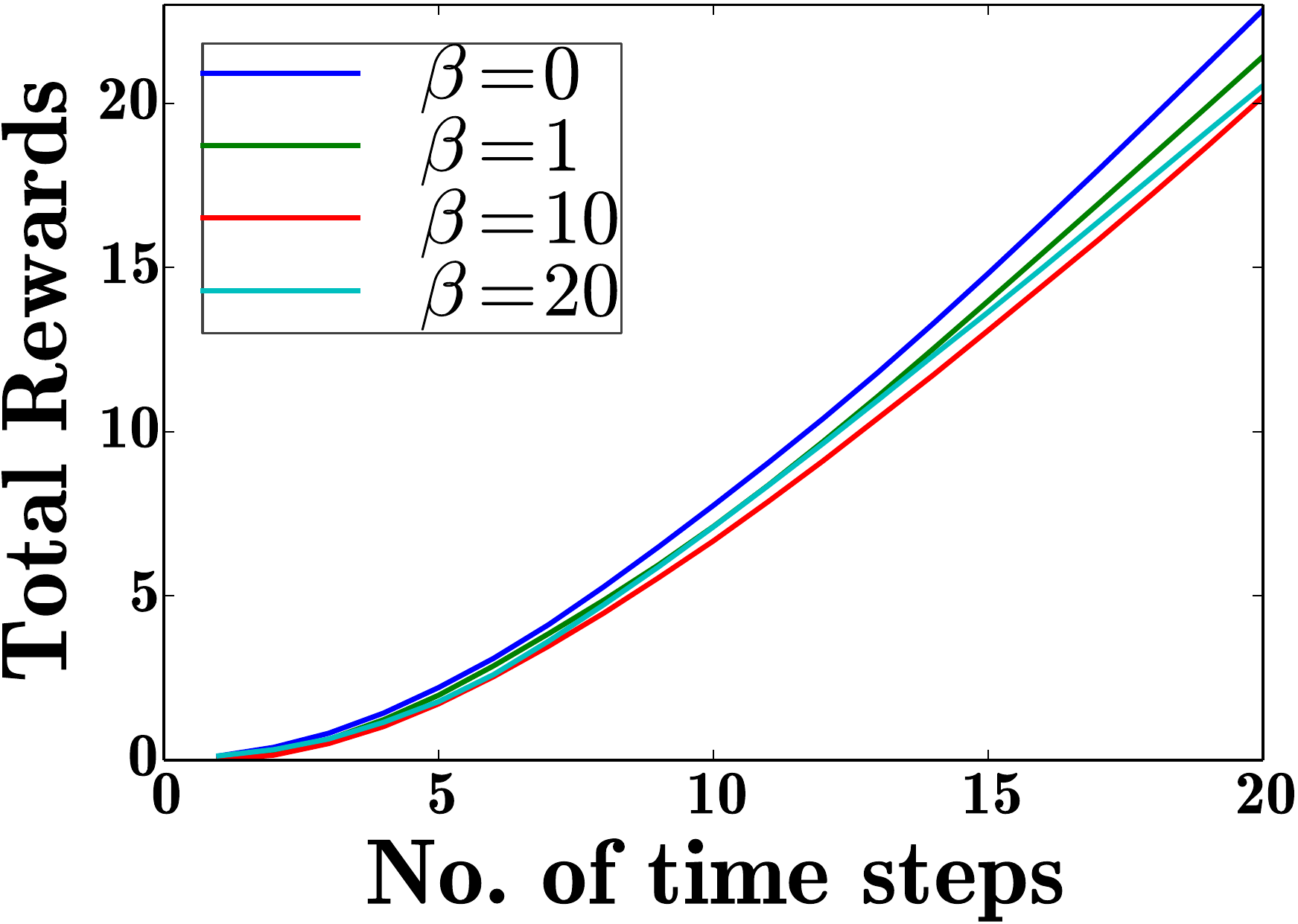} 
\vspace{-2mm}\\
 \hspace{-0mm}{\scriptsize (d)} & \hspace{-4mm}{\scriptsize (e)} & \hspace{-4mm}{\scriptsize (f)}
\end{tabular}\vspace{-4mm}
% Figure spacing
\caption{Graphs of total rewards of $\epsilon$-GPP policies using UCB-based rewards with (a) $H'=4$, $\beta=0$, and varying $\epsilon$, (b) varying $H'=1,2,3,4$ (respectively, $\epsilon=0.002, 0.015, 0.4, 5$) and $\beta=0$, (c) $H'=1$, $\epsilon=0.002$, (d) $H'=2$, $\epsilon=0.015$, (e) $H'=3$, $\epsilon=0.4$, and (f) $H'=4$, $\epsilon=5$, and varying $\beta$ vs. no. of time steps for BO on simulated chl-a field.
The plot of $\epsilon^*=5$ utilizes our anytime variant with a maximum tree size of $7.5 \times 10^4$ nodes while the plot of $\epsilon=250$ effectively implements the nonmyopic UCB \cite{RamosUAI14} assuming maximum likelihood observations during planning.\vspace{-2mm}
%%%%%
% \cite{RamosUAI14}
%(b) Performance and cost for $\epsilon$-GPP with , respectively.
}
\label{fig:gpp_synthetic_linear}
\end{figure}

\begin{figure}[h]
\centering
%\vspace{-3mm}
\begin{tabular}{cc}
\hspace{-0mm}\includegraphics[width=4.5cm]{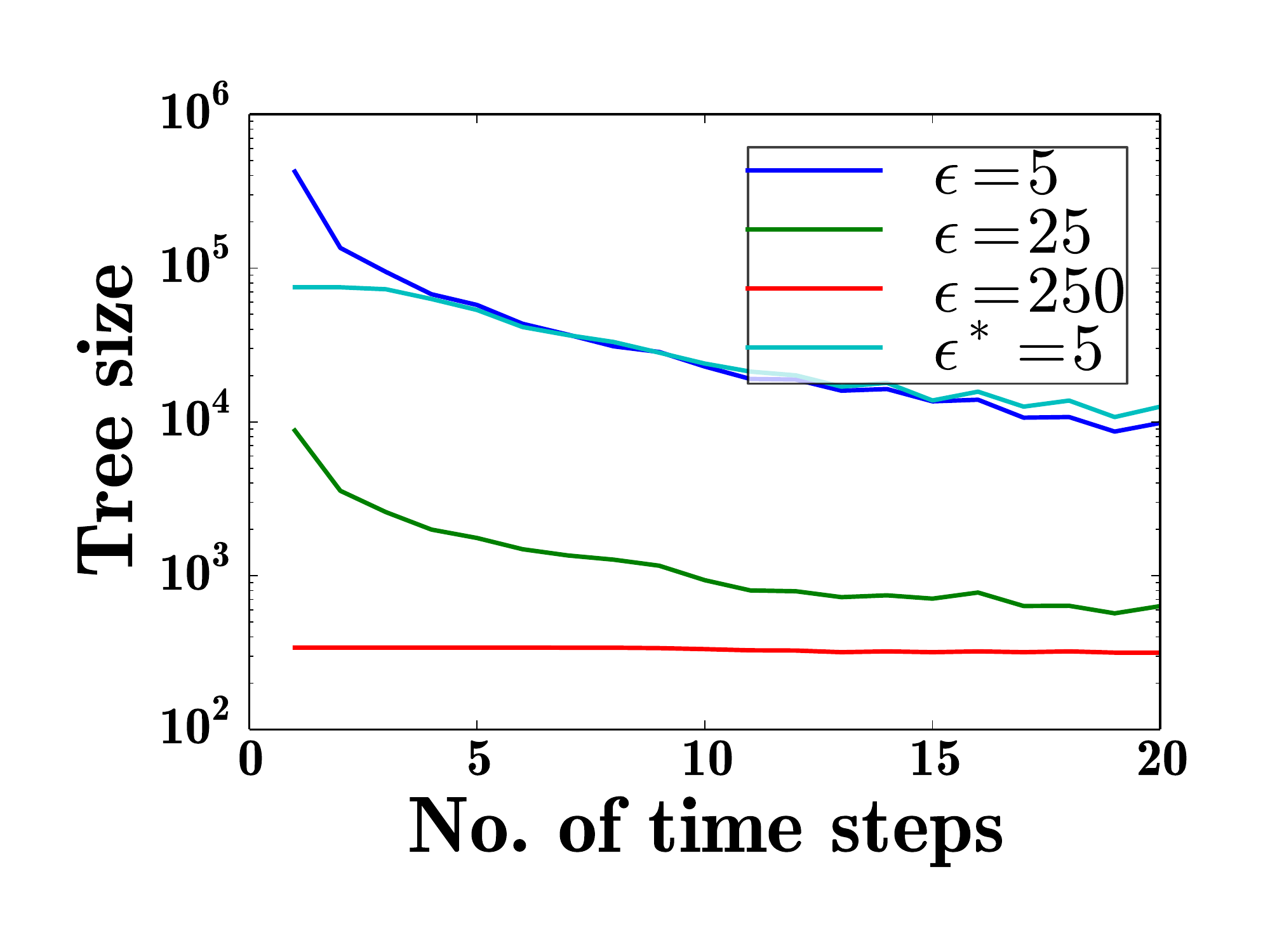} & \hspace{-0mm}\includegraphics[width=4.5cm]{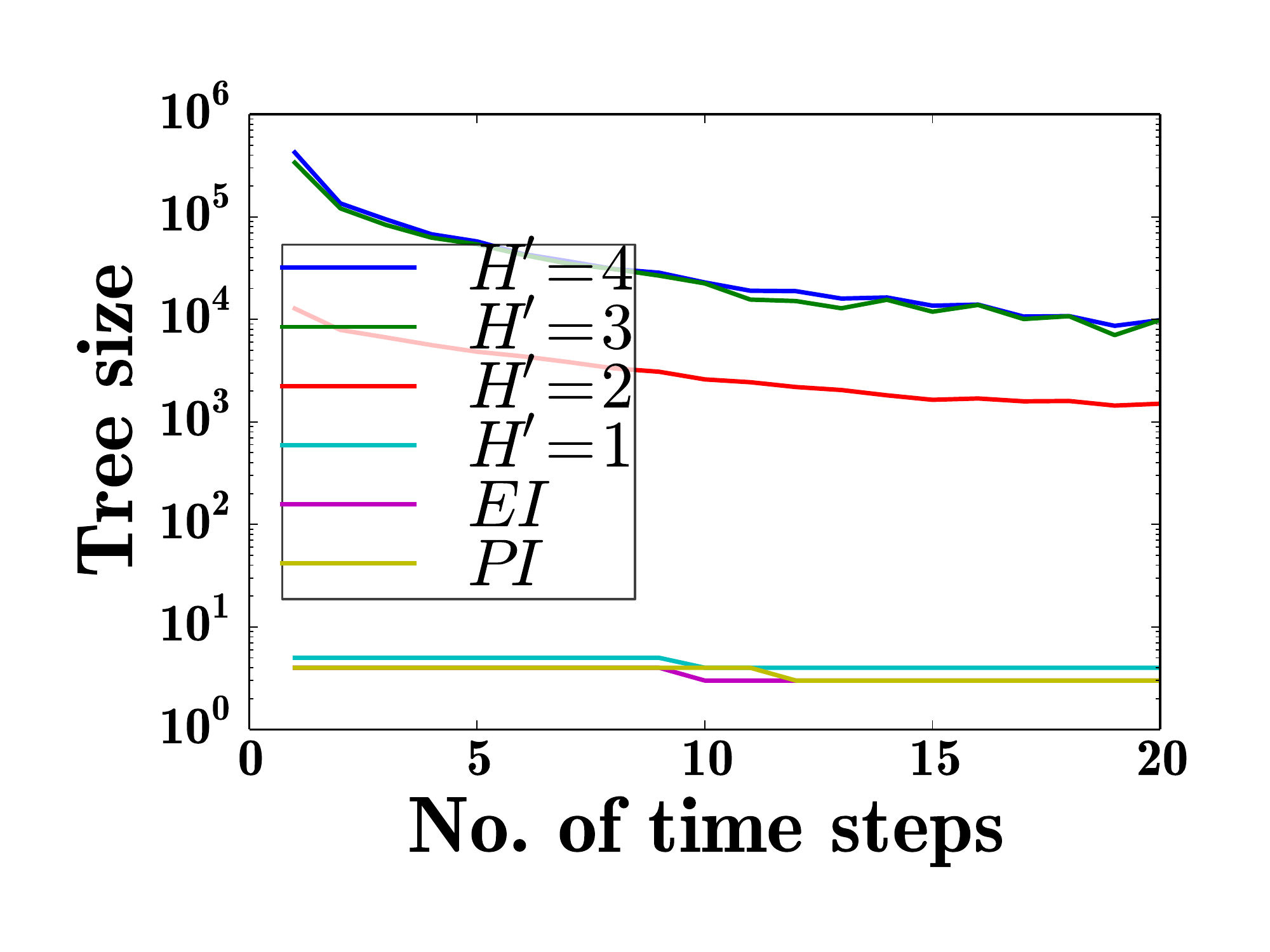}\vspace{-1mm}\\
\hspace{-0mm}{\scriptsize (a)} & \hspace{-0mm}{\scriptsize (b)} 
\end{tabular}%\vspace{-4mm}
% Figure spacing
\caption{Graphs of tree size of $\epsilon$-GPP policies using UCB-based rewards with (a) $H'=4$, $\beta=0$, and varying $\epsilon$ and (b) varying $H'=1,2,3,4$ (respectively, $\epsilon=0.002, 0.015, 0.4, 5.0$) and $\beta=0$ for BO on simulated chl-a field. The plot of $\epsilon^*=5$ utilizes our anytime variant with a maximum tree size of $7.5 \times 10^4$ nodes while the plot of $\epsilon=250$ effectively implements the nonmyopic UCB \cite{RamosUAI14} assuming maximum likelihood observations during planning.}
%%%%%
%\cite{RamosUAI14}
\label{fig:gpp_synthetic_linear2}
\vspace{-0mm}
\end{figure}

\fi

\end{document}